\documentclass[11pt,a4paper]{article}
\usepackage[hyperref]{emnlp2020}
\usepackage{times}
\usepackage{latexsym}

\usepackage{enumitem}
\usepackage{amsmath}
\usepackage{commath}
\DeclareMathOperator{\sign}{sgn}
\usepackage{graphicx}
\usepackage{longtable}
\usepackage{multirow}
\usepackage{multicol}
\usepackage{float}
\usepackage{colortbl}

\usepackage{microtype}
\usepackage[ruled,vlined]{algorithm2e}
\aclfinalcopy % Uncomment this line for the final submission
\title{DQI: Measuring Data Quality in NLP}
\author{Swaroop Mishra \:  Anjana Arunkumar \: Bhavdeep Sachdeva \: Chris Bryan \: Chitta Baral 
\\ Department of Computer Science, Arizona State University
\\ \texttt{\{srmishr1, aarunku5, bssachde, cbryan16, chitta\}@asu.edu}
}

\date{}

\begin{document}
\maketitle
\begin{abstract}
Neural language models have achieved human level performance across several NLP datasets. However, recent studies have shown that these models are not truly learning the desired task; rather, their high performance is attributed to overfitting using spurious biases, which suggests that the capabilities of AI systems have been over-estimated. We introduce a generic formula for Data Quality Index (DQI) to help dataset creators create datasets free of such unwanted biases. We evaluate this formula using a recently proposed approach for adversarial filtering, AFLite. We propose a new data creation paradigm using DQI to create higher quality data. The data creation paradigm consists of several data visualizations to help data creators (i) understand the quality of data and (ii) visualize the impact of the created data instance on the overall quality. It also has a couple of automation methods to (i) assist data creators and (ii) make the model more robust to adversarial attacks. We use DQI along with these automation methods to renovate biased examples in SNLI. We show that models trained on the renovated SNLI dataset generalize better to out of distribution tasks. Renovation results in reduced model performance, exposing a large gap with respect to human performance. DQI systematically helps in creating harder benchmarks using active learning. Our work takes the process of dynamic dataset creation forward, wherein datasets evolve together with the evolving state of the art, therefore serving as a means of benchmarking the true progress of AI. 
\end{abstract}

\section{Introduction}
Recently, a series of works \cite{gururangan2018annotation, poliak2018hypothesis, kaushik2018much, tsuchiya2018performance, tan2019investigating, schwartz2017effect, nadeem2020stereoset} has shown that many of popular datasets, such as SQUAD \cite{rajpurkar2016squad} and SNLI \cite{bowman2015large} have unwanted biases \cite{torralba2011unbiased}, resulting from the annotation process. The spurious biases represent ``unintended correlations between input and output" \cite{bras2020adversarial}. Models exploit these biases as features instead of utilizing the actual underlying features needed to solve a task. Models therefore fail to generalize, and consequently, their performance drops drastically when tested with out of distribution data or adversarial examples \cite{bras2020adversarial,mccoy2019right,zhang2019paws,jia2017adversarial,jin2019bert}. These can limit Machine Learning applications to various domains because of the possibility of serious accidents. For example, ``a medical diagnosis model may consistently classify with high confidence, even while it should flag difficult examples for human intervention. The resulting unflagged, erroneous diagnoses could blockade future machine learning technologies in medicine." \cite{hendrycks2016baseline}. These biases have also led to the overestimation of AI's true advancement \cite{sakaguchi2019winogrande,bras2020adversarial}. 

% Classifiers failing to indicate when they are likely mistaken can limit their adoption or
% cause serious accidents. For example, a medical diagnosis model may consistently classify with
% high confidence, even while it should flag difficult examples for human intervention. The resulting
% unflagged, erroneous diagnoses could blockade future machine learning technologies in medicine.
% More generally and importantly, estimating when a model is in error is of great concern to AI Safety

Hence, in lieu of merely creating and solving new datasets, the Machine Learning community needs to address a core problem, i.e., how can dataset creators create datasets that are free of unwanted biases, and thus help models generalize better? This paper focuses only on NLP, but the same principles are also applicable to other areas such as Vision and Speech.

\begin{figure*}
\includegraphics[width=\linewidth,height=10cm]{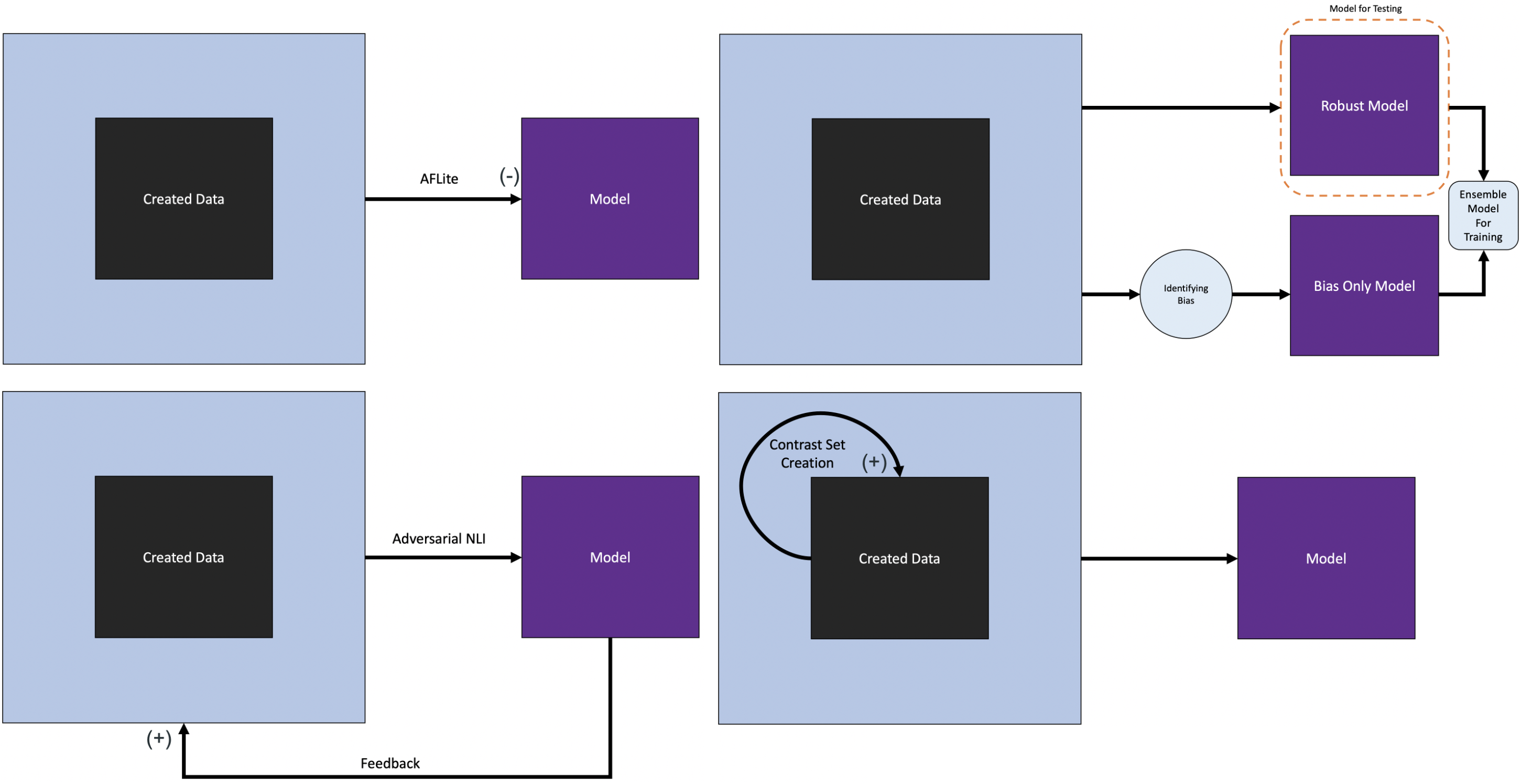}
  \caption{Existing approaches: 1a. Dataset Pruning (top left), 1b. Stopping the model from Exploiting Biases (top right), 1c. Adversarial Dataset Creation (bottom left), 1d. Counterfactual Data Augmentation (bottom right)}
\label{fig:exapproaches}
\end{figure*}

There are mainly four types of approaches to address this problem (i) Dataset pruning (ii) Stopping the model from exploiting biases (iii) Adversarial dataset creation (iv) Counterfactual Data Augmentation. Each type of approach focuses on a specific part of the loop consisting of data and model, as illustrated in Figure \ref{fig:exapproaches}. 

AFLite  \cite{sakaguchi2019winogrande}, REPAIR \cite{li2019repair}, RESOUND \cite{li2018resound} and Dataset Distillation \cite{wang2018dataset} are some of the recent works that use the first approach. AFLite filters dataset biases adversarially to attenuate the overestimation of AI systems' capabilities. On the other hand, Dataset Distillation synthesizes a minimum set of representative data to achieve close to original performance. Similarly, REPAIR resamples data to remove representation biases, and RESOUND samples existing datasets and creates a new dataset to minimize static biases. However, all these approaches do not directly impact the dataset creation process, as data pruning is only done after the data has been created by  crowd workers and/or automated systems. Post-creation, data pruning is a costly operation, as resources invested in creating the initial `biased' data get wasted. Also, these approaches do not prevent a dataset creator from creating biased data in a future data creation process.

The second approach has been studied in several works \cite{clark2019don}. They use a prior knowledge of biases to train a naive model that exploits dataset biases. Then this model is combined with a robust model, and the ensemble is trained. The ensemble is forced to focus on other patterns of data which are not biases. Similarly, DRiFt has been proposed \cite{he2019unlearn}, where initially a biased model is learned, which uses only bias related features. Then a debiased model is trained to fit the residual of the biased model. Another interesting work \cite{mahabadi2019simple} operates along the same lines, and  has an additional lightweight bias-only model which learns dataset biases. They use its prediction to adjust the loss of the base model, to reduce the biases. Apart from the overhead involved in bias identification, the drawbacks of ``wasted resources invested in creating the initial biased data" and ``not preventing dataset creators from creating biased data in future" remain in this type of approach.

Adversarial Filtering algorithm \cite{zellers2018swag} builds a de-biased dataset by iteratively training an ensemble of classifiers, and then utilizing them to filter data. However, this approach is model dependent and the drawbacks of the first two approaches still remain. Similarly, the Adversarial NLI dataset creation process \cite{nie2019adversarial} involves an iterative and adversarial "human-and-model-in-the-loop" procedure. Here, dataset creators have an additional responsibility to fool the model, and the effort required on their part increases as the rounds progress. Also, this process might create biased data itself, since it is adversarial to a specific model. Biased data is relative in nature and has significance with respect to a trained set. Since the model is not trained at every step, the adversarial dataset creation process may not produce bias free data in each and among various splits. This category of approaches might induce its own biases, as studied in a recent work \cite{liu2019inoculation} for NLI stress tests \cite{naik2018stress} and the Adversarial SQuAD dataset \cite{jia2017adversarial}.

Counterfactual Data Augmentation involves asking dataset creators to create samples with counterfactual target labels. This shouldn't disturb the sample's internal coherence, nor make unnecessary changes \cite{kaushik2019learning}. Recently, a new annotation paradigm has been proposed \cite{gardner2020evaluating} where they recommend that dataset authors manually perturb the test instances in small but meaningful ways that change the gold label, creating contrast sets. However, these approaches have too much dependence on authors in identifying a list of phenomena that characterize their dataset. Thus they can lead to the formation of a different, unique set of biases for each dataset they are applied to. Also, this approach does not prevent crowd workers from creating biased data in future.

Overall, existing approaches have seven types of issues: (i) resources invested in creating the initial `biased' data get wasted, (ii) a dataset creator is not prevented from creating biased data in a future data creation process, (iii) important aspects of bias like the dependence of bias on training set, train-test split are ignored, (iv) a set of additional biases is created as a byproduct, (v) the time complexity is high because of the involvement of training at each iteration, (vi) they are specific to a model or task, (vii) there is too much effort required on the part of crowd workers/authors/experts, without providing a suitable and illustrative feedback channel. We introduce a generic formula for DQI to address the first six issues, and a new data creation paradigm with several data visualizations and a couple of user-assistance methods to address the seventh one. 
Data Shapley \cite{ghorbani2019data} has been proposed as a metric to quantify the value of each training datum to the predictor performance. However, their approach was model dependent and task dependent. More importantly, their metric might not signify bias content, as they quantify the value of training datum based on predictor performance, and biases might favor the predictor. So, we focus on building a generic DQI with minimized dependency on models and tasks.

We take inspiration from the Quality Indexes present in other domains such as power quality \cite{bollen2000understanding}, water quality \cite{world1993guidelines}, food quality \cite{grunert2005food} and air quality \cite{jones1999indoor}. We actuate and adapt those in our approach to find the formula for DQI. First, we identify the seven components which cover the space of various possible interactions between samples in an NLP dataset.
% We use these components to introduce the skeleton of the formula for Universal DQI. 
We look for potential leads by going through a series of works which enumerate the various origins of dataset biases, and their impact on performance and robustness. We trace the leads to propose an empirical formula for DQI. We cover many datasets and a hierarchy of tasks ranging from NLI to Text Summarization in our analysis. This is to ensure that our formula is generic and is not overfitted towards a specific task or dataset. We evaluate this formula using AFLite, which is a recent and successful approach for light weight, model agnostic adversarial filtering.

\begin{figure*}
\includegraphics[width=\linewidth,height=12cm]{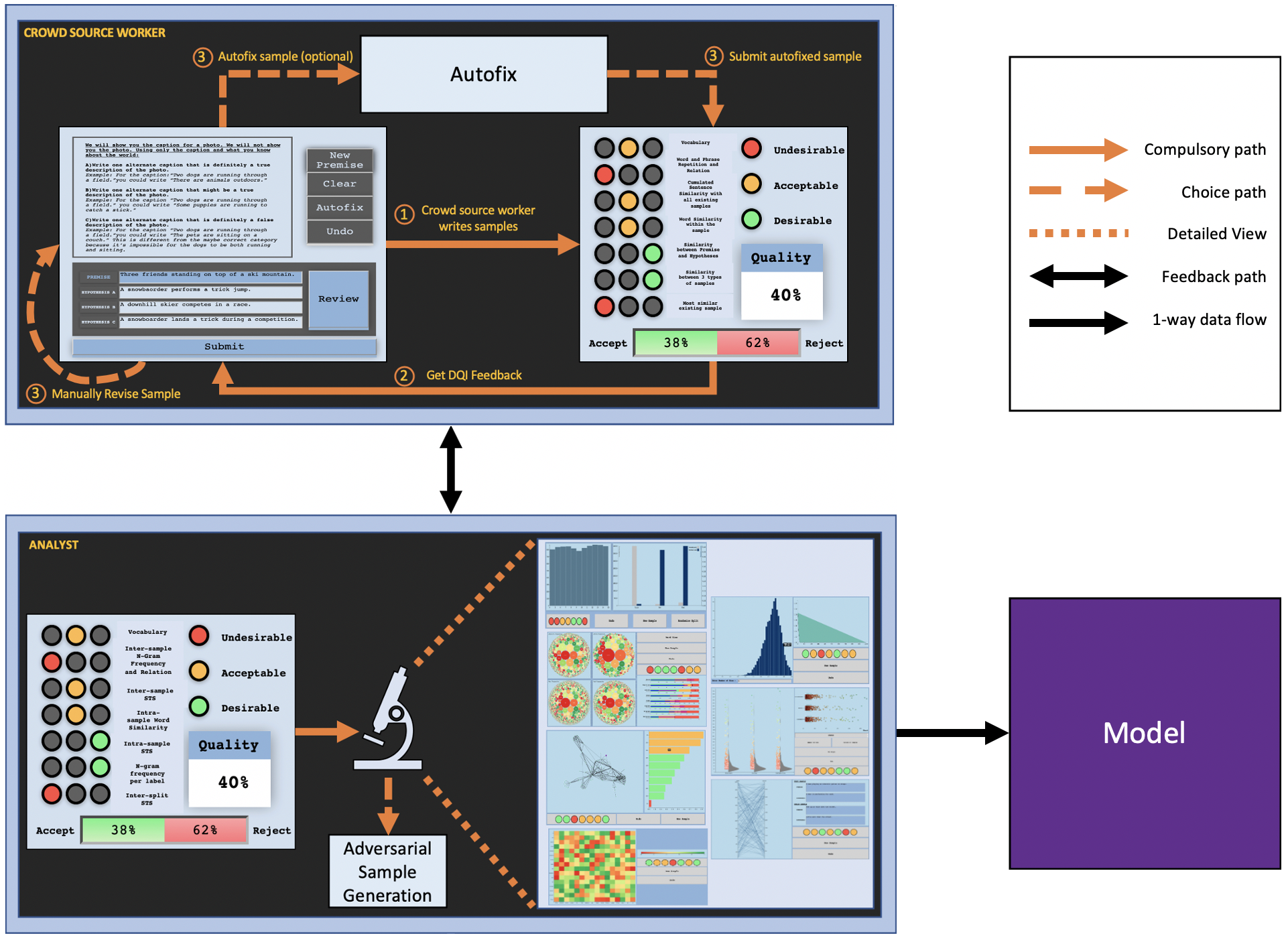}
  \caption{Our approach}
\label{fig:ourapproach}
\end{figure*}

We utilize DQI to propose a new data creation paradigm which consists of several  data  visualizations  to  help  data  creators  (i)  understand  the quality of data and (ii) visualize the impact of their created data instance on the overall quality. In a concurrent work \cite{wang2020vibe}, a tool for measuring and mitigating bias in Image datasets has been proposed. Our data creation paradigm also has a couple of automation methods to (i) assist data creators in rectifying their data creation process to minimize biases and (ii) make the model more robust to adversarial attacks. The automation methods consist of Textfooler \cite{jin2019bert}, a recent technique which has been successful in fooling the state-of-the-art models and Autofix, a model independent version of Textfooler which we propose using DQI. Figure \ref{fig:ourapproach} illustrates our proposed data creation paradigm.

Active learning has been shown to be useful for various NLP tasks \cite{li2020active, sachan2015active, garrette2013learning, kholghi2016active}. DQI systematically helps in creating harder benchmarks using active learning. We apply DQI in an active learning setup to renovate the SNLI dataset \cite{bowman2015large} using the automation methods, and produce a series of benchmarks in an increasing hierarchy of hardness. Inspired by recent datasets \cite{sakaguchi2019winogrande} \cite{nie2019adversarial}, our work takes forward the process of dynamic dataset creation wherein datasets evolve together with the evolving state of the art, therefore serving as a means of benchmarking the true progress of AI. 

We also show  that  models  trained on the  renovated SNLI dataset generalize better to out of distribution tasks. Our work supports the findings of an interesting recent work \cite{bras2020adversarial} where they indicate that biases make benchmarks easier, as models learn to exploit these biases instead of learning actual features. 

\begin{figure*}
\includegraphics[width=\linewidth]{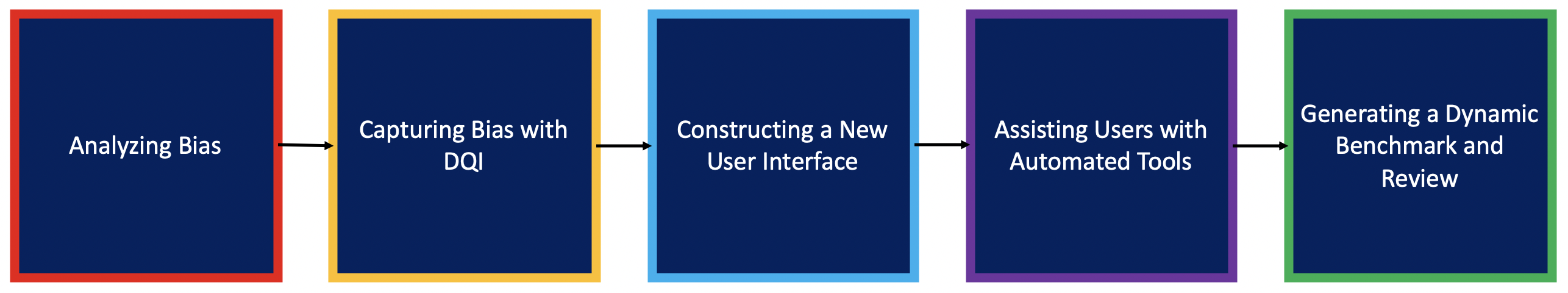}
  \caption{Process Flow}
\label{fig:overall}
\end{figure*}

\begin{figure*}
\includegraphics[width=\linewidth]{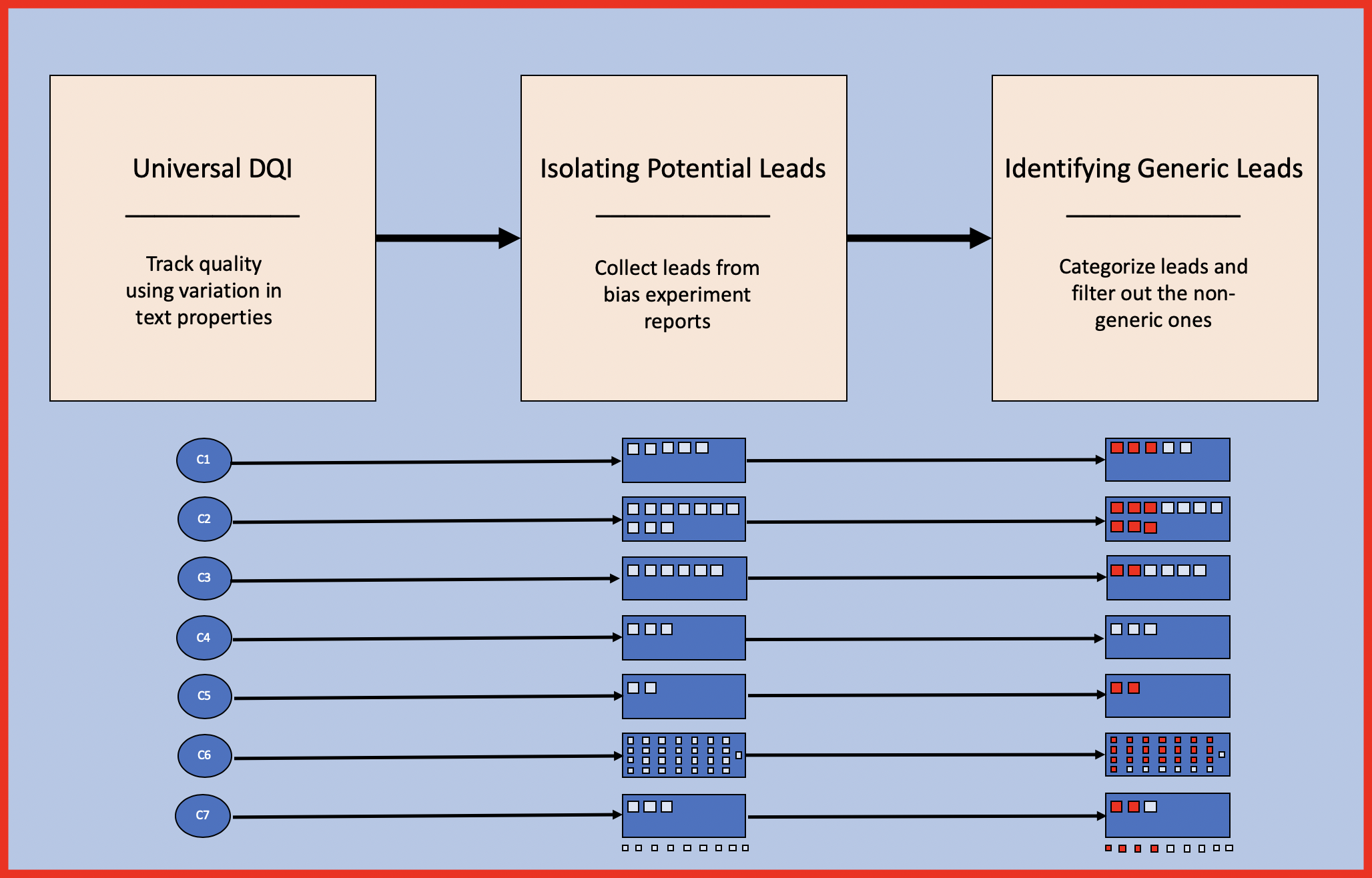}
  \caption{Step 1}
\label{fig:step1}
\end{figure*}

\begin{figure*}
\includegraphics[width=\linewidth]{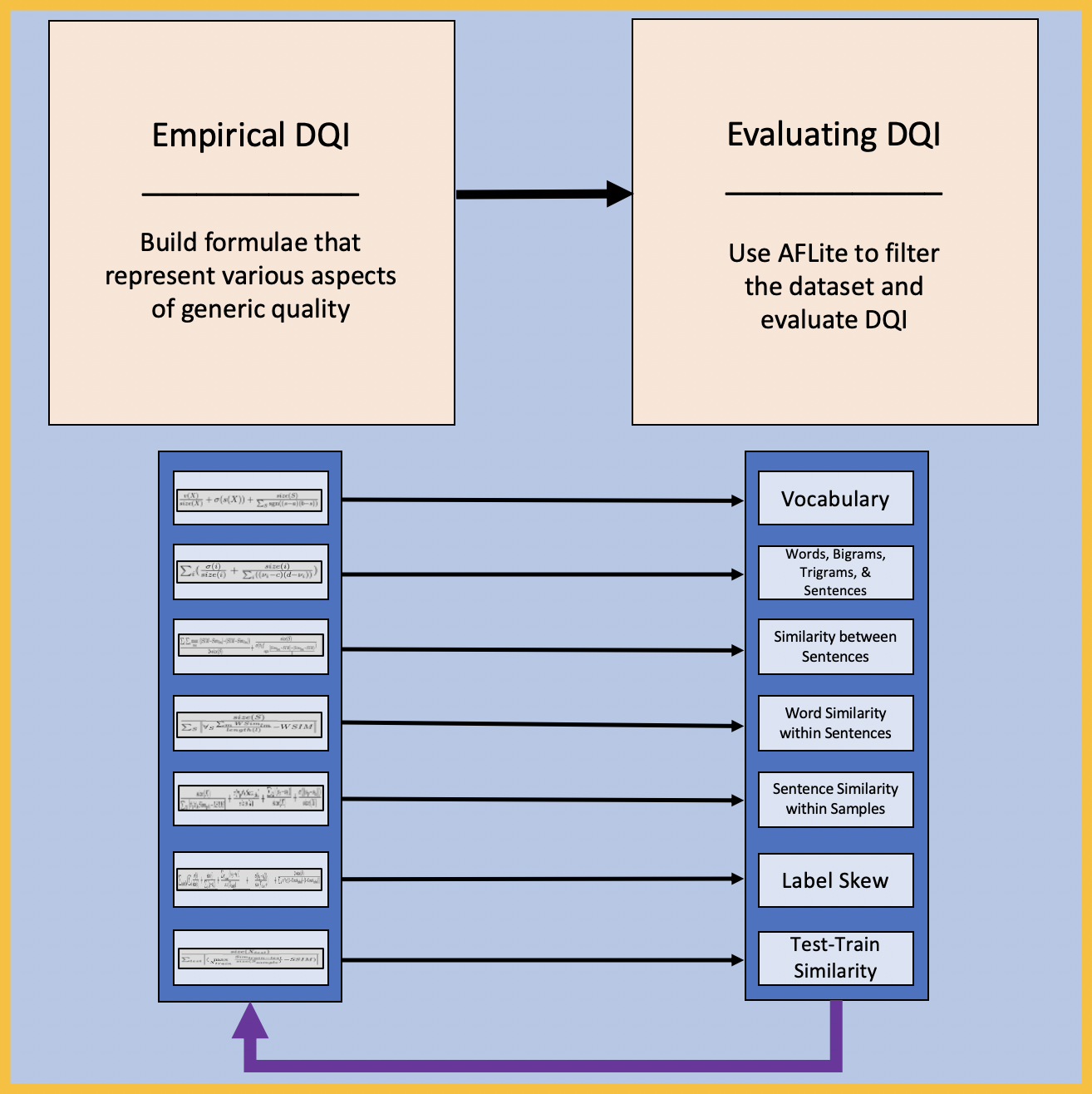}
  \caption{Step 2}
\label{fig:step2}
\end{figure*}

\begin{figure*}
\includegraphics[width=\linewidth]{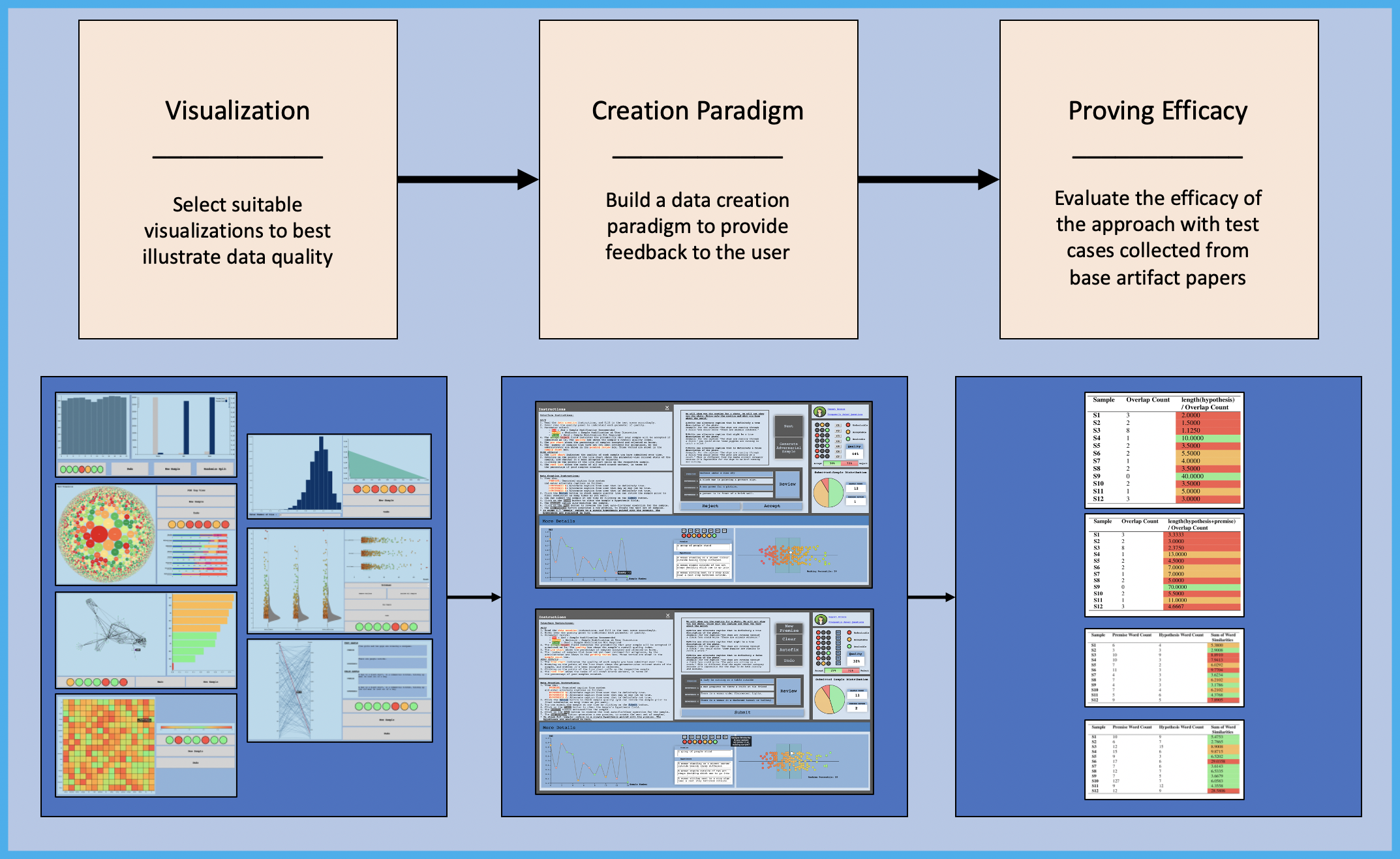}
  \caption{Step 3}
\label{fig:step3}
\end{figure*}

\begin{figure*}
\includegraphics[width=\linewidth]{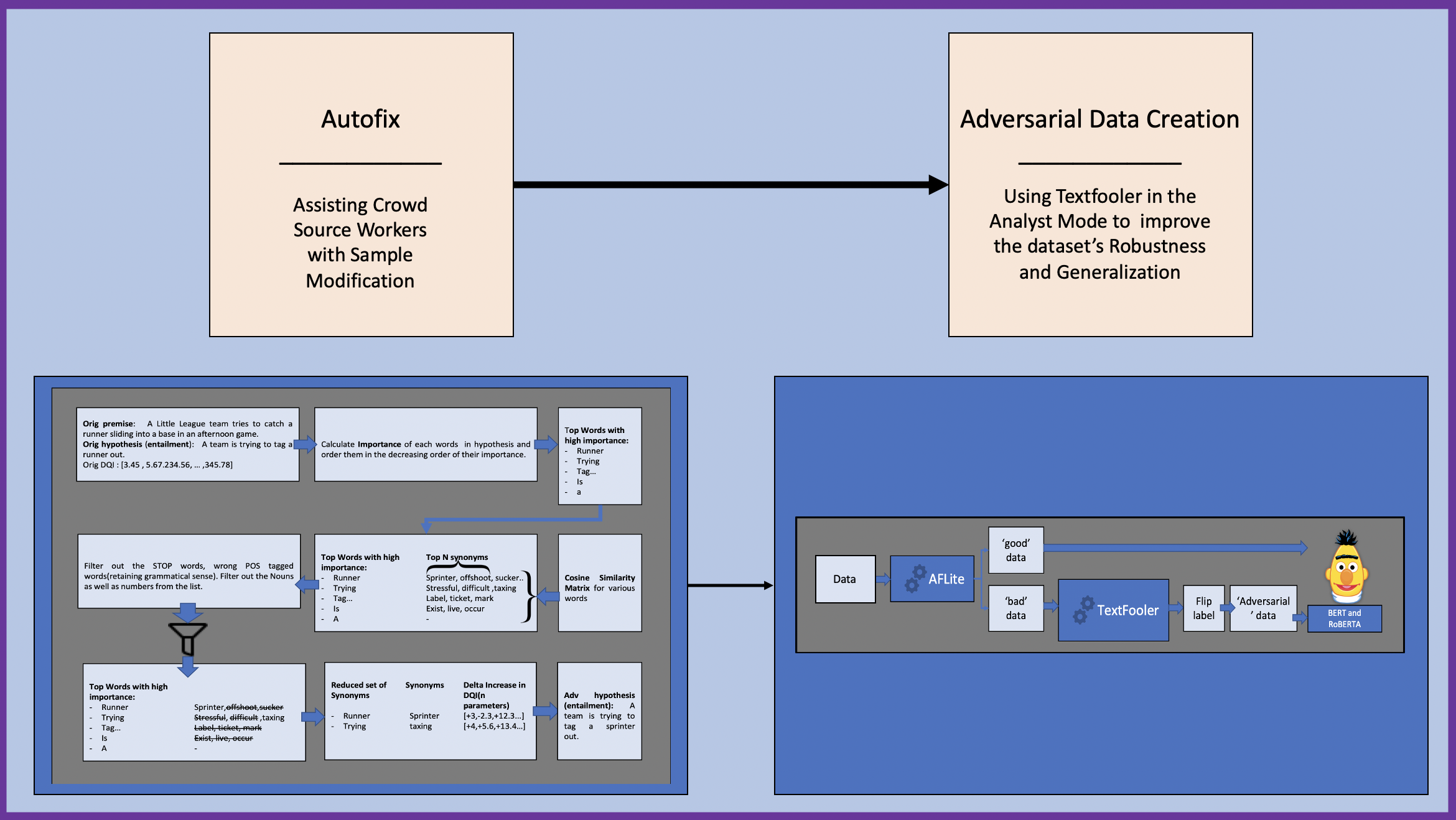}
  \caption{Step 4}
\label{fig:step4}
\end{figure*}

\begin{figure*}
\includegraphics[width=\linewidth]{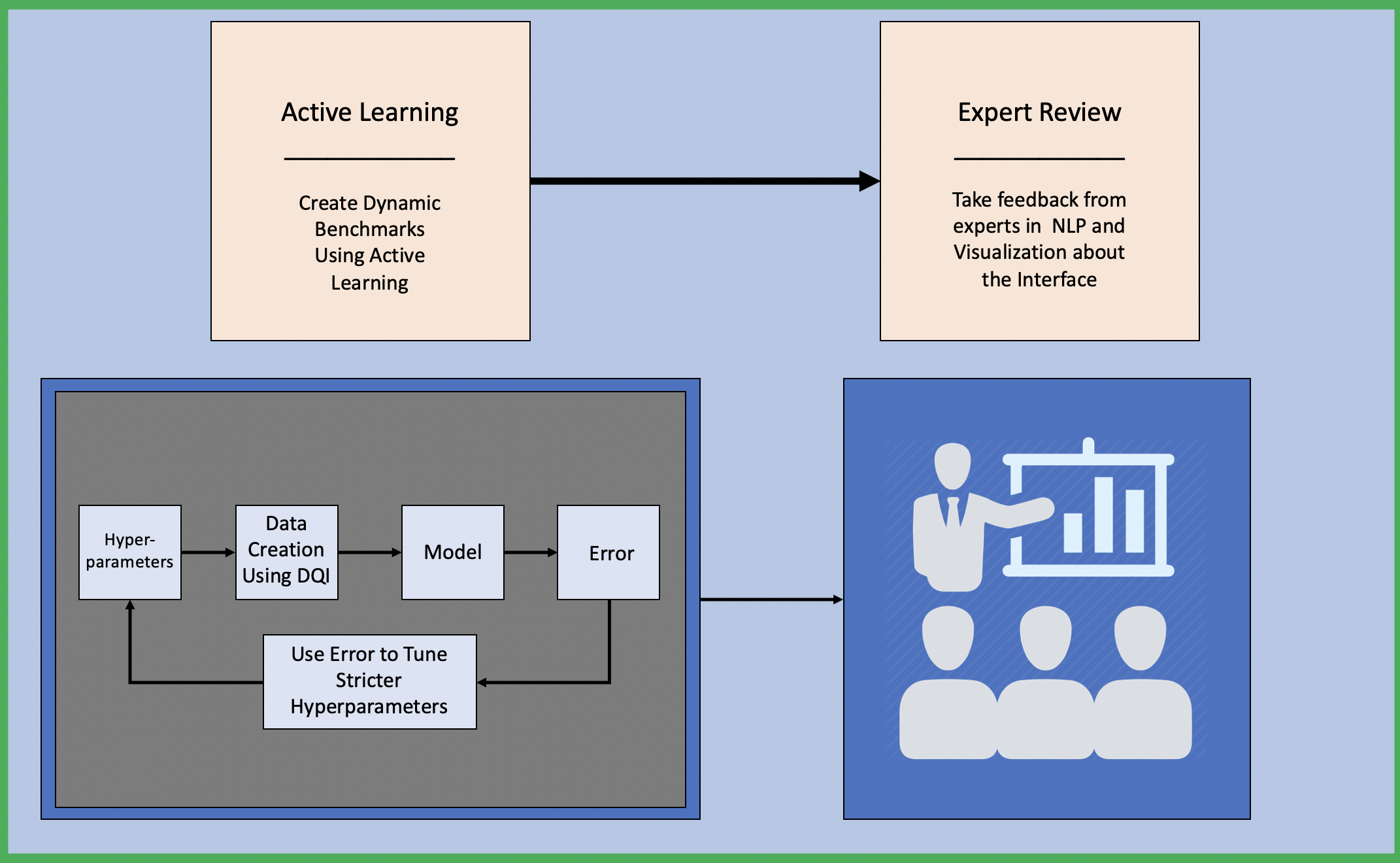}
  \caption{Step 5}
\label{fig:step5}
\end{figure*}

Figure \ref{fig:overall} summarizes our work in this paper using a process flow diagram. Figures \ref{fig:step1}, \ref{fig:step2}, \ref{fig:step3}, \ref{fig:step4} and \ref{fig:step5} provide more details on each of the processes.


\section{Universal DQI}
\label{sec:ttwwoo}
Our data creation paradigm is focused on showing (i) the overall data quality and (ii) the impact of new data created on the overall quality. To show impact, our setting involves the creation of the  $(n+1)^{th}$ data sample, when we already have $n$ data samples. In this paper, higher quality implies lower bias and higher generalization capability. 

We identify seven properties of text, which can represent several components covering the space of various possible interactions between samples in an NLP dataset. This is purely based on our intuition; for example, vocabulary distinguishes natural language from machine languages. Lesser amounts of vocabulary may therefore lead to misunderstanding and concurrently introduce biases. Similarly, if the frequency classes of n-grams are highly unbalanced, it may lead to models (i) ignoring or misunderstanding low frequency n-grams and (ii) memorizing and finding unintended correlations for high frequency n-grams from their surrounding contexts . We also have similar intuitions behind choosing properties like Semantic Textual Similarities (STS) and data splits. The seven properties are as follows:

\begin{itemize}
    \item Vocabulary
    \item Inter-sample N-gram Frequency and Relation
    \item Inter-sample STS 
    \item Intra-sample Word Similarity
    \item Intra-sample STS
    \item N-gram Frequency per Label
    \item Inter-spilt STS
\end{itemize}

% This is done by considering the influence of some particular aspect of already existing data samples on a new sample that is introduced. In turn, this reflects  our primary goal in eliminating data bias during the course of dataset creation rather than after creation. Therefore, the first four categories address the situation of a dataset creator (e.g.: crowd source worker) creating the $n^{th}$ sample, for an existing dataset of (n-1) samples. The fifth category provides a guided approach to benchmarking; this is done by ensuring that bias is removed over all possible train-test splits.

\paragraph{Hyper-parameters and Genericness of Universal DQI}
In various other domains such as water, food, and power we do have hyper-parameters in the quality indices. This is because of the dependence of a quality index on its application; for example, in the case of water quality, the quality of water needed for irrigation is different from the quality of water used for drinking, skin care, fitness, making medicine, and so on. Thus, the allowed limits of water components varies according to the use case. Similarly, we should have hyper-parameters in DQI, determining the tolerance of its components. These must be tuned for different NLP tasks and domains; for example, hyper-parameters for Biomedical NLP may be very different from those used for general NLP. However, we ensure Genericness of our proposed DQI by covering many types of datasets and a hierarchy of tasks ranging from NLI to Text Summarization in our process of developing the formula.

% We did survey, collected parameters and proposed empirical DQI with higher quality means higher generalization capability and less overfitting (so less presence of biases), the range of which needs to be tuned for different models and tasks depending on the objective. This is because, learning pattern of inductive bias is different for different models, thus can’t be generalized. Similarly DQI heavily depends on the objective at hand e.g. in medical domain, vocabularies may not be as overlapping or conflicting as general english, so tolerance level of data quality needs to stricter there. (explain 7 scenarios for which each of the term will be dominant). Also, task wise also, e.g. for NLI, QA, STS, RC, summarization, argumentation, accordingly explain why there can’t be a universal DQI. This is inline with other domain like food, water, power where again there is not a univesal equation, depending on application it changes, like for irrigation, for drinking, for skin care, for fitness, for weight loss etc. Here general thing is intelligence which is based on generalization. That sort of general formula did not exist there, but it must exist here from the point of estimation of AI performance.

\section{Potential leads}\label{plead}
% Give the spice example, instead of application, find from ingredients.

In this section, we comprehensively list potential leads that either (i) directly indicate bias, (ii) inspect the possible existence of bias via model probing, (iii) can be utilized to remove bias.  We consider a range of NLP tasks, in the following order: NLI, Argumentation, Question Answering, Reading Comprehension, and Abstractive Summarization. The ordering reflects the presence of increasing amounts of data per sample across tasks. We do this because bias analysis on lower order tasks can be extended to higher order tasks. This is reflected in Figures \ref{fig:LeadTable1}-\ref{fig:LeadTable3}. 
\paragraph{Justification of Task Ordering}
NLI takes a two sentence input (premise and hypothesis), to output a single label (entailment, neutral, and contradiction). Argumentation takes a four sentence input - claim, reason, warrant, and alternative warrant- and outputs the choice between the warrant and the alternative warrant. Multiple choice questions read either single/multi-line inputs and a set of choices comprising of words/sentences; they output a single choice (number/word/sentence). Open ended questions always output words/one or more sentences, after reading a multi-line input. Reading comprehension questions follow the same patterns as regular question answering samples, in that a multi-line input is read, and a choice/word/phrase/sentence is the output. The output format depends on the patterns of questions asked such as fill in the blanks and sentence completion. Also, the volume of input read is generally much larger than that seen in question answering.  Finally, abstractive summarization deals with both multi-line input and multi-line output.

\paragraph{Exploration} The list of potential leads has been compiled by reviewing literature discussing the impact, identification, isolation, and removal of bias in various datasets. We have extrapolated leads developed for a particular NLP task to a broader set of tasks along with examples \footnote{Refer to Appendix for more details}, such as the `copy' lead, originally used for abstractive summarization \cite{see-etal-2017-get}, split-and-rephrase \cite{gu-etal-2016-incorporating,aharoni-goldberg-2018-split}, and language modelling \cite{merity2016pointer}. Also, many of the leads do not directly signify bias. The papers they were compiled from have not directly mentioned them in relation to bias as well. We generate a lead by relating any model failures to potential bias (e.g.:multistep reasoning, coreference resolution). The leads are binned into seven categories as discussed in Section \ref{sec:ttwwoo}.

% \paragraph{NLI:}
% The datasets considered are classified into 3 groups, following the methodology of \cite{poliak2018hypothesis}, as Human Elicited, Human Judged and Recast.

\subsection{Vocabulary}
This bin deals with leads related to the vocabulary of a dataset. Specifically, the language used in the dataset in terms of its ambiguity and diversity is analyzed.

\label{sec:1}\paragraph{Vocabulary Magnitude:}\hyperref[sec:eg1]{(e.g.)}
We define this as the ratio of a dataset’s vocabulary size to the size of the dataset. The performance drop for MNLI is lesser than SNLI on providing partial input. This has been attributed to the presence of multiple genres in MNLI \cite{poliak2018hypothesis,gururangan2018annotation}.  This indicates that high vocabulary magnitude is desirable, and will reduce model dependency on spurious correlations. 
\paragraph{Vocabulary across POS Tags:}
The above lead also needs to be examined across POS tags to account for the presence of homonyms in vocabulary. The word distribution across samples might also be a good bias indicator.
\label{sec:5}\paragraph{Language Perturbation:}\hyperref[sec:eg5]{(e.g.)}
Correlations exploited by models can be exposed by isolating cases in which certain words or phrases are not used as a part of context in answering. Isolation can be achieved through the generation of examples by replacement of conjunctive \cite{talmor2019olmpics} phrases with meaningless filler words, and observing the extent of change in model accuracy with respect to the perturbed samples. If the learning curve of a model does not change when the input is perturbed or even deleted, then the model shows low language sensitivity. This can also be used to evaluate the influence of prepositional phrases.
\label{sec:6}\paragraph{Semantic Adverb Resolution:}\hyperref[sec:eg6]{(e.g.)}
The ability of models to correctly perceive  and differentiate the usage of adverbs such as ‘always’, ‘sometimes’, ‘often’, and ‘never’ reflects the extent of its reasoning capabilities \cite{talmor2019olmpics}. Therefore, the relationship between the model performance and level of presence of adverbs across samples is a viable lead.
\label{sec:7}\paragraph{Domain Specific Vocabulary:}\hyperref[sec:eg7]{(e.g.)}
Multiple genres dilute bias influence, as model performance decreases on data sets with multiple genres \cite{poliak2018hypothesis,gururangan2018annotation,glockner2018breaking}. In the process of creating multiple genre datasets, a large amount of domain specific vocabulary (e.g.: ordinals, nationalities, countries, etc.) is generated. Therefore the presence of an increased number of domain specific words seems desirable.

\subsection{Inter-sample N-gram Frequency and Relation}
This bin looks at leads that concern n-grams individually or in relation to other n-grams. Replacement based methods seem to provide a viable way to dilute the influence of these leads on bias. 

\label{sec:2}\paragraph{Maximal Word Distance:}\hyperref[sec:eg2]{(e.g.)}
The presence of multiple genres accounts for the robustness of the MNLI dataset in comparison to SNLI \cite{poliak2018hypothesis,gururangan2018annotation}. This can be quantified, in terms of spreading the ‘distances’ of words in the vocabulary to the maximum extent.
\label{sec:3}\paragraph{POS Tag Replacement:}\hyperref[sec:eg3]{(e.g.)}
POS tag replacement is a method to increase the vocabulary size in a controlled manner, as it allows for the balancing of a dataset's word distribution. Erasure, which can be used as an alternate elimination based method to balance word distribution, \cite{li2016understanding} was seen to sometimes generate semantically or grammatically incorrect sentences \cite{zhao2017generating}. In order to generate adversarial examples, Ribeiro et.al.\cite{ribeiro-etal-2018-local} replace sentence tokens by random words of the same POS tag, with a probability proportional to the similarity of their embeddings. Though there is less scope for generating grammatical errors using this method, there are cases where semantic inconsistencies are generated. To address this, we can combine POS tag replacement with the approach of discarding sentences with low resultant bigram frequencies as seen in the work of Glockner et.al. \cite{glockner2018breaking}. Textfooler uses a similar approach for replacement, in that the most important words for the target model are identified, and then replaced with the most semantically similar and grammatically correct words until the prediction is altered \cite{jin2019bert}. 
\label{sec:4}\paragraph{Consecutive Verb Frequency:}\hyperref[sec:eg4]{(e.g.)}
Machine translation results in dropping of consecutive verbs \cite{zhao2017generating}. We extrapolate this as a potential bigram related lead for NLI. 
\label{sec:8}\paragraph{Anonymization of Entities:}\hyperref[sec:eg8]{(e.g.)}
Masking entities across samples during processing will help ensure that the model does not rely on co-occurence based spurious biases in attaching a role to that entity. This is extrapolated from   Hermann et.al.\cite{hermann2015teaching}, originally used in the cloze style preparation of samples in RC datasets. This type of representation bias is also addressed by Li et.al. \cite{li2018resound}, in terms of object, scene and person bias.
\label{sec:12}\paragraph{Metonymy:}\hyperref[sec:eg12]{(e.g.)}
The usage of figures of speech in sentences must be resolved \cite{clark2018knowledge}, which requires effective context usage. It provides a case to examine model dependency on word association.
\label{sec:18}\paragraph{Stereotypes:}\hyperref[sec:eg18]{(e.g.)}
Rudinger et.al.\cite{rudinger-etal-2017-social} has shown that the hypotheses in NLI datasets contain gender, religious, race and age based stereotypes. This can be a form of contextual bias, in that the occurrence of sets of stereotype n-grams could bias the model towards a particular label. This also means that if exceptions to the stereotype were generated as adversarial examples, they would not be handled as similar pattern questions, but rather as contradictions.
\label{sec:21}\paragraph{Out of Distributions in Range}\hyperref[sec:eg21]{(e.g.)}
Models that rely on spurious correlations to solve the NLI task fail on out of sample distributions.  For example, ROBERTA can’t resolve numbers to be ages if they are not in a typical human range \cite{talmor2019olmpics}. 
\paragraph{Handling Conjunctions:}
Models can’t determine if conjunctional clauses are true, which is necessary in sorting, and comparison based reasoning inference chains \cite{talmor2019olmpics}.
\label{sec:22}\paragraph{Unnatural Language:}\hyperref[sec:eg22]{(e.g.)}
This refers to contradictory phrase pairs that arise by substituting adjectives and adverbs of opposing intent. For example, the usage patterns of ‘not’ and ‘very’ are identical in some cases, though the sentence meanings are opposite. Though not very common in occurrence, the resolution of such patterns between pairs and within pairs is necessary as it is indicative of negation \cite{talmor2019olmpics}.
\label{sec:49}\paragraph{Broad Referring Expressions:}\hyperref[sec:eg49]{(e.g.)}
The use of ‘broad’ referring expressions  like ‘the’, ‘this’, ‘that’, and ‘it’ in a test set distribution serves to test the ability of a model to reason based on any referential resolution patterns it has identified in the training set \cite{gundel1993cognitive,mcshane2016resolving,degen2020redundancy}.

\subsection{Inter-sample STS}
This bin deals with leads that can create and dilute bias as a consequence of a new sample's introduction in terms of sentence similarity. Syntactic, semantic, and pragmatic properties of sentences are considered. 

\label{sec:9}\paragraph{Sentence Structure:}\hyperref[sec:eg9]{(e.g.)}
Models learn to infer the meaning of each  class(parse) of sentences, and further extrapolate such parsing to more complex sentences. However, if the distribution of different parse structures is skewed, i.e., a small proportion of parse trees dominates the majority of the training samples, the resulting model may just learn spurious correlations, and thus perform poorly. This lead is created by extrapolating the works of \cite{poliak2018hypothesis}.
\label{sec:10}\paragraph{Multistep Reasoning:}\hyperref[sec:eg10]{(e.g.)}
Multistep reasoning is required to resolve complex sentences, by extrapolating the structures and semantics of simpler sentences. Failure to solve multistep reasoning samples might be an indicator of learning spurious correlations. This is evinced by two cases, namely compositional and numerical reasoning samples. Both follow a chain of inferences, with numerical reasoning additionally quantifying and solving arithmetic questions. Language models have been seen to struggle to resolve compositional questions even with supervision \cite{talmor2019olmpics}. Accurate numerical reasoning resolution has also been a deficiency  in inference models \cite{naik2018stress}.
\label{sec:11}\paragraph{Inter-Sentence Antithesis:}\hyperref[sec:eg11]{(e.g.)}
A special case of pattern exploitation in language modelling  is in converse examples, wherein two samples have identical linguistic patterns, and only differ with a single word or phrase of opposing meaning \cite{naik-etal-2018-stress}. Incorrect resolution of this case might suggest a model's dependency on annotation artifacts.
\label{sec:14}\paragraph{Sentence Length Variation:} \hyperref[sec:eg14]{(e.g.)}
Sentence length should vary across samples to ensure that models don't use it as an annotation artifact. \cite{gururangan2018annotation}.
\label{sec:15}\paragraph{Start Tokens:}\hyperref[sec:eg15]{(e.g.)}
The presence of repeated start tokens in the premise and hypothesis, could bias a model to only focus on certain parts of the input. This is extrapolated from the work of Sugarawa et.al. \cite{sugawara2018makes}. 
\label{sec:16}\paragraph{Ellipsis Resolution:}\hyperref[sec:eg16]{(e.g.)}
The presence of ellipsis in samples has been a point of shortfall for language models \cite{clark2018knowledge}, due to their reliance on factitious relations in NLI datasets.

\subsection{Intra-sample Word Similarity}
This bin concerns intra-sample bias, in the form of word similarities. Specifically, bias seen within the premise and/or within the hypothesis statements of a sample is dealt with.

\label{sec:13}\paragraph{Presupposition and Query:}\hyperref[sec:eg13]{(e.g.)}
Sometimes, sentences indicate an already implied fact, which is utilized as the basis for a further query on a specific attribute/case of that fact within a hypothesis \cite{clark2018knowledge}. This can indicate a model's ability to resolve context.
\label{sec:17}\paragraph{Coreference Resolution:}\hyperref[sec:eg17]{(e.g.)}
Coreferences can be a result of the usage of pronouns, as well as abstractive words like ‘each’ and ‘some’ \cite{gururangan2018annotation,cirik2018visual}. This coreference may occur in both the premise and hypothesis or in either one, with an actual entity stated in the respective other. The inability to correctly resolve coreferences suggests the misuse of or disregarding of context, due to dependence on biases.
\label{sec:19}\paragraph{Taxonomy Trees:}\hyperref[sec:eg19]{(e.g.)}
Consider the conjunction of two objects that can be grouped under a generic super-class. The first object’s closest parent on the taxonomy tree is taken as the superset across both objects. This applies even if the second object does not fall into that superset. For example, ‘horse and crow’ would be grouped as ‘animal’, but ‘crow and horse’ may be grouped as ‘bird’ in some cases \cite{talmor2019olmpics}. 

\subsection{Intra-sample STS}
This bin is concerned with another aspect of intra-sample bias, i.e., that which is seen between the premise and hypothesis statements. 

\label{sec:20}\paragraph{Overlap:}\hyperref[sec:eg20]{(e.g.)}
Overlap in terms of words seen in the premise-hypothesis pair could be indicative of label. Failure to resolve antonymy and negation is a special case of this \cite{naik2018stress}. This feature is used as a bias indicator in the construction of the adversarial dataset HANS, in three ways: (i) assuming that a premise entails all hypotheses constructed from words in the premise , (ii)assuming that a premise entails all of its contiguous subsequences, and (iii) assuming that a premise entails all complete subtrees in its parse tree \cite{mccoy2019right}. 
\label{sec:24}\paragraph{Sentence Similarity:}\hyperref[sec:eg24]{(e.g.)}
Studies have shown that high sentence similarity biases systems towards assigning the label of ‘entailment’, and low similarity towards ‘neutral’ \cite{naik2018stress}. This is dependent on word overlap levels between the sentences \cite{clark2018knowledge}. 
\subsection{N-gram Frequency per Label}
This bin contains leads that reflect the dominating causes of bias introduced due to the influence of existing labels on the new sample's label. Leads are shortlisted in terms of bias originating from (i) premise, (ii) hypothesis, and (iii)both.

\label{sec:55}\paragraph{Erasure:}\hyperref[sec:eg55]{(e.g.)}
Li et.al. \cite{li2016understanding} erase different levels of representation used by models, and use reinforcement learning to erase minimal sets of input words to flip model decisions. This technique can indirectly help identify certain elements producing annotation artifacts by extrapolating the minimal set of input words responsive to models.
\label{sec:25}\paragraph{Negation:}\hyperref[sec:eg25]{(e.g.)}
Terms such as ‘no’ or ‘not’ are indicators of universal negation, and containing samples are predisposed to be labeled as contradiction in SNLI \cite{poliak2018hypothesis}.
\label{sec:26}\paragraph{Antonymy:} \hyperref[sec:eg26]{(e.g.)}
Discarding antonymy due to the absence of explicit negation is an indication of model bias \cite{naik2018stress}. 
\label{sec:27}\paragraph{WL Mapping:}\hyperref[sec:eg27]{(e.g.)}
This lead is a measure of the level of correlation within a class label. $P(l/w)$ gives the conditional probability of the occurrence of a label(l) given a word(w). If it has value 0 or 1, the label becomes trivial \cite{poliak2018hypothesis}. Such a skew leads to inference on the basis of word presence, a spurious bias.
\label{sec:28}\paragraph{PL Mapping:}\hyperref[sec:eg28]{(e.g.)}
Pattern exploitation can be extended to phrase level dependencies of labels, measured as $P(l/p)$, i.e. P(label/phrase). 
\label{sec:29}\paragraph{Vocabulary Score:}\hyperref[sec:eg29]{(e.g.)}
We define this lead as a constant length vector of: (i) the number of labels a given word is present in, (ii) the individual counts of the word in each label. This will help prevent the skew of labels given a particular word; for example, the word ’sleep’ and its variations were found to be indicators of contradiction in SNLI, as they were predominantly present in samples with that label \cite{poliak2018hypothesis}.
\label{sec:30}\paragraph{Overlap Rate:}\hyperref[sec:eg30]{(e.g.)}
This is a measure in the work of Dasgupta et.al.\cite{dasgupta2018evaluating}, which measures the bias of a model towards entailment or neutral by calculating the number of overlap words divided by the number of words in a sample.
\label{sec:31}\paragraph{Copying:}\hyperref[sec:eg31]{(e.g.)}
Copy augmented modeling has proven useful in works on the split and rephrase task \cite{aharoni-goldberg-2018-split,gu-etal-2016-incorporating}. The mechanism has also been used by See et.al. \cite{see-etal-2017-get} for abstractive summarization, and by Merity et.al. \cite{merity2016pointer} for language modelling. We propose the use of an iterative copy mechanism, to copy different n-grams of words between the premise and hypothesis statements. By noting the points at which the label changes, we can isolate the most informative word overlap sets.
\label{sec:32}\paragraph{Hypothesis Only Prediction:}\hyperref[sec:eg32]{(e.g.)}
This lead is used to test dependencies between the label and hypothesis, to prevent partial answering based on correlation \cite{tan2019investigating}.
\label{sec:33}\paragraph{Cue Influence:}\hyperref[sec:eg33]{(e.g.)}
Niven et.al. \cite{niven2019probing} address the presence and nature of artifacts, and their contribution to Warrant only predictions in the ARCT dataset. They evaluate this using three metrics: applicability, productivity, and coverage. This can be extrapolated to finding the influence of cues on hypothesis only prediction in NLI.
\label{sec:34}\paragraph{Length Mismatch:}\hyperref[sec:eg34]{(e.g.)}
The length of a sentence can indicate its label class, as entailment or neutral for shorter and longer sentences respectively. Additionally, length mismatches between the premise and hypothesis can predispose the model to predict non-entailment labels \cite{poliak2018hypothesis,gururangan2018annotation,naik-etal-2018-stress}.
\label{sec:35}\paragraph{Grammaticality:}\hyperref[sec:eg35]{(e.g.)}
Tests on the FN+ dataset have shown that sentences with poor grammar are classified under non-entailment labels \cite{poliak2018hypothesis}.
\label{sec:36}\paragraph{PMI:}\hyperref[sec:eg36]{(e.g.)}
PMI represents a scaled conditional probability of word-label dependency. It measures how likely they are to co-occur, given their independent probabilities, and joint probability under a state of conditional independence \cite{naik2019exploring,gururangan2018annotation}.
\label{sec:37}\paragraph{Scripts:}\hyperref[sec:eg37]{(e.g.)}
A way to break down complex inference chains is to identify common scripts \cite{clark2018knowledge} based on the incorporation of real world knowledge . For example, ‘X wants power and therefore tries to acquire it, Y doesn’t want X to have power and tries to thwart X’ is a common script for inference chains.
\label{sec:38}\paragraph{Numerical Reasoning:}\hyperref[sec:eg38]{(e.g.)}
The accurate quantification of numbers is essential to correct label prediction. Language models often fail at numerical reasoning \cite{naik2018stress}. Additionally, the presence of numbers predisposes bias against entailment, as entailment examples in SNLI are seen to have numerical information  abstracted with words like ‘some’ or ‘few’ \cite{gururangan2018annotation}.
\label{sec:39}\paragraph{Gender:}\hyperref[sec:eg39]{(e.g.)}
The absence of gender information is an indicator of entailment in SNLI \cite{gururangan2018annotation}.
\label{sec:40}\paragraph{Hypernyms and Hyponyms:}\hyperref[sec:eg40]{(e.g.)}
Models follow a super-set/sub-set structured approach, in the form of hypernyms and hyponyms \cite{richardson2019does}, when assigning entailment. Glockner et.al.\cite{glockner2018breaking} generate entailment samples by replacing words with their synonyms, hyponyms and hypernyms. Contradiction samples are generated by replacing words with mutually exclusive co-hyponyms and antonyms. Co-hyponym resolution is an issue for biased NLI models. Therefore, the above methods of sample generation produce adversarial samples . Models using DIRT \cite{lin2001dirt} based methods suffer from the problem of forming prototypical hypernyms as spurious biases while solving. For example, a chair might serve as a super-set for its legs, even though it is not a true hypernym \cite{levy2015supervised}.
\label{sec:41}\paragraph{Modifiers and Superlatives:}\hyperref[sec:eg41]{(e.g.)}
The use of modifiers such as 'tall' and 'sad', and superlatives like ‘first’ and ‘most’ is predominantly seen in the neutral class \cite{gururangan2018annotation}.
\label{sec:42}\paragraph{Causal Phrases:}\hyperref[sec:eg42]{(e.g.)}
Phrases like ‘because of’ and ‘due to’ are associated with the neutral class, as they add specificity \cite{gururangan2018annotation}.
\label{sec:43}\paragraph{Absence Indicators:}\hyperref[sec:eg43]{(e.g.)}
Words like ‘sleep’ or ‘naked’ indicate the absence of an object in the sentence, and therefore are associated primarily with the contradiction class \cite{gururangan2018annotation}.
\label{sec:44}\paragraph{Ambiguity:}\hyperref[sec:eg44]{(e.g.)}
Cases where external knowledge or chain reasoning is required to solve referential cues are classified as neutral \cite{naik2018stress}.
\label{sec:45}\paragraph{Bigram Entropy:}\hyperref[sec:eg45]{(e.g.)}
High entropy bigrams can be used as indicators of entailment and neutral labels. Here entropy is calculated as  \cite{tan2019investigating}: 
% H(c | w) = −\sum_{c}^{}p(c | w)\log p(c | w)
This can be extended to phrases as well, extrapolating on the forms of representation bias discussed by Li et.al. \cite{li2018resound}, in the form of object, scene, and person bias.
\label{sec:46}\paragraph{Paraphrasing:}\hyperref[sec:eg46]{(e.g.)}
Paraphrased question generation is often used to generate additional samples \cite{sugawara2018makes}. PAWS is an adversarial dataset for paraphrase identification. It employs word swapping and back translation to generate challenging paraphrase pairs \cite{zhang2019paws}. However, the limit of paraphrasing is an important lead to be considered, i.e., at what point does the semantic meaning change? An example of this is the inability of a model to distinguish between the meanings of ‘same’ and ‘about the same’ \cite{clark2018knowledge}. \label{sec:47}\paragraph{Multiple Cases:}\hyperref[sec:eg47]{(e.g.)}
This lead is extrapolated from Sugawara et.al. \cite{sugawara2018makes}. It deals with possible ambiguity in answer choice selection. This occurs when there are multiple span matches among answer choices to the passage span selected by the question. In the context of NLI, this can be viewed as an indicator for neutral and non-neutral label assignment.
\label{sec:48}\paragraph{Modality and Belief:}\hyperref[sec:eg48]{(e.g.)}
Modality details how things could, must, or could not have been. Belief is viewed as a true/false construct when deciding if a modality holds for NLI. This is reflected in patterns followed by human annotators, as seen in Bowman et.al., Williams et.al. \cite{bowman2015large,williams2017broad}. 
\label{sec:51}\paragraph{Shuffling Premises:}\hyperref[sec:eg51]{(e.g.)}
Shuffling of premises in the test set and checking model performance can help to understand the influence of premises in deciding label \cite{tan2019investigating}. 
\label{sec:52}\paragraph{Concatenative Adversaries:}\hyperref[sec:eg52]{(e.g.)}
The addition of distracting phrases added in conjunction with premise hypothesis pairs might help test the model’s reliance on spurious biases  \cite{naik2018stress,jia2017adversarial}.
\label{sec:56}\paragraph{Crowdsource Setting:}\hyperref[sec:eg56]{(e.g.)}
Analysis of the story cloze task \cite{mostafazadeh-etal-2016-corpus} shows that there is a difference in the writing styles employed by annotators in different sub-tasks \cite{schwartz2017effect}. Following the order of composing a full story, or a one line coherent / incoherent ending, the following patterns are observed: (i) decrease in sentence length, (ii) fewer pronouns, (iii) decrease in use of coordinations like 'and', (iv) less enthusiastic and increasingly negative language. These are also found to be indicators of deceptive text, by Qin et.al. \cite{qin2004exploratory}. Their work categorizes deceptive text on the basis of nineteen parameters, classified into five categories: quantity, vocabulary complexity, sentence complexity, specificity and expressiveness, and informality. Yancheva et.al. \cite{yancheva2013automatic} include the mean number of clauses per utterance and the Stajner- Mitkov measure of complexity as highly informative syntactic features for deception in text. Liars tend to use fewer self-references, more negative emotion words, and fewer markers of cognitive complexity, i.e., fewer 'exclusive' words, and more 'motion' verbs like walk and go \cite{newman2003lying}. These features can all be applied as leads in NLI as they provide spurious biases for distinguishing both contradiction labels, as well as annotator patterns.
\label{sec:57}\paragraph{Sample Perturbation:}\hyperref[sec:eg57]{(e.g.)}
Kaushik et.al. \cite{kaushik2019learning} use a human-in-the-loop system to create counterfactual samples for a dataset. When a model is trained on these samples, it fails on the original data, and vice versa. Augmenting the revised samples however, reduces the correlations formed from the two sets individually. Gardner et.al. \cite{gardner2020evaluating} create contrast sets by perturbing samples to change the gold label, to view a model's decision boundary around a local instance. Model performance on contrast sets decreases, thus creating new benchmarks.

\subsection{Inter-split STS}
This bin talks about the necessity of optimal dissimilarity between training and test sets. All the leads of the previous groups must be optimized within each spilt as well. 

\label{sec:53}\paragraph{Variation of Split:}\hyperref[sec:eg53]{(e.g.)}
Recent studies have shown that benchmarking is done improperly, due to the presence of fixed training and test sets \cite{tan2019investigating}. Also, evaluation metrics are mistakenly treated as exact quantities. They should instead be treated as estimates of random variables corresponding to true system performance. Therefore, many works either do not use proper statistical tests- such as hypothesis testing- for system comparison/ do not report which tests were used.  The absence of proper testing can result in type 1 errors \cite{gorman-bedrick-2019-need}.
\paragraph{Annotator Bias:}
Geva et.al. \cite{geva2019we} show that model performance improves when annotator identifiers are included as training features. Models are also not able to generalize to test samples created by annotators if those annotators did not contribute at all to the training set. This leads to the model seemingly fitting the annotators and not the task. To mitigate this bias, Geva et.al. propose that annotator sets be made disjoint for train and test sample generation.
\paragraph{World Definition:}
The negative set of a dataset defines what the dataset considers to be “the rest of the world”. If that set is not representative, or unbalanced, it could produce classifiers that are overconfident and not discriminative \cite{torralba2011unbiased}. 

\subsection{Miscellaneous}
This bin houses a few cases of leads which deal with bias originating from model interaction, human evaluation, and gold-label determination. These cannot be sorted into the previous categories defined, as (i) we are focusing on model-independent development, (ii) we are not considering any flaws in gold-label assignment to data, and (iii) we are only concerned with the data creation phase, and not the data validation phase.

\label{sec:54}\paragraph{Innoculation Cost:}\hyperref[sec:eg54]{(e.g.)}
This is used in the context of question answering, by Richardson et.al. \cite{richardson2019does}, and is defined as the improvement in performance seen after the innoculation of a language model. In innoculation, training is done on new tasks using small sample sets. This aims at fine tuning the model to perform robustly on out of distribution samples without re-purposing the model entirely. This data could be solved using available knowledge in the model.  A similar approach is also seen in Nie et.al. \cite{nie2019adversarial}, who use an adversarial human-and-model-in-the-loop procedure, to generate a new adversarial dataset, on which a model is trained to improve its performance. However, both these approaches might introduce their own set of biases. 
\label{sec:61}\paragraph{Disagreement:}\hyperref[sec:eg61]{(e.g.)}
If disagreement amongst annotators looks like random noise, then data with low reliability can be tolerated by a machine learning model. If this disagreement contains patterns, then a model can use these patterns as a spurious bias, to boost its performance. By testing for correlation between two annotators, some of these patterns can be identified. However, not all patterns picked up by the model  will necessarily show up on the correlation test- a scenario which could arise if the number of samples with disagreement is too low \cite{reidsma2008reliability}.
\paragraph{Random Labelling:}
Zhang et.al. \cite{zhang2016understanding} train models on datasets where the true labels are replaced by random labels. It is seen that models can achieve zero training error, even on the randomly labelled data. Therefore, without changing the model, model size, hyper parameters, and optimizer, the generalization error of a model can be forced to increase considerably. Explicit regularization techniques like weight decay, dropout, and data augmentation are also found to be insufficient for controlling generalization error. Stochastic gradient descent with unchanged hyper parameter settings can optimize weights to fit to random labels perfectly, even though the true meaning of the labels is lost. They conclude that optimization is easy even if the resulting model does not generalize. So the reasons for optimization being easy differs from the true cause of generalization.
\paragraph{Re-Optimizing Weights:}
REPAIR formulates bias minimization as an optimization problem, by redistributing weights to penalize easy examples for a classifier. By maximizing the ratio between loss on the re-weighted dataset and the uncertainty of ground truth labels, the bias is reduced \cite{li2019repair}.
\paragraph{Ranking Artifacts:}
We propose that annotation artifacts \cite{gururangan2018annotation} as well as some other leads be ranked based the extent of their influence on label. Using this ranking, the artifact combinations and occurrences that give rise to a greater amount of bias can be isolated.
\paragraph{Human Performance Measurement:}
Gardner et.al. \cite{gardner2020evaluating} measure human performance on the contrast sets they create, by evaluating themselves on the contrast sets. The authors know the intricacies of the dataset creation process and the motives behind creating the dataset. Therefore, author evaluation can bias the reporting of human performance levels.
\paragraph{Order of Input:}
Dodge et.al. \cite{dodge2020fine} study how the different orders in which training data is fed to the model affect the achieved validation performance of the model. This evidences that some data orderings serve as better random seeds than others. These orderings are particular to a dataset. This ordering can be linked to the influence of dataset bias.
\paragraph{Models of Annotation:}
Paun et. al. \cite{paun2018comparing} have analyzed several models of annotation to improvise the  traditional way of calculating and handling gold standard labels, annotator accuracies and bias minimization, and item difficulties and error patterns. Bayesian models of annotation have been shown to be better than traditional approaches of majority voting and coefficients of agreement.
\paragraph{Exposure Bias:}
A model's way of handling data may introduce bias. For example, exposure bias is introduced because of the difference in exposing data to the model during training and inference phase \cite{caccia2018language}.

%The way models deal with data may introduce bias
% \begin{table*} 
% \centering
%         \scriptsize
%         \resizebox{2.0\columnwidth}{!}{
% \begin{tabular}{|l|l|l|l|l|}
% \hline
%  &
%   Vocabulary &
%   Inter-sample N-gram Frequency and Relation &
%   Inter-sample STS &
%   Intra-sample Word Similarity \\ \hline
% Considered Leads &
%   \begin{tabular}[c]{@{}l@{}}Vocabulary Magnitude,\\ Vocabulary across POS Tags, \\ Domain SpecificVocabulary\end{tabular} &
%   \begin{tabular}[c]{@{}l@{}}Maximal Word Distance,\\ POS Tag Replacement,\\ Stereotypes,\\ Out of Distributions in Range\end{tabular} &
%   \begin{tabular}[c]{@{}l@{}}Sentence Structure, \\ Sentence Length Variation\end{tabular} &
%   \\ \hline
% Unconsidered Leads &
%   \begin{tabular}[c]{@{}l@{}}Language Perturbation,\\ Semantic Adverb Resolution\end{tabular} &
%   \begin{tabular}[c]{@{}l@{}}Consecutive Verb Frequency,\\  Anonymization of Entities,\\  Metonymy,\\  Handling Conjunctions,\\  Unnatural Language, \\ Broad Referring Expressions\end{tabular} &
%   \begin{tabular}[c]{@{}l@{}}Multistep Reasoning, \\ Inter-Sentence Antithesis, \\ Start Tokens, \\ Ellipsis Resolution\end{tabular} &
%   \begin{tabular}[c]{@{}l@{}}Presupposition and Query, \\ Coreference Resolution, \\ Taxonomy Trees\end{tabular} \\ \hline
% \end{tabular}
% }
% \caption{Lead Categorization 1}
% \label{lcategory1}
% \end{table*}
\begin{table*}
\centering
        \scriptsize
        \resizebox{2.0\columnwidth}{!}{
\begin{tabular}{lllll}
\hline
 &
  \textbf{Vocabulary} &
  \textbf{Inter-sample N-gram Frequency andRelation} &
  \textbf{Inter-sample STS} &
  \textbf{Intra-sample Word Similarity} \\ \hline
\textbf{Considered Leads} &
  \begin{tabular}[c]{@{}l@{}}Vocabulary Magnitude,\\ Vocabulary across POS Tags, \\ Domain SpecificVocabulary\end{tabular} &
  \begin{tabular}[c]{@{}l@{}}Maximal Word Distance,\\ POS Tag Replacement,\\ Stereotypes,\\ Out of Distributions in Range\end{tabular} &
  \begin{tabular}[c]{@{}l@{}}Sentence Structure, \\ Sentence Length Variation\end{tabular} &
  \\ \hline
\textbf{Unconsidered Leads} &
  \begin{tabular}[c]{@{}l@{}}Language Perturbation,\\ Semantic Adverb Resolution\end{tabular} &
  \begin{tabular}[c]{@{}l@{}}Consecutive Verb Frequency,\\  Anonymization of Entities,\\  Metonymy,\\  Handling Conjunctions,\\  Unnatural Language, \\ Broad Referring Expressions\end{tabular} &
  \begin{tabular}[c]{@{}l@{}}Multistep Reasoning, \\ Inter-Sentence Antithesis, \\ Start Tokens, \\ Ellipsis Resolution\end{tabular} &
  \begin{tabular}[c]{@{}l@{}}Presupposition and Query, \\ Coreference Resolution, \\ Taxonomy Trees\end{tabular} \\ \hline
\end{tabular}
}
\caption{Lead Categorization 1}
\label{lcategory1}

\vspace{1cm}
\centering
        \scriptsize
        \resizebox{2.0\columnwidth}{!}{
\begin{tabular}{lllll}
\hline
 &
  \textbf{Intra-sample STS} &
  \textbf{N-gram Frequency per Label} &
  \textbf{Inter-split STS} &
  \textbf{Miscellaneous} \\ \hline
\textbf{Considered Leads} &
  \begin{tabular}[c]{@{}l@{}}Overlap, \\ Sentence Similarity\end{tabular} &
  \begin{tabular}[c]{@{}l@{}}Erasure, \\ Negation, \\ WL Mapping, \\ PL Mapping, \\ Vocabulary Score, \\ Overlap Rate, \\ Hypothesis Only Prediction, \\ Cue Influence, \\ Length Mismatch, \\ Grammaticality, \\ PMI, \\ Gender, \\ Modifiers and Superlatives, \\ Causal Phrases, \\ Absence Indicators, \\ Bigram Entropy\end{tabular} &
  \begin{tabular}[c]{@{}l@{}}Variation of Split, \\ Annotator Bias\end{tabular} &
  \begin{tabular}[c]{@{}l@{}}Ranking Artifacts, \\ Human Performance Measurement, \\ Models of Annotation\end{tabular} \\ \hline
\textbf{Unconsidered Leads} &
  &
  \begin{tabular}[c]{@{}l@{}}Antonymy, \\ Copying, \\ Scripts, \\ Numerical Reasoning, \\ Hypernyms and Hyponyms, \\ Ambiguity, \\ Paraphrasing, \\ Multiple Cases, \\ Modality and Belief, \\ Shuffling Premises, \\ Concatenative Adversaries, \\ Crowdsource Setting, \\ Sample Perturbation,\end{tabular} &
  World Definition &
  \begin{tabular}[c]{@{}l@{}}Innoculation Cost, \\ Disagreement, \\ Random Labelling, \\ Re-Optimizing Weights, \\ Order of Input, \\ Exposure Bias\end{tabular} \\ \hline
\end{tabular}
}
\caption{Lead Categorization 2}
\label{lcategory2}
\end{table*}
\section{Identification of Generic Leads:}
% We categorize potential leads into three parts (i) Generic Leads (ii)
We find that certain leads are specific to models. They help in probing models and analyzing bias better, and thus can be used as guidelines in creating bias-minimized data or tools to visualize the bias exploitation process in models. However, these have to be updated every time we have a new SOTA model. So, we don't include them in our development of generic DQI. Table \ref{lcategory1} and \ref{lcategory2} enlists filtered leads across categories. We use leads to extend our intuition, but don't rely on them completely. For example, we don't consider any leads for the Intra-sample Word Similarity Category.

% See there are certain things which are purely our intuition or based on commonsense to fool DQI. This were not found in potential leads covered in literature

\paragraph{Scope for Model-specific\footnote{SOTA Model such as ROBERTA} DQI using Active Learning:} We include model-specific leads since they can be utilized as constraints to develop model-specific DQI which can be further utilized in creating hard datasets or understanding bias in models. For example, Semantic Adverbs should be present a minimum number of times in a dataset. The same is true for Domain Specific Words as they force models to learn and not look for patterns. Similarly, Consecutive Verb Frequency should have a minimum threshold for certain verbs. Also there should be sufficient number of figures of speech. The idea is there should be a minimum number of patterns which are difficult for the SOTA model to crack while solving a dataset. This is to force models to not rely on spurious biases in order to solve that dataset. Our proposed workflow of data creation paradigm can be used to prepare such datasets by just extending our DQI to model specific DQIs. Active Learning can be used to make the dataset hard using errors that a model make to retune hyper-parameters in DQI. The use of DQI in the active learning process helps partially automate the feedback process, and reduces the load on crowd workers. Human bias also gets minimized using constraints based on DQI in our data creation paradigm. However, we limit this paper to generic DQI.

\section{Empirical DQI}
We utilize generic leads to expand our intuition described in Section \ref{sec:ttwwoo} and propose the formula for Empirical DQI. We ensure that there is at least one term representing each category in the overall DQI. We enlist DQI component terms $DQI_C$ representing each of the categories.

\paragraph{Vocabulary:} We define Average Vocabulary as the number of unique words per total number of data samples in the dataset. Higher the average vocabulary, higher the quality of data. Sentence length also should be within an upper and lower limit, as shorter and longer sentences have a propensity to introduce artifacts. There are 2 hyper-parameters $a$ and $b$ representing lower and higher thresholds of sentence length. Also, the frequency distribution of sentence length should have higher variance to prevent the model from over fitting towards a specific length. Let $X$ represent a dataset, $v$ be the vocabulary, $s$ be sentence length, $S$ represent the set of all sentences in the dataset and $size$ represent the total number of samples.\\\\
% \begin{equation}
$DQI_{c1}=\frac{v(X)}{size(X)}+\sigma(s(X))*\frac{\sum_{S}\sign((s-a)(b-s))}{size(S)}$
% \end{equation}
\paragraph{Inter-sample N-gram Frequency and Relation:} Lesser the variance of frequency of words, higher the data quality. This also holds for each category of POS tags. We normalize individual frequencies by dividing with size. Every word should have a minimum frequency, so that models get the necessary favorable bias. There should also be a upper bound so that models do not get a chance to use highly frequent words as bias. The frequency distribution of bigrams, trigrams, and full sentences should not be skewed. They should have lower variance to have higher quality. Again, each of these should have a minimum and maximum frequency value. Let $i \epsilon \{Words, Verbs, Adjectives, Nouns, Adverbs,$
$Bigram, Trigram, Sentences\}$ and $\nu$ represent frequency. Minimum and maximum threshold, defined similarly to the thresholds of the first component, are represented as $c$ and $d$.\\\\
% \resizebox{\hsize}
% \begin{equation}
$DQI_{c2}=\sum_{i}(\frac{1}{\sigma (\frac{i(\nu)}{size(i)})}*\frac{\sum_{i}((\nu_{i}-c)(d-\nu_{i}))}{size(i)})$
% \end{equation}
%\sum_{l}\mathbf{rem}{\frac{F_k}{f_{x_l}}}
%numerator can be size(X)
% $\abs{Sim_{lm}-SIM}$
\paragraph{Inter-sample STS: } Every sentence should have another sentence in the dataset which has some minimum similarity score, and there should be some minimum number of such similar sentences. However, the distribution should have lower variance for ensuring higher quality. Semantic Textual Similarity (STS), paraphrasing or identification of duplicates are the options to implement this. Here, $l$ spans the dataset, $Sim_{lm}$ stands for sentence similarity between the $l^{th}$ sentence and $m^{th}$ sentence where $m$ spans every other sentence in the dataset, $e$ is a hyperparmeter dependent on the dataset size which says how many sentences should have the minimum simialrity score. $SIM$ represents the minimum similarity value which is a hyperparameter, and $\max_{me}$ stands for $e$ number of maximum values.\\\\
%Computationally complex, so we might have to convert to simpler formulae.
% \begin{equation}
$DQI_{c3}=\frac{size(S)}{\sigma(\forall_{l} \nu_{\sign\frac{\abs{Sim_{lm}-SIM}-(Sim_{lm}-SIM)}{2}})+1}+\frac {2*size(S)}{(\sum_{l}\sum_{e}{\max\limits_{me}\frac{}{}(\abs{Sim_{lm}-SIM}}-(Sim_{lm}-SIM)))+1}$
% \end{equation}
% $DQI_{c3}=\frac{size(S)}{\sigma(\forall_{l} \nu_{\sign\frac{\left|{Sim_{lm}-SIM}\right|-(Sim_{lm}-SIM)}{2}})+1}+\frac {2*size(S)}{(\sum_{l}\sum_{e}{\max\limits_{me}\frac{}{}(\left|{Sim_{lm}-SIM}}\right|-(Sim_{lm}-SIM)))+1}$

% $DQI_{c3}=\frac{1}{Var(\sum_{m}(f_{(Sim_{lm}-SIM_l)-\left|{Sim_{lm}-SIM_l}}\right|}))}+\frac{\sum_{l}{\max_{ma}(Sim_{lm}-SIM_l)-\left|{Sim_{lm}-SIM_l}}\right|}{size(X)}$
%look to add the same for words also (not mandatory though as sentence covers most part of it)
\paragraph{Intra-sample Word Similarity: } Summation of similarity of a word to every other word in the sentence should have a minimum value. The closer the average similarity score is towards the minimum value, the higher is the data quality. Here, $WSim_{lm}$ stands for word similarity between the  $l^{th}$ word and the $m^{th}$ word where $m$ spans every word in the sentence except the $l^{th}$ word, $l$ spans S, $WSIM$ represents the minimum word similarity value which is a hyperparameter dependent on dataset size.\\\\
% \begin{equation}
$DQI_{c4}=\frac{size(S)}{\sum_{S}({\forall_{l}\abs{\frac{\sum_{m}WSim_{lm}}{length(l)}-WSIM}})+1}$
% $DQI_{c4}=\frac{size(S)}{\sum_{S}({\forall_{l}\left|\frac{\sum_{m}WSim_{lm}}{length(l)}-WSIM}\right|)+1}$
% \end{equation}
\paragraph{Intra-sample STS: } This represents similarity between the premise and hypothesis in NLI, question and answer in QA, and passage and answer in RC. Similarity should not be too high or too low, so that the model does not have the scope to exploit it as bias. However, the variance should be high so that the model does not get biased by always expecting a data with fixed premise-hypothesis similarity. A similar analogy holds for the variation of sentence length among premise and hypothesis. Also there should be lower word overlap and word similarity among premise and hypothesis. Here $p$ represents sentences from one side, such as premises for NLI, and $h$ represents sentences from the other side, such as hypothesis for NLI; $s_p$ represents premise length and $s_h$ represents hypothesis length, $uw$ represents unique words, $q$ spans the sample, $Wsim$ represents word similarity, $hyp$ represents hypothesis. $ISIM$ represents the minimum similarity value which is a hyper-parameter.\\\\
% \begin{equation}
% $DQI_{c5}=\frac{size(X)}{({\sum_{X}\abs{{\forall_{p}\forall_{h}Sim_{ph}-ISIM}}}})+1} +\frac{size(X)}{\sum_{X}\abs{(s_p-s_h)}+1}+\frac{\sigma(\abs{(s_p-s_h)})}{size(X)}+ \frac{\sigma(\forall_{p}\forall_{h}Sim_{ph})}{size(X)}+\frac{\sum_X(\frac{s_p+s_h}{\forall_{uw}\sum_{q} \sign(2-\nu_{sample})})}{size(X)}+\frac{\sum_X(\frac{1}{\forall_{uw}\sum_{hyp.} \max\limits_{premise}Wsim})}{size(X)}$
% $DQI_{c5}=\frac{size(X)}{({\sum_{X}\left|{\forall_{p}\forall_{h}Sim_{ph}-ISIM}}\right|})+1} +\frac{size(X)}{\sum_{X}\left|(s_p-s_h)\right|+1}+\frac{\sigma(\left|(s_p-s_h)\right|)}{size(X)}+ \frac{\sigma(\forall_{p}\forall_{h}Sim_{ph})}{size(X)}+\frac{\sum_X(\frac{s_p+s_h}{\forall_{uw}\sum_{q} \sign(2-\nu_{sample})})}{size(X)}+\frac{\sum_X(\frac{1}{\forall_{uw}\sum_{hyp.} \max\limits_{premise}Wsim})}{size(X)}$
% \end{equation}
$DQI_{c5}=\frac{size(X)}{\sum_{x}\left | \forall_{p}\forall_{h}Sim_{ph}-ISIM \right |+1}+\frac{size(X)}{\sum_{X}\left | (s_p-s_h) \right |+1}+\frac{\sigma(\left | (s_p-s_h) \right |)}{size(X)}+\frac{\sigma(\forall_{p} \forall_{h} Sim_{ph} )}{size(X)}+\frac{\sum_X(\frac{s_p+s_h}{\forall_{uw}\sum_{q} \sign(2-\nu_{sample})})}{size(X)}+\frac{\sum_X(\frac{1}{\forall_{uw}\sum_{hyp.} \max\limits_{premise}Wsim})}{size(X)}$
\paragraph{N-gram Frequency per Label: }These frequency distributions should not be skewed towards a specific label. Also, the lesser variance there is across labels, the higher the data  quality. Here, the hyper-parameter $g$ is the upper limit for total number of words (and others in $i$) across any individual label. $Count_{label}$ is a vector of size 3 which represent how many times a word (and others in $i$) has been assigned each of the labels.\\\\
% \begin{equation}
$DQI_{c6}=\sum_{labels}(\sum_{i}\frac{1}{\sigma(\frac{i(\nu)}{size(i)})}*\frac{\sum_{i}((g-\nu_{i}))}{size(i)}+\frac{size(X_{label})}{(\sum_{X_{label}}(\left|(s_p-s_h)\right|))+1}+\frac{\sigma(\left|(s_p-s_h)\right|)}{size(X_{label})})+\sum_{i}\frac{size(i(X))}{(\sum_{i(X)}\sigma(\forall_{X}\frac{(\left|{1-Count_{label}}\right|-(1-Count_{label}))}{2}))+1}$
% $DQI_{c6}=\frac{1}{\sum_{i}\sum_{l}Var(f_{lx_i})+\sum_{j}\sum_{l}Var(f_{lx_j})}$
% \end{equation}
\paragraph{Inter-split STS: } For a sample in the test data, the most similar training data sample should have a similarity value within an upper cap. The similarity level between the train and test samples should also have a minimum lower cap. The closer the similarity value is towards the lower cap, the higher the data quality. $X_{train}$ and $X_{test}$ represent data in the train and test spilts respectively.
% $S_{sample}$ enlists total number of sentences in a sample, e.g.  it is 2 for NLI.
$Sim_{train-test}$ stands for similarity between the train and test data and $SSIM$ stands for the spilt overlap allowance which is a hyper-parameter.\\\\
% \begin{equation}
$DQI_{c7}=\frac{size(X_{test})}{(\sum_{test}{\left|\max\limits_{X_{train}}{Sim_{train-test}}-SSIM\right|})+1}$
% \end{equation}
%this is to avoid the crowd-worker from repeating all words to make an example. Also, note that we don't have test cases to cover extremes. We should write those.
% We have not added any property related to grammaticality as we assume that all the sentences in the dataset are valid sentences, which means that they will be grammatically corect.
%histogram should have higher variance.
\\\\
We propose the empirical formula of DQI as a function of all components.\\\\
% \begin{equation}
$DQI=f(DQI_1, DQI_2, DQI_3, DQI_4, DQI_5, \\
DQI_6, DQI_7)$\\
% \end{equation}
$f$ depends on both task and dataset, and thus needs to be experimentally tuned.

% \caption{Term-wise and Overall Values for $DQI_{c1}$} with a=3, b=30.

\section{DQI Evaluation and Discussion} \label{dqieval}
We use AFLite \cite{sakaguchi2019winogrande}, a recently proposed approach for adversarial filtering, to evaluate DQI. First, we filter SNLI and divide it into two categories (i) good (ii) bad  where `good' and `bad' refer to the set of samples retained and removed respectively. We calculate DQI components for each of the category and analyze results.
\subsection{Vocabulary}
\paragraph{Which characteristics of data are covered? }
This component takes the following characteristics of data into account:
(i) size of the existing vocabulary, (ii) sentence length distribution, and (iii) contribution of sentences to vocabulary given their length. 
\paragraph{Termwise Breakdown:}
The first term measures the magnitude of vocabulary of the data. The second measures the standard deviation of sentence lengths. A penalty is imposed by the last term, to check if a sentence's length lies in an acceptable range. This range is a hyperparameter that is determined based on the distribution of sentence lengths seen in the dataset.
\begin{table}
\centering
\scriptsize
\resizebox{1.0\columnwidth}{!}{%
\begin{tabular}{lllll}
\hline
\textbf{Term} & \textbf{T1} & \textbf{T2} & \textbf{T3} & \textbf{DQI C1} \\
\hline
\textbf{Good} & \textbf{1.8996} & \textbf{6.0409}   & \textbf{0.9532} & \textbf{7.6578} \\
\textbf{Bad}  & 0.6416 & 5.8135   & 0.9494 & 6.1609 \\\hline
              &            &            &            &                
\end{tabular}%
}
\caption{Term-wise and Overall Values for $DQI_{c1}$}.
\label{tab:dqi1}
\end{table}
\paragraph{Which category has higher DQI?}
Table \ref{tab:dqi1} shows that the good category of data has higher DQI than the bad category. Of the three terms in this component, the first term showed the most significant difference. Even though the second and third term are higher for the good category, the difference is less than expected. We were expecting a higher difference because, sentence length has been found to be an important parameter related to bias in SNLI, as discussed in section \ref{plead}. 
\begin{figure}
\includegraphics[width=8cm]{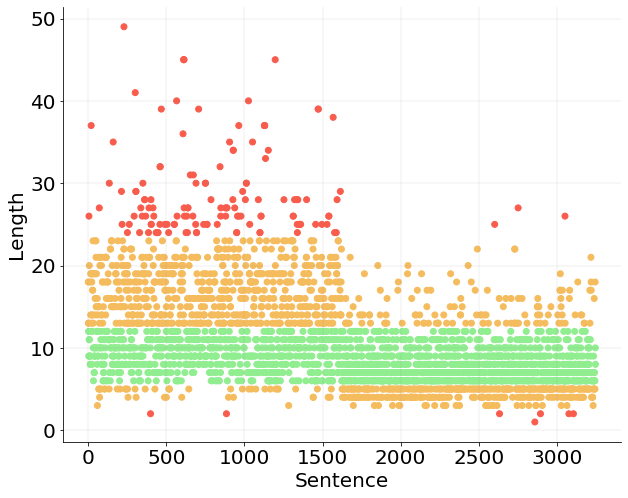}
  \caption{Sentence Lengths for Good Category}
\label{fig:goodlength}
\end{figure}
\begin{figure}
\includegraphics[width=8cm]{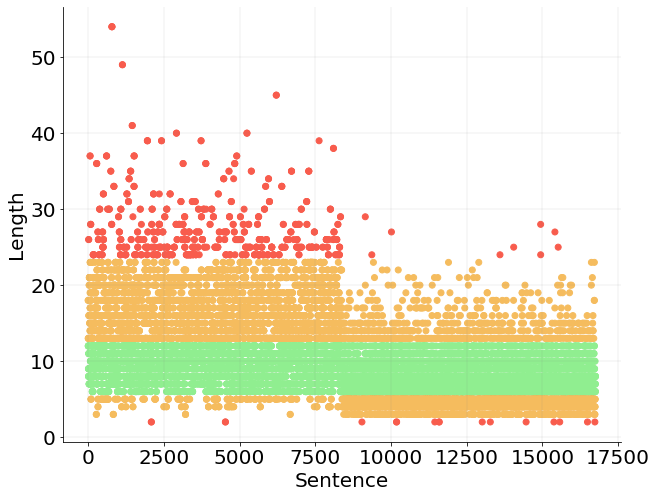}
  \caption{Sentence Lengths for Bad Category}
\label{fig:badlength}
\end{figure}
\begin{figure}
\includegraphics[width=8cm]{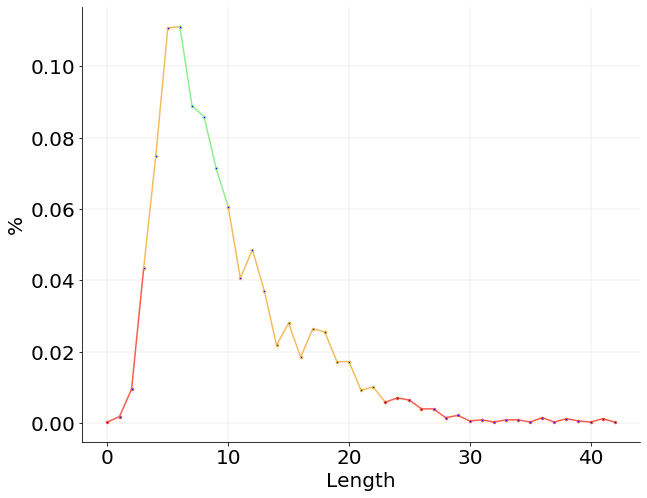}
  \caption{Sentence Length vs. Percentage of Samples for Good Category}
\label{fig:goodlengthper}
\end{figure}
\begin{figure}
\includegraphics[width=8cm]{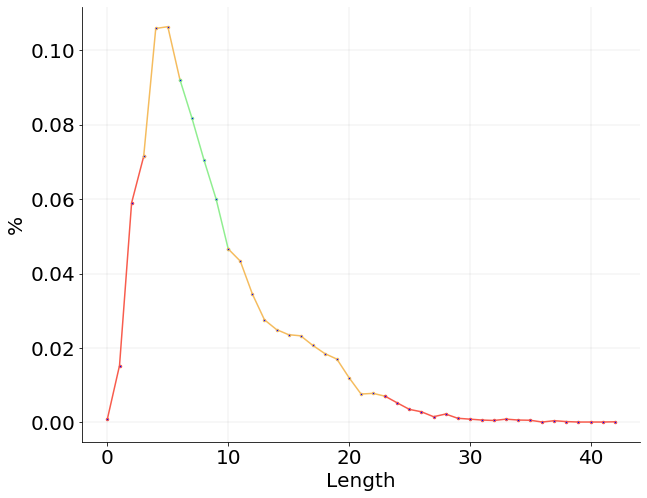}
  \caption{Sentence Length vs. Percentage of Samples for Bad Category}
\label{fig:badlengthper}
\end{figure}
\begin{figure}
\includegraphics[width=8cm]{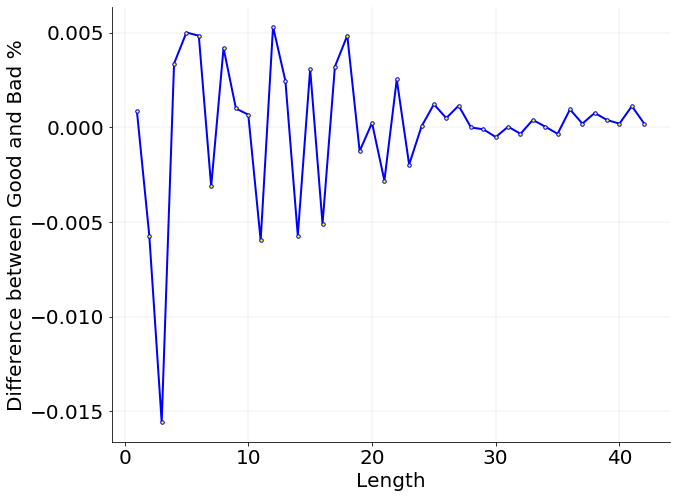}
  \caption{Difference between Splits of Sentence Length vs. Percentage of Samples}
\label{fig:lengthdff}
\end{figure}
\paragraph{Sentence length variation not significant across category}
We analyze sentence length variation closely across the good and bad categories. Figure \ref{fig:goodlength} and Figure \ref{fig:badlength} show that sentence length variation follows a  similar pattern in both categories. We further find the percentage of samples for various sentence lengths and calculate the difference between them across categories. Figures \ref{fig:goodlengthper}, \ref{fig:badlengthper} and \ref{fig:lengthdff} further confirm that there is no significant difference in sentence length variation across category. This might indicate that AFLite is not appropriately removing data with bias associated with sentence length.
%does not take the length of sentences into account while determining the split.
\subsection{Inter-sample N-gram Frequency and Relation:}
\paragraph{Which Characteristics of Data are Covered? }
The data is analyzed at different granularities using this component, namely in terms of POS Tags, Words, Bigrams, Trigrams and Sentences. The POS tags considered are those of Adjectives, Adverbs, Verbs, and Nouns. The terms are constructed to (i) analyze the distribution of the granularity considered, and (ii) impose an acceptable range of values for each granularity.
\begin{table} 
\centering
\scriptsize
\resizebox{1.0\columnwidth}{!}{%
\begin{tabular}{lllll}
\hline
\textbf{Granularity} & \textbf{Split} & \textbf{T1} & \textbf{T2} & \textbf{Contribution} \\
\hline
\textbf{Words} &\textbf{Good}& \textbf{121.9512} & \textbf{0.7269}   & \textbf{88.6463} \\
 &\textbf{Bad}& 52.3560 & 0.6500   & 34.0314 \\\hline
\textbf{Adjectives} &\textbf{Good}& \textbf{31.7460} & \textbf{0.2966}   & \textbf{9.4159} \\
 &\textbf{Bad}& 16.9205 & 0.3590  & 6.0745 \\\hline
\textbf{Adverbs} &\textbf{Good}& \textbf{21.0970} & \textbf{0.1847}  & \textbf{3.8966} \\
 &\textbf{Bad}& 10.7875 & 0.1732   & 1.8684 \\\hline
\textbf{Verbs} &\textbf{Good}& \textbf{43.6681} & \textbf{0.2349} & \textbf{10.2576} \\
 &\textbf{Bad}& 16.5289 & 0.1893  & 3.1289 \\\hline
\textbf{Nouns}&\textbf{Good}& \textbf{49.2611} & \textbf{0.4351}  & \textbf{21.4335} \\
 &\textbf{Bad}& 21.0084 & 0.3685   & 7.7416 \\\hline
\textbf{Bigrams} &\textbf{Good}& \textbf{1296.3443} & \textbf{0.9374}  & \textbf{1215.1931} \\
 &\textbf{Bad}& 873.2862 & 0.9355   & 816.9592 \\\hline
\textbf{Trigrams} &\textbf{Good}& \textbf{7686.3951} & \textbf{0.9546} & \textbf{7337.4328} \\
  &\textbf{Bad}& 6119.9510 & 0.9422  & 5766.2178 \\\hline
\textbf{Sentences} &\textbf{Good}& 9070.7819 & \textbf{0.6607} & \textbf{5993.0656} \\
 &\textbf{Bad}& \textbf{14537.0541} & 0.2705   & 3932.2731\\\hline
 \textbf{Sentences} &\textbf{Good}& \textbf{3.0656} & \textbf{0.6607} & \textbf{3.7263} \\
 \textbf{(Not Normalized)} &\textbf{Bad}& 1.2655 & 0.2705   & 1.0607\\\hline
\textbf{DQIC2} &\textbf{Good}& - & - & \textbf{8668.3012} \\
  &\textbf{Bad}& -& - & 6636.3641 \\\hline
  
              &            &            &        &                   
\end{tabular}%
}
\caption{Term-wise and Overall Values for $DQI_{c2}$, Good Split}
\label{tab:dqi2}
\end{table}

% \textbf{Sentences} &\textbf{Good}& \textbf{3.0656} & 0.6607 & 3.7263 \\
%  &\textbf{Bad}& 1.2655 & 0.2705   & 1.0607 \\\hline
%sentences, good: 9070.7819, bad: 14537.0541

\paragraph{First Term:}
The first term measures the standard deviations of the granularities. In order to ensure high data quality, there should be minimal variance in the frequency distributions across all granularities, i.e., variance is inversely proportional to data quality. Normalization based on the number of units per granularity is done to ensure a fair comparison. This is in order to ensure that no single unit in any granularity provides spurious bias for the model to learn. Therefore, the good split of AFLite is expected to have lower standard deviations for all granularities compared to the bad split. Table \ref{tab:dqi2} shows that this property holds for everything except sentences. We investigate and find that, this is because sentences are repeated very few times unlike words and other granularities. So, we decide to find the first term without normalization. Table \ref{tab:dqi2} shows that the property also holds for sentences without normalization.

\paragraph{Second Term:}
The range in the second term is a hyperparameter, which is decided for each granularity based on its distribution in the dataset. Each unit considered for all granularities should have a minimum frequency, in order for the model to get favorable bias. On the other hand, they must have an upper limit so that the model does not get a chance to use it as a spurious bias.  The second term is directly proportional to the data quality. This means that each good split granularity should have a higher value than its corresponding bad split granularity. As shown in Table \ref{tab:dqi2}, this passes in all cases. 

%The use of a modified signum function
% \begin{table*} 
% \centering
% \scriptsize
% \resizebox{1.2\columnwidth}{!}{%
% \begin{tabular}{lllll}
% \hline
% \textbf{Granularity} & \textbf{Split} & \textbf{T1} & \textbf{T2} & \textbf{DQI C2} \\
% \hline
% \textbf{Sentences} &\textbf{Good}& \textbf{3.0656} & 0.6607 & 3.7263 \\
%  &\textbf{Bad}& 1.2655 & 0.2705   & 1.0607 \\\hline
% \textbf{Words} &\textbf{Good}& \textbf{121.9512} & 0.7269   & 0.7351 \\
%  &\textbf{Bad}& 52.3560 & 0.6500   & 0.6691 \\\hline
% \textbf{Adjectives} &\textbf{Good}& \textbf{31.7460} & 0.2966   & 0.3281 \\
%  &\textbf{Bad}& 16.9205 & 0.3590  & 0.4181 \\\hline
% \textbf{Adverbs} &\textbf{Good}& \textbf{21.0970} & 0.1847  & 0.2321 \\
%  &\textbf{Bad}& 10.7875 & 0.1732   & 0.2659 \\\hline
% \textbf{Verbs} &\textbf{Good}& \textbf{43.6681} & 0.2349 & 0.2578 \\
%  &\textbf{Bad}& 16.5289 & 0.1893  & 0.2498 \\\hline
% \textbf{Nouns}&\textbf{Good}& \textbf{49.2611} & 0.4351  & 0.4554 \\
%  &\textbf{Bad}& 21.0084 & 0.3685   & 0.4161 \\\hline
% \textbf{Bigrams} &\textbf{Good}& \textbf{1296.3443} & 0.9374  & 0.9382 \\
%  &\textbf{Bad}& 873.2862 & 0.9355   & 0.9356 \\\hline
% \textbf{Trigrams} &\textbf{Good}& \textbf{7686.3951} & 0.9546 & 0.9547 \\
%   &\textbf{Bad}& 6119.9510 & 0.9422  & 0.9424 \\\hline
%               &            &            &        &                   
% \end{tabular}%
% }
% \caption{Term-wise and Overall Values for $DQI_{c2}$, Good Split}
% \label{tab:my-table}
% \end{table*}

%sentences, good: 9070.7819, bad: 14537.0541

\begin{figure}
\includegraphics[width=8cm]{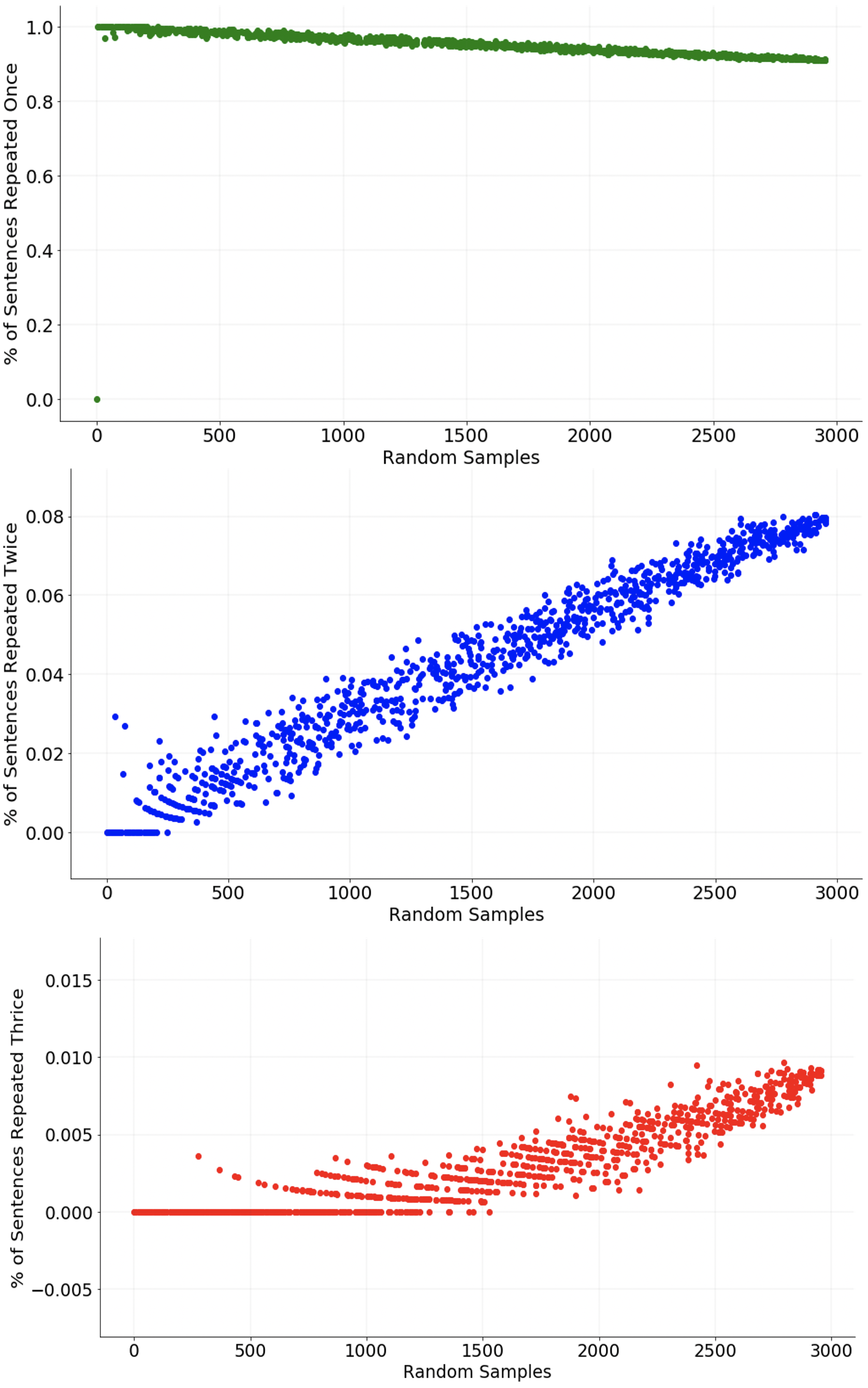}
  \caption{Distribution of repetition in randomly sampled sentence subsets of good split}
\label{fig:dqic6good}
\end{figure}
\begin{figure}
\includegraphics[width=8cm,height=20cm]{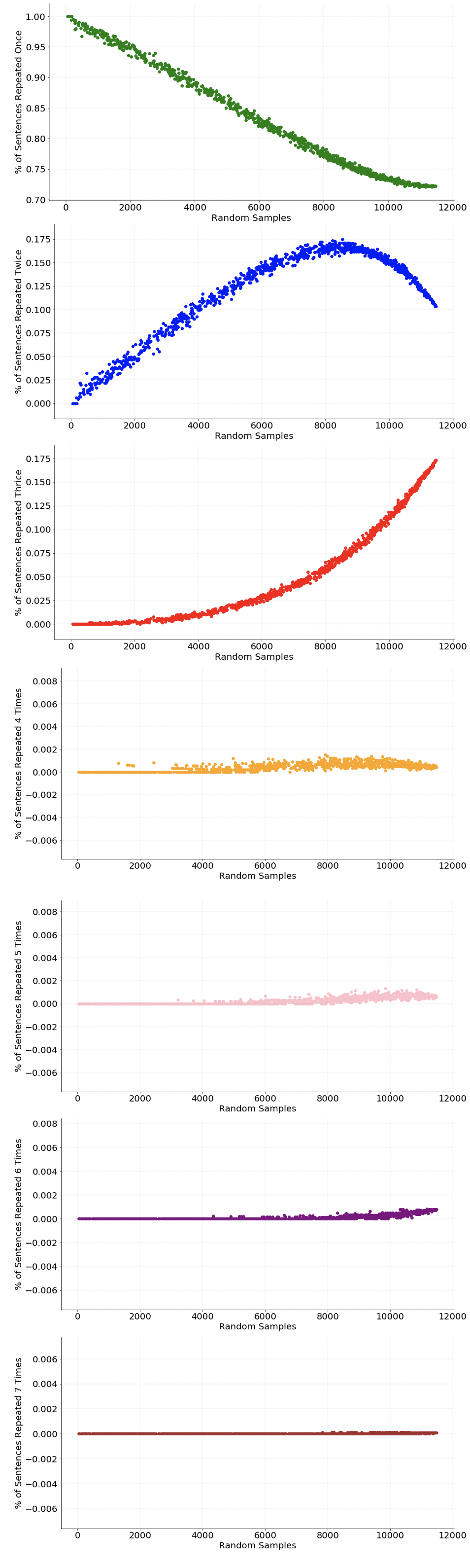}
  \caption{Distribution of repetition in randomly sampled sentence subsets of bad split}
\label{fig:dqic6bad}
\end{figure}

\subsection{Inter-sample STS}
\begin{table} 
\centering
\scriptsize
\resizebox{1.0\columnwidth}{!}{%
\begin{tabular}{llll}
\hline
\textbf{Split} & \textbf{SIML=0.3} & \textbf{SIML=0.35}& \textbf{SIML=0.4} \\
\hline
\textbf{Good} & 9.1320 & 11.3955   & 14.3267 \\
\textbf{Bad} & \textbf{10.3842} & \textbf{13.1062}   & \textbf{16.6390} \\\hline
              &            &            &                            
\end{tabular}%
}
\caption{Term 1 for $DQI_{c3}$}
\label{tab:dqi31}
\end{table}

\begin{table} 
\centering
\scriptsize
\resizebox{1.0\columnwidth}{!}{%
\begin{tabular}{llll}
\hline
\textbf{Split} & \textbf{e=0.25} & \textbf{e=0.33}& \textbf{e=0.5} \\
\hline
\textbf{Good} & \textbf{0.0468} & \textbf{0.0244}   & \textbf{0.0103} \\
\textbf{Bad} & 0.0404 & 0.0216   & 0.0094 \\\hline
              &            &            &                          
\end{tabular}%
}
\caption{Term 2 for $DQI_{c3}$, with SIML=0.4}
\label{tab:dqi32}
\end{table}

\begin{table} 
\resizebox{1.0\columnwidth}{!}{%
  \begin{tabular}{llll}
  \hline
    \multirow{2}{*}{\textbf{Sample Set}} &\multicolumn{3}{c}{\textbf{DQI C3 (e=0.5)}}\\
      &SIM=0.5&SIM=0.6&SIM=0.7\\\hline      
     \textbf{Good} &9.4123 &11.4508&14.3370  \\
     \textbf{Bad} & \textbf{10.3936} & \textbf{13.1156} & \textbf{16.7024} \\
     \hline
  \end{tabular}%
  }
 \caption{$DQI_{C3}$}
 \label{tab:dqi3final}
\end{table}

% \paragraph{Inter-sample STS}
%Semantic Textual Similarity (STS) is analyzed by this component. 
\paragraph{Which  Characteristics  of  Data  are  Covered?}
Here, similarities are calculated between (i) every possible pair of individual sentences in the good category, and (ii) every possible sentence pair in the bad category. We take random samples of the bad category with size equal to that of the good category to perform experiments on a minimal computational budget. We consider multiple random samples for a fair comparison. The terms: (i) check if sentences meet the minimum similarity threshold required for providing favorable bias to a model, and (ii) provide a bound on the number of sentences that have this minimum score.
\begin{figure}
\includegraphics[width=7cm]{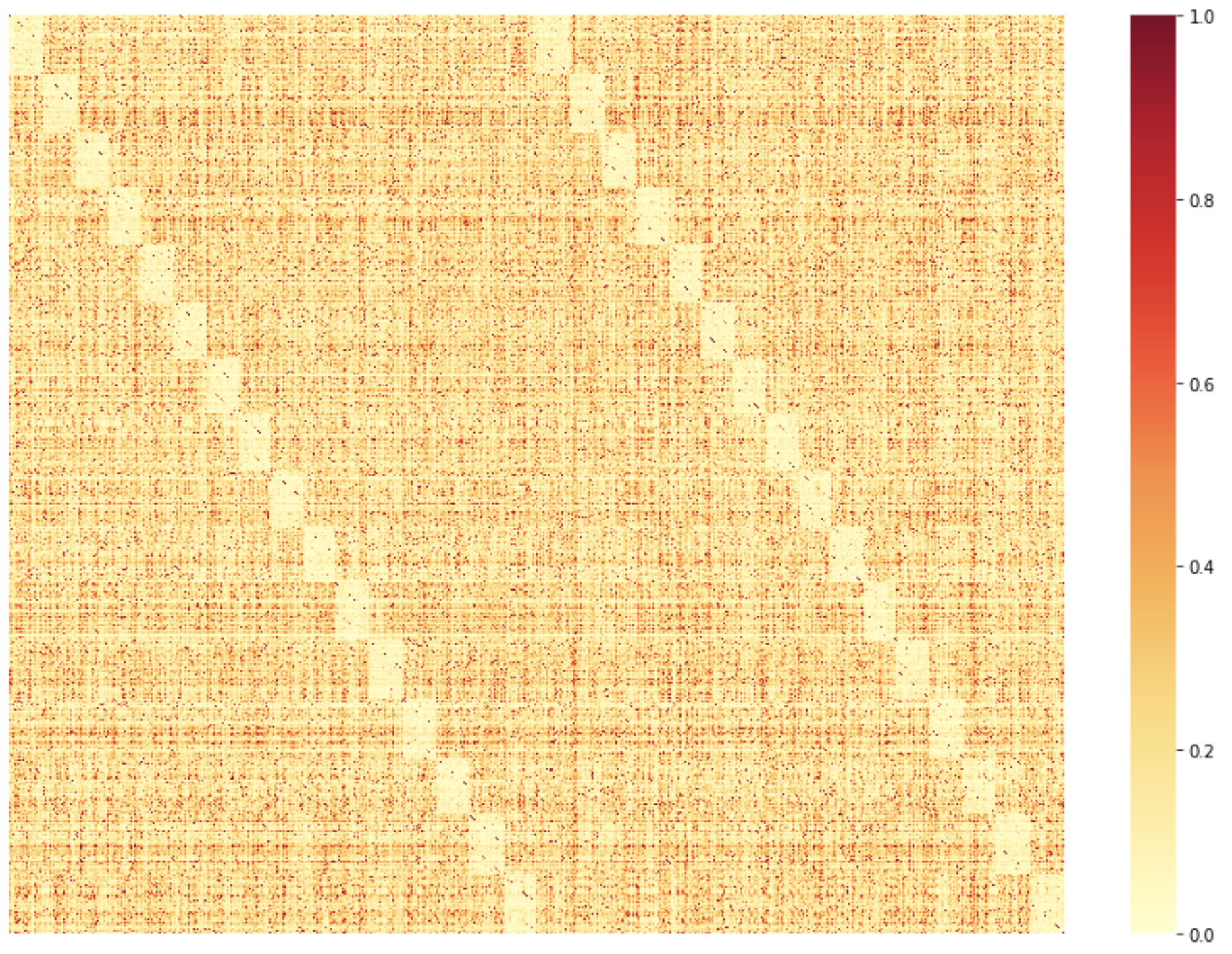}
  \caption{Sentence Similarity for Good Category}
\label{fig:dqi3goodsentence}
\end{figure}
\begin{figure}
\includegraphics[width=7cm]{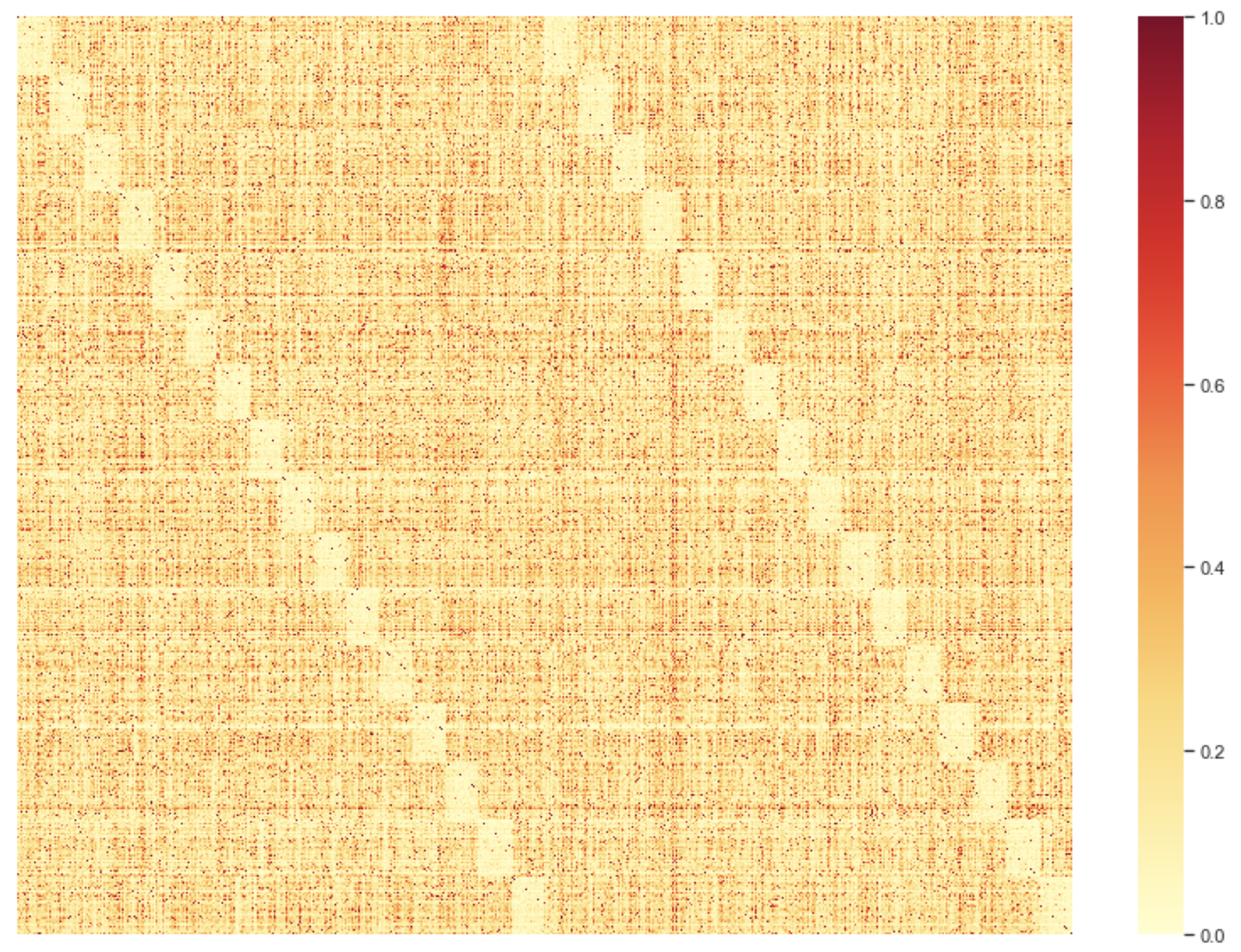}
  \caption{Sentence Similarity for Bad Category}
\label{fig:dqi3badsentence}
\end{figure}
\paragraph{First Term}
The first term has a hyperparameter that dictates the minimum similarity threshold. Over the dataset, given each sentence in turn, all other sentences are checked against it and those which don't meet the threshold are counted. The standard deviation of this series should be low, and is  inversely proportional to the term's value. The accountability of this term is similar to class imbalance. Table \ref{tab:dqi31} shows that the good category has lower value than the bad category. We analyze it further and can see the same pattern in Figure \ref{fig:dqic6good}, \ref{fig:dqic6bad} This might indicate that, AFLite is not considering imbalance due to sentence similarity.
\paragraph{Second Term}
The second term utilizes two hyperparameters, the threshold from the first term and the lower bound on the number of sentences that should meet this threshold. The summation term should therefore be low, as it counts the number of sentences that fail to meet the threshold. Table \ref{tab:dqi32} shows that all categories pass this.
% \begin{table*} 
% \centering
% \scriptsize
% \resizebox{1.2\columnwidth}{!}{%
% \begin{tabular}{llll}
% \hline
% \textbf{Split} & \textbf{SIML=0.3} & \textbf{SIML=0.35}& \textbf{SIML=0.4} \\
% \hline
% \textbf{Good} & 0.1095 & 0.0878   & 0.0698 \\
% \textbf{Bad} & 0.0963 & 0.0763   & 0.0601 \\\hline
%               &            &            &                            
% \end{tabular}%
% }
% \caption{Term 1 for $DQI_{c3}$}
% \label{tab:my-table}
% \end{table*}
% \begin{table*} 
% \centering
% \scriptsize
% \resizebox{1.2\columnwidth}{!}{%
% \begin{tabular}{llll}
% \hline
% \textbf{Split} & \textbf{a=0.25} & \textbf{a=0.33}& \textbf{a=0.5} \\
% \hline
% \textbf{Good} & 269.4854 & 344.2961   & 486.4793 \\
% \textbf{Bad} & 272.9440 & 349.7231   & 497.1452 \\\hline
%               &            &            &                          
% \end{tabular}%
% }
% \caption{Term 2 for $DQI_{c3}$, with SIML=0.4}
% \label{tab:my-table}
% \end{table*}
% \begin{table*} 
% \centering
% \scriptsize
% \resizebox{1.2\columnwidth}{!}{%
% \begin{tabular}{llll}
% \hline
% \textbf{Split} & \textbf{a=0.25} & \textbf{a=0.33}& \textbf{a=0.5} \\
% \hline
% \textbf{Good} & 0.0936 & 0.0487   & 0.0205 \\
% \textbf{Bad} & 0.0807 & 0.0432   & 0.0188 \\\hline
%               &            &            &                          
% \end{tabular}%
% }
% \caption{Term 2 for $DQI_{c3}$, with SIML=0.4}
% \label{tab:my-table}
% \end{table*}
\subsection{Intra-sample Word Similarity}
% \paragraph{Average Similarity with Existing Words in the Sentence}
\paragraph{Which  Characteristics  of  Data  are  Covered?}
This component consists of a single term, that captures how close the similarity values between all words in a single sentence are to a minimum word similarity value, which is a hyperparameter. The closer the mean of all similarities is to the hyperparameter value, the higher the data quality. This follows from the reasoning that words that a low sum implies noisy data and a high sum implies high pair wise bias in the data. Therefore, the denominator of the term should be as low as possible, meaning that the DQI component should be higher for the good category than the bad category. For a hyperparameter value of 0.5, we observe that the good category has a higher component value than the bad category.

\begin{table} 
\centering
\scriptsize
\resizebox{0.5\columnwidth}{!}{%
\begin{tabular}{ccc}
\hline
\textbf{Split} & \textbf{DQIC4}  \\
\hline
\textbf{Good} & 0.000372 \\
\textbf{Bad} & 0.000062 \\\hline
              &            &                 
\end{tabular}%
}
\caption{$DQI_{c4}$}
\label{tab:my-tablec4}
\end{table}
%, with WSIML=0.5
\begin{table} 
\centering
\scriptsize
\resizebox{1.0\columnwidth}{!}{%
\begin{tabular}{lllll}
\hline
\textbf{Split} & \textbf{ISIM=0.3}& \textbf{ISIM=0.4}& \textbf{ISIM=0.5}& \textbf{ISIM=0.6}\\
\hline
\textbf{Good} & \textbf{2.2349} & \textbf{2.8763} & \textbf{4.0125} & \textbf{6.3065}\\
\textbf{Bad}& 2.2215 & 2.8558 & 3.9784 & 6.2237 \\\hline
              &            &            &            &              
\end{tabular}%
}
\caption{Term 1 for $DQI_{c5}$}
\label{tab:dqi51}
\end{table}
\begin{table} 
\centering
\scriptsize
\resizebox{1.0\columnwidth}{!}{%
\begin{tabular}{llllll}
\hline
\textbf{Split} & \textbf{T2}& \textbf{T3}& \textbf{T4}& \textbf{T5}& \textbf{T6} \\
\hline
\textbf{Good} & \textbf{0.1439} & \textbf{0.0038} & \textbf{6.4064e-05}& \textbf{20.3518}& \textbf{0.0903}\\
\textbf{Bad} & 0.1430& 0.0007& 1.2711e-05&19.9288 &0.0900 \\\hline
              &   & &   & &   
\end{tabular}%
}
\caption{Terms 2,3,4,5,6 for $DQI_{c5}$}
\label{tab:dqi52}
\end{table}
\begin{table} 
\centering
\scriptsize
\resizebox{0.5\columnwidth}{!}{%
\begin{tabular}{ll}
\hline
\textbf{Split} & \textbf{DQI C5} \\
\hline
\textbf{Good} & \textbf{24.6024}\\
\textbf{Bad} & 24.1409\\\hline
              &                                      
\end{tabular}%
}
\caption{$DQI_{c5}$, with ISIM=0.5}
\label{tab:my-tablec5}
\end{table}
\subsection{Intra-sample STS}
% \paragraph{Inter Sentence Similarity in a Data}
\paragraph{Which  Characteristics  of  Data  are  Covered?}
Premise-Hypothesis similarity within samples is addressed by this component. Five aspects of the dataset are analyzed: (i) how far premise-hypothesis pairs are from a particular similarity threshold, (ii) how much the length variation between premise and hypothesis is, (iii) how much the variation in similarities across all pairs in a dataset is, (iv) what the level of word overlap between the premise and hypothesis is, and (v) what the maximum level of word similarity between the premise and hypothesis is.
\paragraph{First Term}
The first term computes if the sentence similarity across a given sample meets a threshold, which is a hyperparameter. This sum should be low and so the term should be high, because if the similarity between premise hypothesis pairs is far from the hyperparameter, the sample might give rise to spurious bias. Table \ref{tab:dqi51} shows that the good category has higher component value than the bad category for a range of hyperparameters.
% This also ensures that most samples fall below the threshold defined in the first term.
\paragraph{Second Term}
The second term measures the length variation in the good and bad categories, between the premise and hypothesis. This variation is computed as a mean of differences. The mean should be less so that the model does not get a chance to use hypothesis length as an artifact. Even though the term has a higher value for good category, it appears to be almost the same for both categories. Table \ref{tab:dqi52} shows this behavior. 
% This might indicate that, the length artifact is not be properly captured by AFLite. 
\paragraph{Third Term}
The variance should be high to cover all possible cases, so that the model does not adhere to fixed length difference and over-fit. This explains the 3rd term. Table \ref{tab:dqi52} shows that the term is higher for the good category
\paragraph{Fourth Term}
The fourth term measures the overall variance of within sample similarity over all samples. This is normalized to account for datasets' differing sizes. It should be high to ensure that the model does not get over-fitted to a certain similarity between premise and hypothesis. Here, the term is slightly higher for the good category.
\paragraph{Fifth Term}
The word overlap level between the premise and hypothesis should be low. The stop words are removed from the dataset and the number of words that overlap are counted for each sample and summed. The length of the concatenated premise and hypothesis sentences is divided by the count to normalize this term. The term is higher for the good category.
\paragraph{Sixth Term}
Another way of capturing word related bias within the sample is to pick the maximally similar words from the premise of each word in the hypothesis. This may help account for those words actually used as context. The maximal similarities found are summed and reciprocated, and then normalized by multiplying by the size of the dataset considered. This term is seen to be higher for the good category.
\paragraph{Overall Component value does not have a significant difference across categories} 
This component captures several major leads as discussed in Section \ref{plead}. So, we were expecting a significant difference across categories for this component. However, 
Table 14 says that the component value of the good category is not very different from that of the bad category. This might indicate that AFLite is not accurately filtering data with high premise-hypothesis similarity and length difference.
% the argument of 2nd and 3rd term is also applicable for 1st and 4th term.

% \begin{table*} 
% \centering
% \scriptsize
% \resizebox{1.2\columnwidth}{!}{%
% \begin{tabular}{lll}
% \hline
% \textbf{Split} & \textbf{Variance} & \textbf{Mean} \\
% \hline
% \textbf{Good} & 0.02297 & 6.9464\\
% \textbf{Bad} & 0.0040 & 6.9925\\\hline
%               &            &                           
% \end{tabular}%
% }
% \caption{Term 2 for $DQI_{c5}$}
% \label{tab:my-table}
% \end{table*}

% \begin{table*} 
% \centering
% \scriptsize
% \resizebox{1.2\columnwidth}{!}{%
% \begin{tabular}{ll}
% \hline
% \textbf{Split} & \textbf{T3} \\
% \hline
% \textbf{Good} & 6.6515e-06\\
% \textbf{Bad} & 1.3533e-06\\\hline
%               &                                      
% \end{tabular}%
% }
% \caption{Term 3 for $DQI_{c5}$}
% \label{tab:my-table}
% \end{table*}
\begin{table} 
\centering
\scriptsize
\resizebox{1.0\columnwidth}{!}{%
\begin{tabular}{ccccc}
\hline
\textbf{Split/Label} & \textbf{Entailment} & \textbf{Neutral}& \textbf{Contradiction} \\
\hline
\textbf{Good} & 1110 & 1430   & 708 \\
\textbf{Bad} & 5626 & 5008   & 6118 \\\hline
              &            &            &       &                    
\end{tabular}%
}
\caption{Sample counts for Splits across Labels}
\label{tab:dqi6spilt}
\end{table}
\begin{table} 
\centering
\scriptsize
\resizebox{1.0\columnwidth}{!}{%
\begin{tabular}{llll}
\hline
\textbf{Split-Label} & \textbf{T1}& \textbf{T2} \\
\hline
\textbf{Good-Entailment}  & 8829.2425 & \textbf{0.9387}\\
\textbf{Bad-Entailment}  & \textbf{21655.2868} & 0.8571\\
\textbf{Good-Neutral}  & 7467.5349 & 0.8699\\
\textbf{Bad-Neutral}  & \textbf{31616.2545} & \textbf{0.9141}\\
\textbf{Good-Contradiction} & 4932.7421 & \textbf{0.9210} \\
\textbf{Bad-Contradiction} & \textbf{29145.0957} & 0.8783 \\\hline
              &            &            &                          
\end{tabular}%
}
\caption{Terms 1 and 2 for $DQI_{c6}$, Sentence Granularity}
\label{tab:dqi61sent}
\end{table}
\begin{table} 
\centering
\scriptsize
\resizebox{1.0\columnwidth}{!}{%
\begin{tabular}{llll}
\hline
\textbf{Split-Label}  & \textbf{T1}& \textbf{T2} \\
\hline
\textbf{Good-Entailment}  & \textbf{142.8571} & \textbf{0.7277}\\
\textbf{Bad-Entailment}  & 81.9672 & 0.6110\\
\textbf{Good-Neutral}  & \textbf{153.8462} & \textbf{0.9118}\\
\textbf{Bad-Neutral}  & 117.6471 & 0.7071\\
\textbf{Good-Contradiction}  & \textbf{163.9344} & \textbf{0.6764} \\
\textbf{Bad-Contradiction}  & 101.0101 & 0.6088 \\\hline
              &            &            &                          
\end{tabular}%
}
\caption{Terms 1 and 2 for $DQI_{c6}$, Word Granularity}
\label{tab:dqi61word}
\end{table}

\begin{table} 
\centering
\scriptsize
\resizebox{1.0\columnwidth}{!}{%
\begin{tabular}{llll}
\hline
\textbf{Split-Label}  & \textbf{T1}& \textbf{T2} \\
\hline
\textbf{Good-Entailment}  & \textbf{42.1230} & \textbf{0.34114}\\
\textbf{Bad-Entailment}  & 26.4201 & 0.30551\\
\textbf{Good-Neutral}  & \textbf{48.8998} & 0.46865\\
\textbf{Bad-Neutral}  & 38.1534 & \textbf{0.47497}\\
\textbf{Good-Contradiction}  & \textbf{43.1593} & 0.31019 \\
\textbf{Bad-Contradiction}  & 29.2826 & \textbf{0.32385} \\\hline
              &            &            &                          
\end{tabular}%
}
\caption{Terms 1 and 2 for $DQI_{c6}$, Adjective Granularity}
\label{tab:dqi61adj}
\end{table}

\begin{table} 
\centering
\scriptsize
\resizebox{1.0\columnwidth}{!}{%
\begin{tabular}{llll}
\hline
\textbf{Split-Label}  & \textbf{T1}& \textbf{T2} \\
\hline
\textbf{Good-Entailment}  & \textbf{18.4128} & 0.056911\\
\textbf{Bad-Entailment}  & 11.0963 & \textbf{0.05816}\\
\textbf{Good-Neutral}  & 8.6798 & 0.09709\\
\textbf{Bad-Neutral}  & \textbf{14.6135} & \textbf{0.43124}\\
\textbf{Good-Contradiction}  & \textbf{37.9795} & \textbf{0.34286} \\
\textbf{Bad-Contradiction}  & 23.7192 & 0.21583 \\\hline
              &            &            &                          
\end{tabular}%
}
\caption{Terms 1 and 2 for $DQI_{c6}$, Adverb Granularity}
\label{tab:dqi61adv}
\end{table}

\begin{table} 
\centering
\scriptsize
\resizebox{1.0\columnwidth}{!}{%
\begin{tabular}{llll}
\hline
\textbf{Split-Label}  & \textbf{T1}& \textbf{T2} \\
\hline
\textbf{Good-Entailment}  & \textbf{41.7885} & \textbf{0.16091}\\
\textbf{Bad-Entailment}  & 22.9410 & 0.05348\\
\textbf{Good-Neutral}  & \textbf{48.9476} & 0.17946\\
\textbf{Bad-Neutral}  & 38.9105 & \textbf{0.20192}\\
\textbf{Good-Contradiction} & \textbf{53.5045} & \textbf{0.20000} \\
\textbf{Bad-Contradiction}  & 34.6380 & 0.13589 \\\hline
              &            &            &                          
\end{tabular}%
}
\caption{Terms 1 and 2 for $DQI_{c6}$, Verb Granularity}
\label{tab:dqi61verb}
\end{table}

\begin{table} 
\centering
\scriptsize
\resizebox{1.0\columnwidth}{!}{%
\begin{tabular}{llll}
\hline
\textbf{Split-Label}  & \textbf{T1}& \textbf{T2} \\
\hline
\textbf{Good-Entailment}  & \textbf{59.2768} & \textbf{0.49650}\\
\textbf{Bad-Entailment}  & 34.3643 & 0.38238\\
\textbf{Good-Neutral}  & \textbf{62.7353} & \textbf{0.44534}\\
\textbf{Bad-Neutral}  & 46.4253 & 0.40586\\
\textbf{Good-Contradiction}  & \textbf{66.3570} & \textbf{0.45653} \\
\textbf{Bad-Contradiction}  & 39.9202 & 0.37431 \\\hline
              &            &            &                          
\end{tabular}%
}
\caption{Terms 1 and 2 for $DQI_{c6}$, Noun Granularity}
\label{tab:dqi61noun}
\end{table}

\begin{table} 
\centering
\scriptsize
\resizebox{1.0\columnwidth}{!}{%
\begin{tabular}{llll}
\hline
\textbf{Split-Label}  & \textbf{T1}& \textbf{T2} \\
\hline
\textbf{Good-Entailment}  & 1131.7133 & \textbf{0.93307}\\
\textbf{Bad-Entailment}  & \textbf{1173.5409} & 0.93206\\
\textbf{Good-Neutral}  & 1261.2663 & 0.93783\\
\textbf{Bad-Neutral}  & \textbf{1598.1514} & \textbf{0.94117}\\
\textbf{Good-Contradiction}  & 1100.8597 & \textbf{0.94325} \\
\textbf{Bad-Contradiction}  & \textbf{1369.0528} & 0.93387 \\\hline
              &            &            &                          
\end{tabular}%
}
\caption{Terms 1 and 2 for $DQI_{c6}$, Bigram Granularity}
\label{tab:dqi61bigram}
\end{table}

\begin{table} 
\centering
\scriptsize
\resizebox{1.0\columnwidth}{!}{%
\begin{tabular}{llll}
\hline
\textbf{Split-Label}  & \textbf{T1}& \textbf{T2} \\
\hline
\textbf{Good-Entailment}  & 5921.2942 & \textbf{0.94672}\\
\textbf{Bad-Entailment}  & \textbf{7757.5306} & 0.93496\\
\textbf{Good-Neutral}  & 6414.8208 & 0.94517\\
\textbf{Bad-Neutral}  & \textbf{10229.7186} & \textbf{0.95015}\\
\textbf{Good-Contradiction}  & 5478.1014 & \textbf{0.95359} \\
\textbf{Bad-Contradiction}  & \textbf{8984.3224} & 0.94430 \\\hline
              &            &            &                          
\end{tabular}%
}
\caption{Terms 1 and 2 for $DQI_{c6}$, Trigram Granularity}
\label{tab:dqi61trigram}
\end{table}
\begin{figure}
\includegraphics[width=8cm,height=20cm]{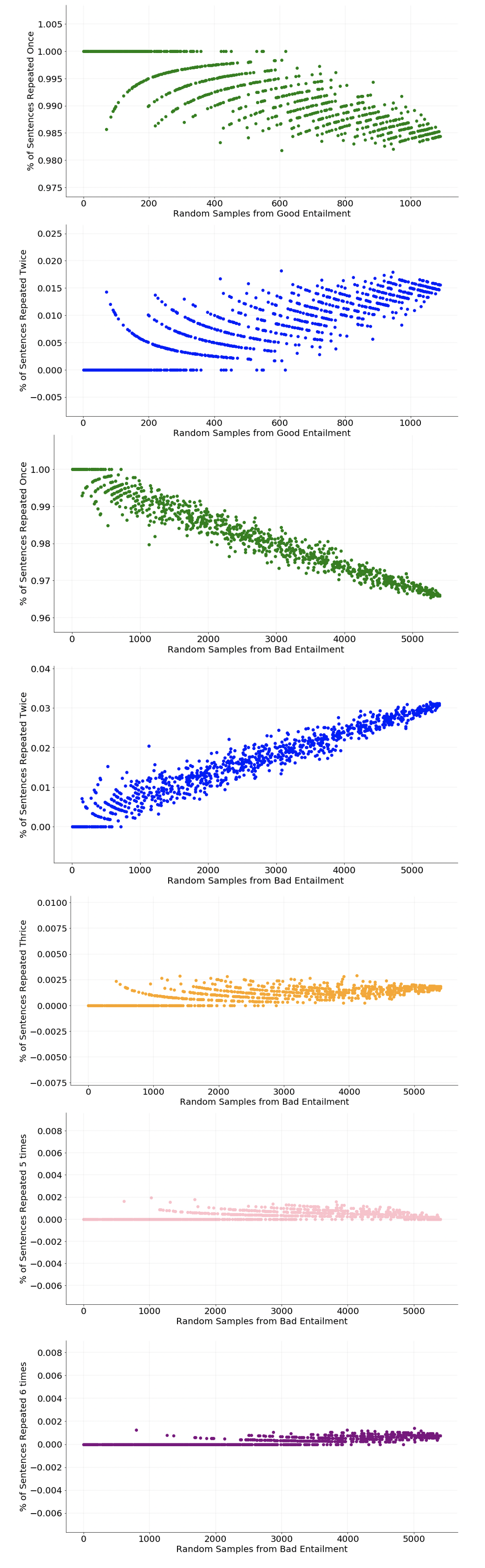}
  \caption{Distribution of repetition in randomly sampled sentence subsets of entailment samples from good and bad splits}
\label{fig:dqic6goodsentencee}
\end{figure}
\begin{figure}
\includegraphics[width=8cm,height=20cm]{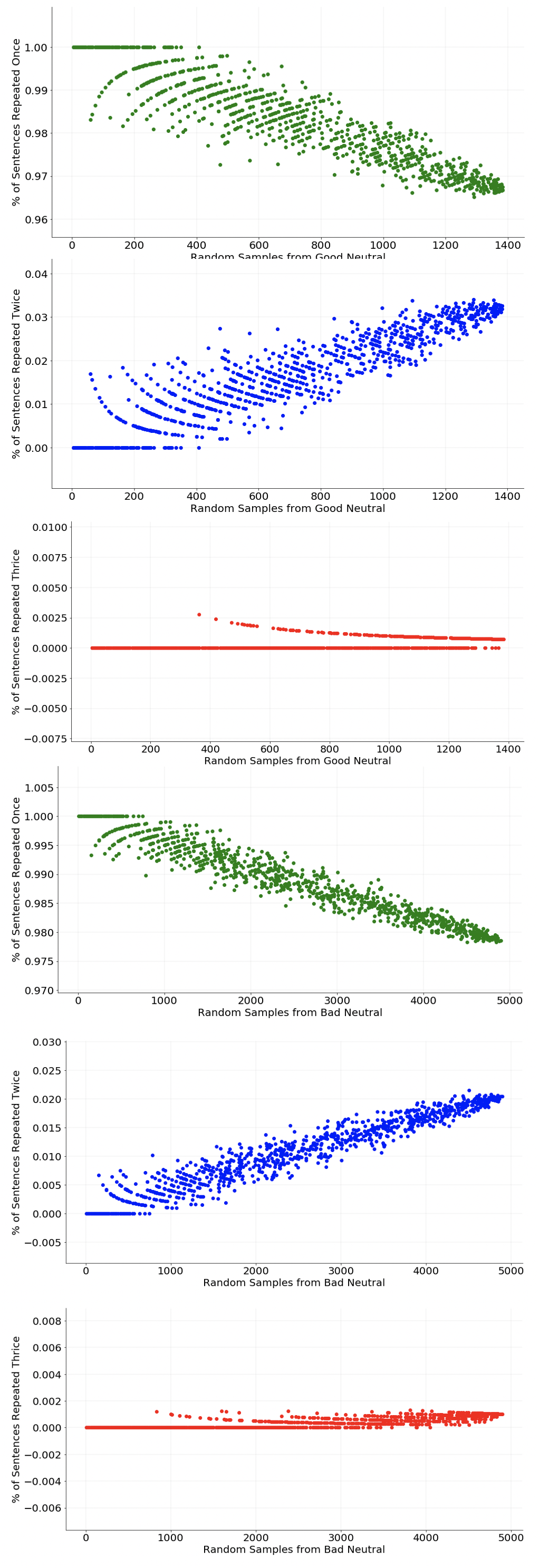}
  \caption{Distribution of repetition in randomly sampled sentence subsets of neutral samples from good and bad splits}
\label{fig:dqic6goodsentencen}
\end{figure}
\begin{figure}
\includegraphics[width=8cm,height=20cm]{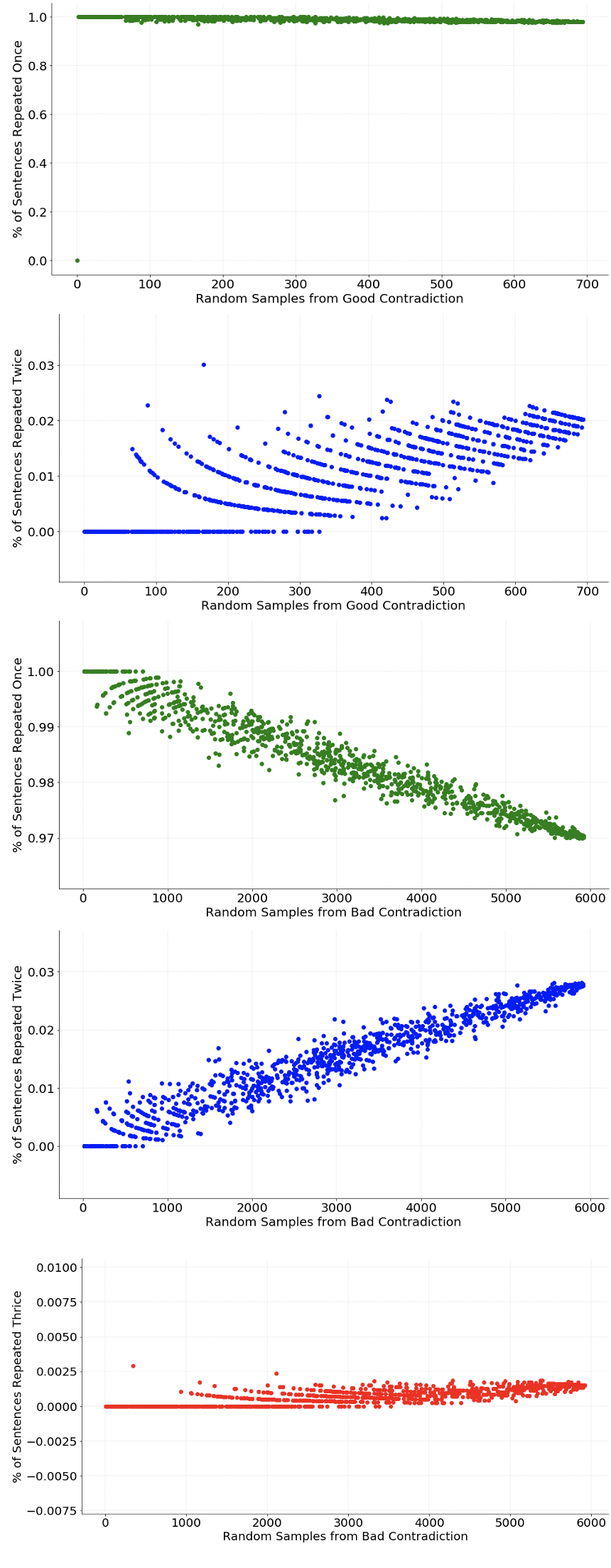}
  \caption{Distribution of repetition in randomly sampled sentence subsets of contradiction samples from good and bad splits}
\label{fig:dqic6goodsentencec}
\end{figure}
\begin{table*} 
\centering
\scriptsize
\resizebox{2.0\columnwidth}{!}{%
\begin{tabular}{lllllll}
\hline
\textbf{Split-Repetition} & \textbf{1} & \textbf{2}& \textbf{3} & \textbf{4}& \textbf{5}& \textbf{6}\\
\hline
\textbf{Good-Entailment} & 0.984446 & 0.015554 & 0 & 0 & 0 & 0 \\
\textbf{Bad-Entailment} &  0.965976& 0.030880&    0.001849&    0 &0.000740&    0.000555 \\
\textbf{Good-Neutral} &0.966739&    0.032538&    0.000723& 0 & 0 &0 \\
\textbf{Bad-Neutral} &0.978563&    0.020416&    0.001021& 0 & 0 & 0 \\
\textbf{Good-Contradiction} &    0.979827&    0.020173  & 0 & 0 & 0 & 0 \\
\textbf{Bad-Contradiction} &     0.978563&    0.020416&    0.001021 & 0 & 0 & 0 \\\hline
              &      &      &    &    &    &      
\end{tabular}%
}
\caption{Sentence Granularity Repetitions}
\label{tab:dqi6sentrep}
\end{table*}
\begin{table} 
\centering
\scriptsize
\resizebox{0.7\columnwidth}{!}{%
\begin{tabular}{ll}
\hline
\textbf{Split-Label} & \textbf{T3}\\
\hline
\textbf{Good-Entailment} & \textbf{0.1457} \\
\textbf{Bad-Entailment} & 0.1330 \\
\textbf{Good-Neutral} & 0.1496 \\
\textbf{Bad-Neutral} & \textbf{0.1571} \\
\textbf{Good-Contradiction} & 0.1313 \\
\textbf{Bad-Contradiction} & \textbf{0.1434} \\\hline
              &                                    
\end{tabular}%
}
\caption{T3 for $DQI_{c6}$}
\label{tab:dqi63}
\end{table}

\begin{table} 
\centering
\scriptsize
\resizebox{0.7\columnwidth}{!}{%
\begin{tabular}{ll}
\hline
\textbf{Split-Label} & \textbf{T4}\\
\hline
\textbf{Good-Entailment} & \textbf{0.0100} \\
\textbf{Bad-Entailment} & 0.0021 \\
\textbf{Good-Neutral} & \textbf{0.0084} \\
\textbf{Bad-Neutral} & 0.0022 \\
\textbf{Good-Contradiction} & 0.0197 \\
\textbf{Bad-Contradiction} & \textbf{0.0020} \\\hline
              &                                    
\end{tabular}%
}
\caption{T4 for $DQI_{c6}$}
\label{tab:dqi64}
\end{table}
\begin{table} 
\centering
\scriptsize
\resizebox{1.0\columnwidth}{!}{%
\begin{tabular}{lll}
\hline
\textbf{Granularity/Split} & \textbf{Good}& \textbf{Bad}\\
\hline
\textbf{Sentences} & \textbf{15.3475} & 11.6614\\
\textbf{Words} & \textbf{0.9313} & 0.6596  \\
\textbf{Adjectives} & \textbf{1.2190} & 0.9185  \\
\textbf{Adverbs} & \textbf{1.5708} & 1.1850 \\
\textbf{Verbs} & \textbf{0.9667} & 0.7001 \\
\textbf{Nouns} & \textbf{1.0623} & 0.7358  \\
\textbf{Bigrams} & 0.3646 & \textbf{0.4893} \\
\textbf{Trigrams} & 0.1860 & \textbf{0.2760} \\\hline
              &      &
\end{tabular}%
}
\caption{T5 for $DQI_{c6}$}
\label{tab:dqi65}
\end{table}

\begin{table} 
\centering
\scriptsize
\resizebox{0.7\columnwidth}{!}{%
\begin{tabular}{ll}
\hline
\textbf{Split-Label} & \textbf{DQI C6}\\
\hline
\textbf{Good} & \textbf{556.6914} \\
\textbf{Bad} &  320.2893\\\hline
              &                                    
\end{tabular}%
}
\caption{$DQI_{c6}$}
\label{tab:dqic6final}
\end{table}
\subsection{N-gram Frequency per Label}
% \paragraph{Words, Bigram, Trigram, and Sentence Frequencies per Label}
\paragraph{Which  Characteristics  of  Data  are  Covered?}
The features of data that lead to label bias are captured by this component. The data is analyzed at different granularities, as in the second component. Terms reflect the following characteristics of data: (i) distribution of of each granularity across labels, (ii) range of frequencies of units in each granularity per label, (iii) distribution of each granularity within each label, and (iv) average length between the premise and hypothesis in each sample, for all samples across labels.
\paragraph{Contradiction samples are seen to be more prone to spurious bias}
In order to compute the terms, the good and bad splits of data were further divided into three subsets each, corresponding to the gold labels of samples. On creating these subsets, we note that the ratio of contradiction samples in the good and bad categories is much higher than that seen in the case of entailment and neutral labels. Table \ref{tab:dqi6spilt} shows this.
% That is, AFLite considers very few contradiction samples in its good split. This indicates that hyperparameters for the contradiction sets must be set more strictly than for the other two labels, as contradiction samples seem to be more inherently prone to bias.
\paragraph{First Term}
Standard deviation is computed individually for each label, and then summed across labels in the first term. Following component two, the standard deviation is expected to be inversely proportional to data quality. We have normalized standard deviation and inverted it so that the term becomes directly proportional to DQI. Tables \ref{tab:dqi61sent}, \ref{tab:dqi61word}, \ref{tab:dqi61adj}, \ref{tab:dqi61adv}, \ref{tab:dqi61verb}, \ref{tab:dqi61noun}, \ref{tab:dqi61bigram}, \ref{tab:dqi61trigram} show that this passes in most cases, and fails for the bigram and trigram and sentence granularities across all labels, and adverb granularities in the neutral label, as the standard deviations seen of the good split are greater in these cases. We closely observe sentence repetition across labels. Based on the plots for sentence granularity distribution in each label, we observe that there is more repetition of sentences in the case of the bad split in the entailment and contradiction labels, but more in the good split for neutral labels. Figure \ref{fig:dqic6goodsentencee}, \ref{fig:dqic6goodsentencen},\ref{fig:dqic6goodsentencec} illustrate this. 
Since we find a higher percentage of unique sentences in the bad category compared to the good category in case of the neutral label, we analyze this further and find that sentences do not repeat significantly across labels, as shown in Table \ref{tab:dqi6sentrep}. However, failure in Bigrams and Trigrams might indicate that AFLite is not handling those cases appropriately.
\paragraph{Second Term}
The second term defines an acceptable range of values for units in each granularity, which is a hyperparameter that differs for different granularities. It follows the second term of the second component's relationship with data quality, i.e. direct proportionality. Interestingly, this fails only in the neutral label for a few granularities i.e. sentence, adjective, adverb, verb, bigram and trigram, and passes for everything else. This might indicate that, AFLite is not filtering appropriately for neutral category.

% Failure cases are bigram, trigram, sentence, adjective, adverb, and verb granularities for the neutral label, adverb granularity for the entailment label, and adjective and noun granularities for the contradiction label.
\paragraph{Third Term}
The variation in sentence lengths within a sample, i.e., the differences between the premise and hypothesis lengths per sample across all samples is calculated for each label in the third term. The mean should be lesser and close to 0 so that the model doesn't get a chance to use hypothesis length as a hyperparameter. Interestingly, it again fails for the neutral label along with the  contradiction label, as shown in Table \ref{tab:dqi63}
%This is found to be almost the same for each label across the good and bad splits.
Hence, we might infer that AFLite does not appropriately capture the artifact of sentence length across labels.
\paragraph{Fourth Term}
The fourth term calculates the standard deviation of sentence length difference between premise and hypothesis across labels. The standard deviation needs to be higher to ensure that there exist samples of varying difference betweeen premise and hypothesis length, and the model is not overfitted towards a fixed length difference. It passes for entailment and neutral label. It fails for contradiction label though both the terms are very close in that case. Table \ref{tab:dqi64} shows this.
\paragraph{Fifth Term}
The fifth term first computes the frequency of each unit in a granularity, to form vectors of length three for each unit. The standard deviation of this vector is calculated for each unit, if the unit is repeated. If the unit is not repeated, then it is not considered in our calculation. The sum of these standard deviations is calculated across the given granularity. This sum is normalized by division by the size of the set of units for that granularity, across all labels. The expectation is that the good split will show lower values of this term compared to the bad split, as lesser variance within labels is desirable. So, the term has been reversed to have direct proportionality with DQI. This is seen to fail in case of bigram and trigram granularities, as shown in Table \ref{tab:dqi65}.
\paragraph{Overall}
It is observed that bigrams and trigrams do not pass a majority of cases. Hence, they may not be informative/utilized enough by AFLite. The same is true for samples with the neutral label.
\begin{figure}
\includegraphics[width=7cm]{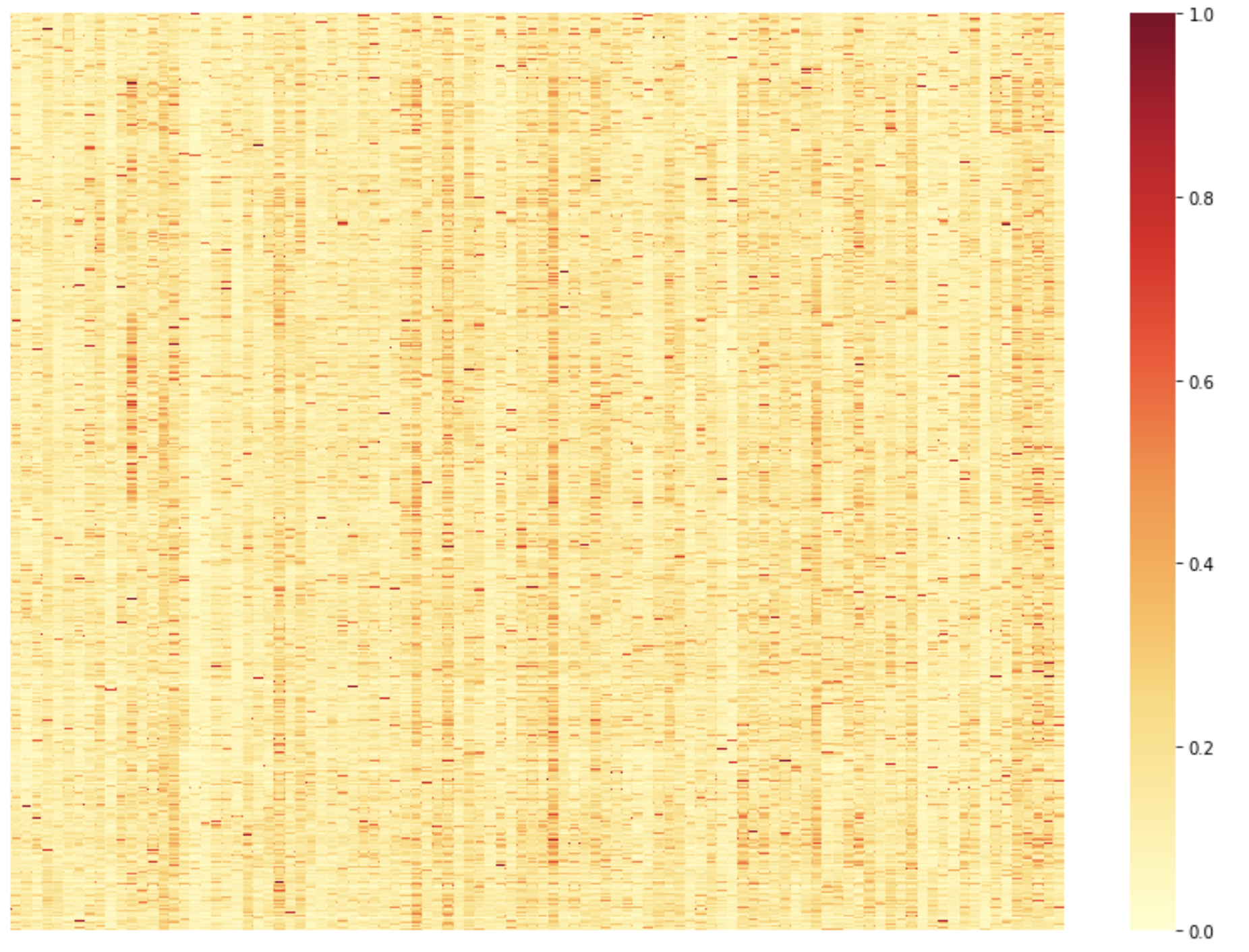}
  \caption{Sample Similarity: Test Good vs. Train Good}
\label{fig:dqi7good}
\end{figure}
\begin{figure}
\includegraphics[width=7cm]{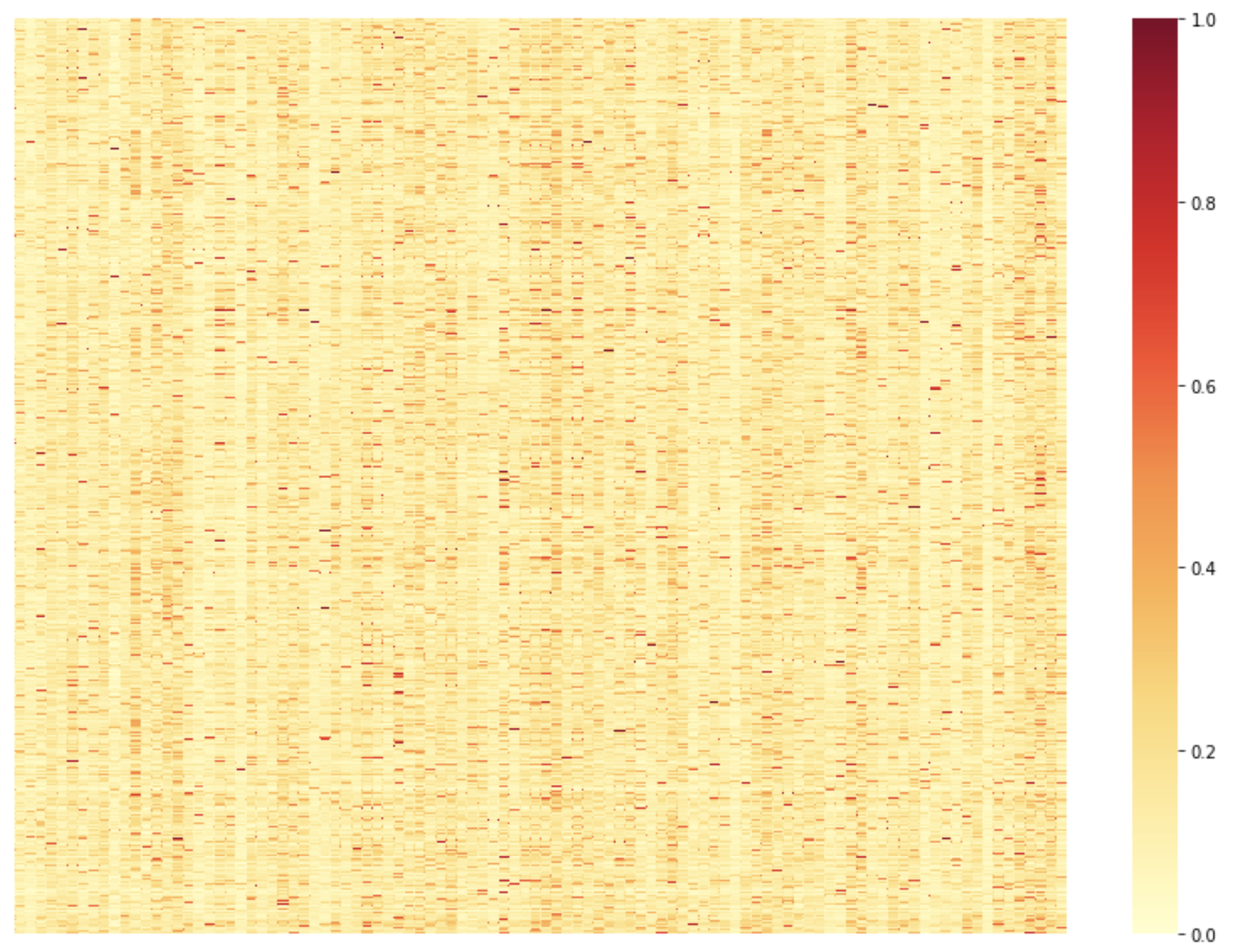}
  \caption{Sample Similarity: Test Good vs. Train Bad}
\label{fig:dqi7bad}
\end{figure}

\subsection{Inter-split STS}
% \paragraph{Similarity between Training Data and Test Data}
\paragraph{Which  Characteristics  of  Data  are  Covered?}
This component measures similarity between the training and test splits. We take random samples of the train bad category with a size equal to that of the train good category. We also consider 100 samples each of the test set for the good and bad categories. This is to perform experiments on a  minimal computational budget. However, we consider multiple random samples of both for a fair comparison. The maximum similar training sample for each test sample is found, and this pair's similarity value is checked against a bound value, which is a hyper-parameter. The sum of the terms should be low because it ensures the similarity is not too high or too low. A high value implies data leakage between train and test, and low value implies training set and test set are very different which unnecessarily makes the dataset hard, thus bad. So, this consists of only one term, which should be high in value, as only a small number of samples should be far from the threshold.
\paragraph{Which category has higher DQI?}
Table \ref{tab:dqi7} shows that the good category of data has higher DQI than the bad category. However, both the values are very similar. So, we analyze further and find that, there is no significant difference in similarity plots among categories, as illustrated in Figure \ref{fig:dqi7good} and \ref{fig:dqi7bad}. We were expecting a higher difference because, the train-test split has been found to be an important parameter related to bias in SNLI, as discussed in section \ref{plead}. This might indicate that, AFLite is not properly incorporating this lead while filtering.
% \begin{table*} 
% \centering
% \scriptsize
% \resizebox{1.2\columnwidth}{!}{%
% \begin{tabular}{llll}
% \hline
% \textbf{Split} & \textbf{SSMIL=0.2}& \textbf{SSMIL=0.3}& \textbf{SSMIL=0.4}\\
% \hline
% \textbf{Good} & 327.88003916241 & 237.34712743949004 & 158.77052428924998 \\
% \textbf{Bad} & 341.36231981418996 & 252.04458983771002 & 176.77938284252997 \\\hline
%               &        &        &                            
% \end{tabular}%
% }
% \caption{$DQI_{c7}$}
% \label{tab:my-table}
% \end{table*}

\begin{table} 
\centering
\scriptsize
\resizebox{1.0\columnwidth}{!}{%
\begin{tabular}{llll}
\hline
\textbf{Split} & \textbf{SSMIL=0.2}& \textbf{SSMIL=0.3}& \textbf{SSMIL=0.4}\\
\hline
\textbf{Good} & \textbf{0.0031} & \textbf{0.0042} & \textbf{0.0063} \\
\textbf{Bad} & 0.0029 & 0.0040 & 0.0057 \\\hline
              &        &        &                            
\end{tabular}%
}
\caption{$DQI_{c7}$}
\label{tab:dqi7}
\end{table}
\begin{figure*}
\includegraphics[width=\linewidth,height=8.55cm]{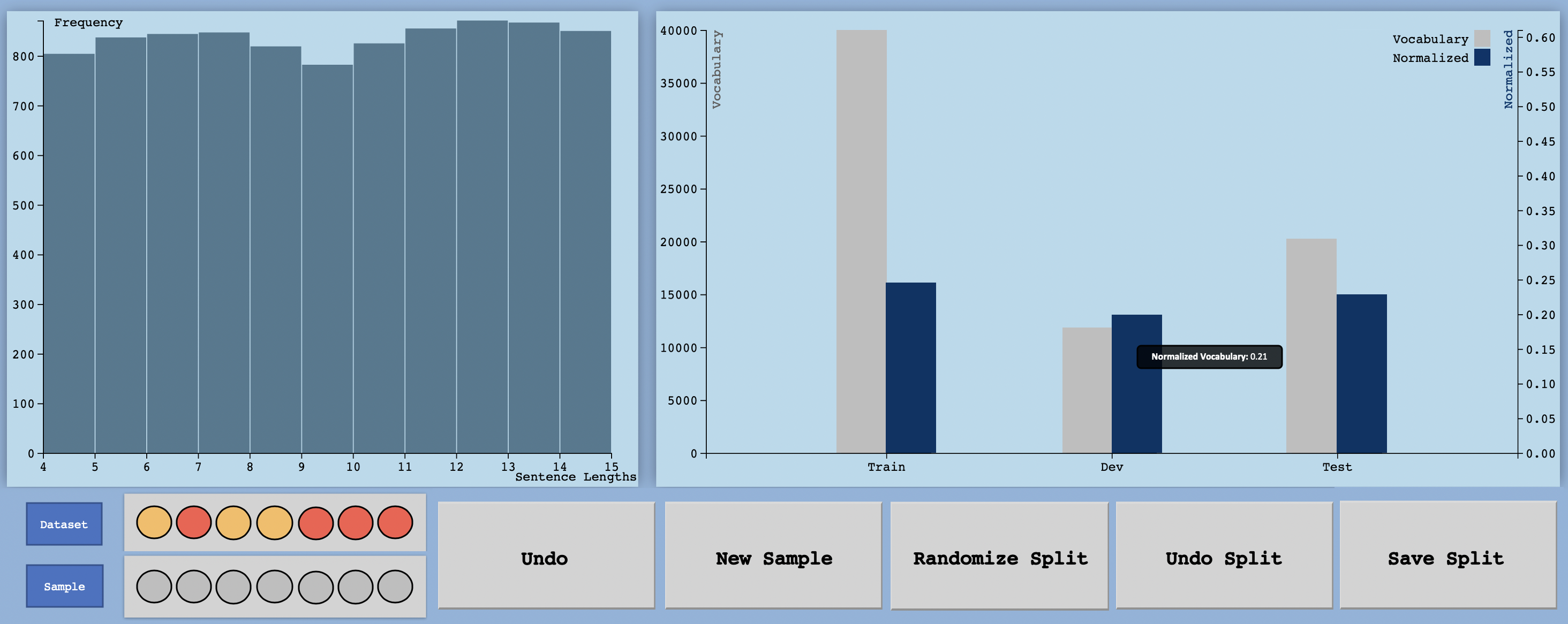}
  \caption{$DQI_{c1}$ Visualization Prior to New Sample Addition}
  \label{fig:Vis1before}
\end{figure*}
\section{Visualization of DQI}
\label{dqivizl}
% \paragraph{Method of Selection:}
\paragraph{Careful Selection of Visualizations}
Prior to the design of test cases and a user interface, data visualizations highlighting the effects of sample addition are built. Considering the complexity of the formulas for the components of empirical DQI, we carefully select visualizations to help illustrate and analyze the effect to which individual text properties are affected. 
% The 'Vocabulary' and 'Similarity between Training Data and Test Data' components are visualized split-wise. All other components are visualized in terms of the whole dataset.
%, given the overall component values.
\paragraph{All DQI Component Values are Shown for Each Visualization:}
We show all DQI component values for each visualization, since the user needs to optimize across several dependent components while selecting the best quality data. All DQI component values are tracked across different visualizations using two separate panels present at the bottom of the screen. The first panel shows the component-wise values as colored circles for the overall dataset prior to adding the sample. The second panel is initially a set of grayscale circles. Once the new sample is added, both the panels are updated. The first panel may not show any color changes, as it represents the overall dataset. The second however, will now display colored circles based on the DQI component values of the individual new sample. The values of the components can be viewed with a tooltip.
\paragraph{Traffic Signal Color Scheme: }
The color combination of Red-Yellow-Green used in all the visualizations represents the quality of the component/property being observed/analyzed. Here, red represents an undesirable quality value, yellow a permissible value, and green an ideal value. The color scale follows a pattern of red-yellow-green-yellow-red unless otherwise specified, centered around the ideal value of a component.
\subsection{Vocabulary}
\begin{figure*}
\includegraphics[width=\linewidth,height=8.55cm]{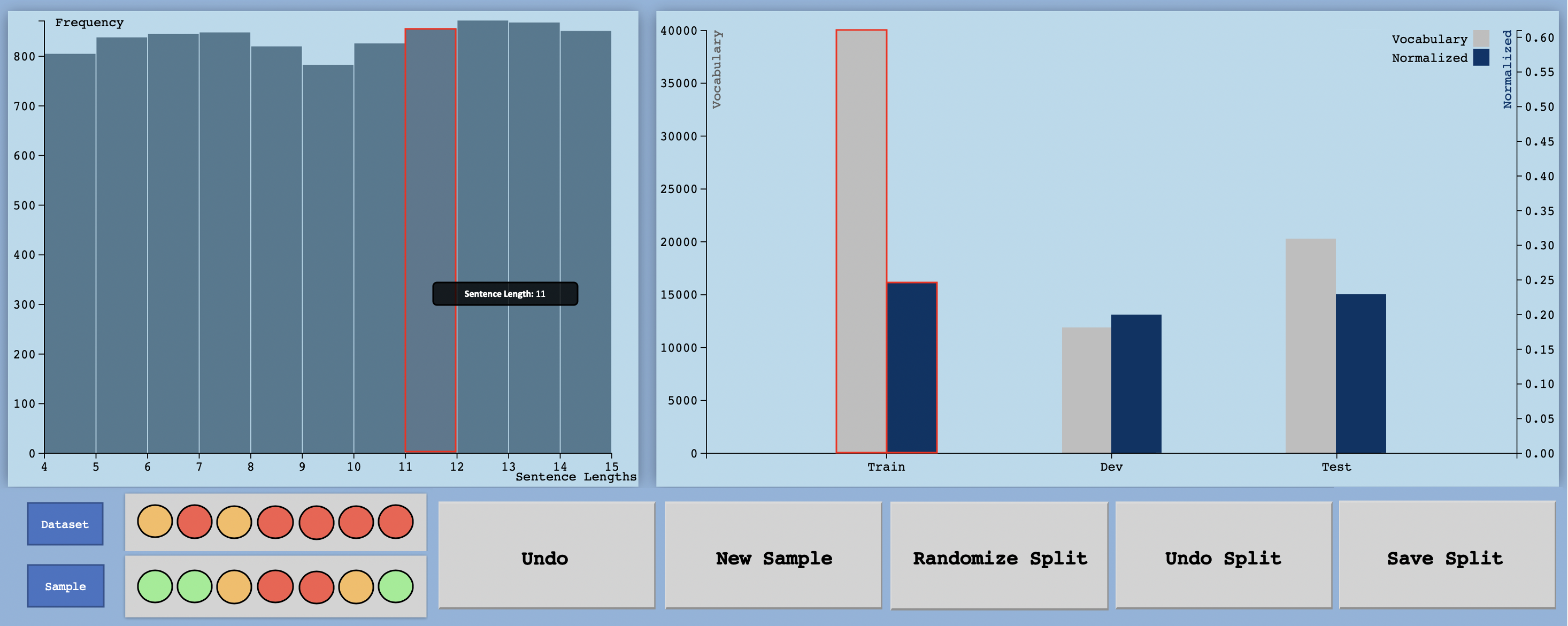}
  \caption{$DQI_{c1}$ Visualization On New Sample Addition}
  \label{fig:Vis1after}
\end{figure*}

\paragraph{Which Characteristics of Data are Visualized?}
The contribution of samples to the size of the vocabulary is tracked using a dual axis bar chart. This displays the vocabulary size, along with the vocabulary magnitude, across the train, dev, and test splits for the dataset. Also, the distribution of sentence lengths is plotted as a histogram. Each sample contributes two sentences, i.e., the premise and hypothesis statements. Figure \ref{fig:Vis1before} illustrates this.
\paragraph{Interactions:}
Interactions are supported through a tooltip and buttons. The tooltip displays the quantities in both charts on mouseover, and the buttons are used to update the chart. There are five tasks supported by the buttons:
\begin{itemize}[noitemsep]
    \item \textbf{Addition of a New Sample \textit{(New Sample)}:} The new sample is added to the train split by default. A script to calculate the new words this sample contributes to the vocabulary set is run, and the bar chart is accordingly updated. The sentence lengths of the premise and hypothesis statements are used to update the histogram. The updated portions of both the charts are highlighted, as shown in Figure \ref{fig:Vis1after}. The component value panels are updated as well. The previous state of the visualization is saved in a set of variables. 
    \item \textbf{Removal of a New Sample \textit{(Undo)}:} This reverses the operations of 'addition of a new sample' by using the saved state variables to restore the visualizations back to their original state.
    \item \textbf{Randomization of Split \textit{(Randomize Split)}:} The samples are distributed randomly between the train, dev, and test splits, using a 70:10:20 split ratio. Once the split is randomized, the new sample cannot be removed from the split anymore, as it is not necessarily a part of the train set. In order to account for annotator bias, the annotator id of dataset samples is used to create mutually exclusive annotator sets across splits.  Additionally, the split is designed such that if a premise has multiple hypothesis statements and is therefore repeated across samples, then all samples containing that premise belong to the same split. This split operation can be performed multiple times, as an attempt to understand the effect of data ordering on the DQI component values for the overall dataset. The previous state of the visualization is saved in a set of variables. 
    \item \textbf{Undo Split \textit{(Undo Split)}:} This reverses the operations of 'randomization of split' by using the saved state variables to restore the visualizations back to their original state. Only the latest randomization operation is reversed.
    \item \textbf{Save Split \textit{(Save Split)}:} Once the split is satisfactory, this button can be used to freeze this split state for the remainder of the analysis. On addition of the next sample, this frozen state is used for the initialization of the visualizations.
\end{itemize}
\begin{figure*}
\includegraphics[width=\linewidth,height=8.55cm]{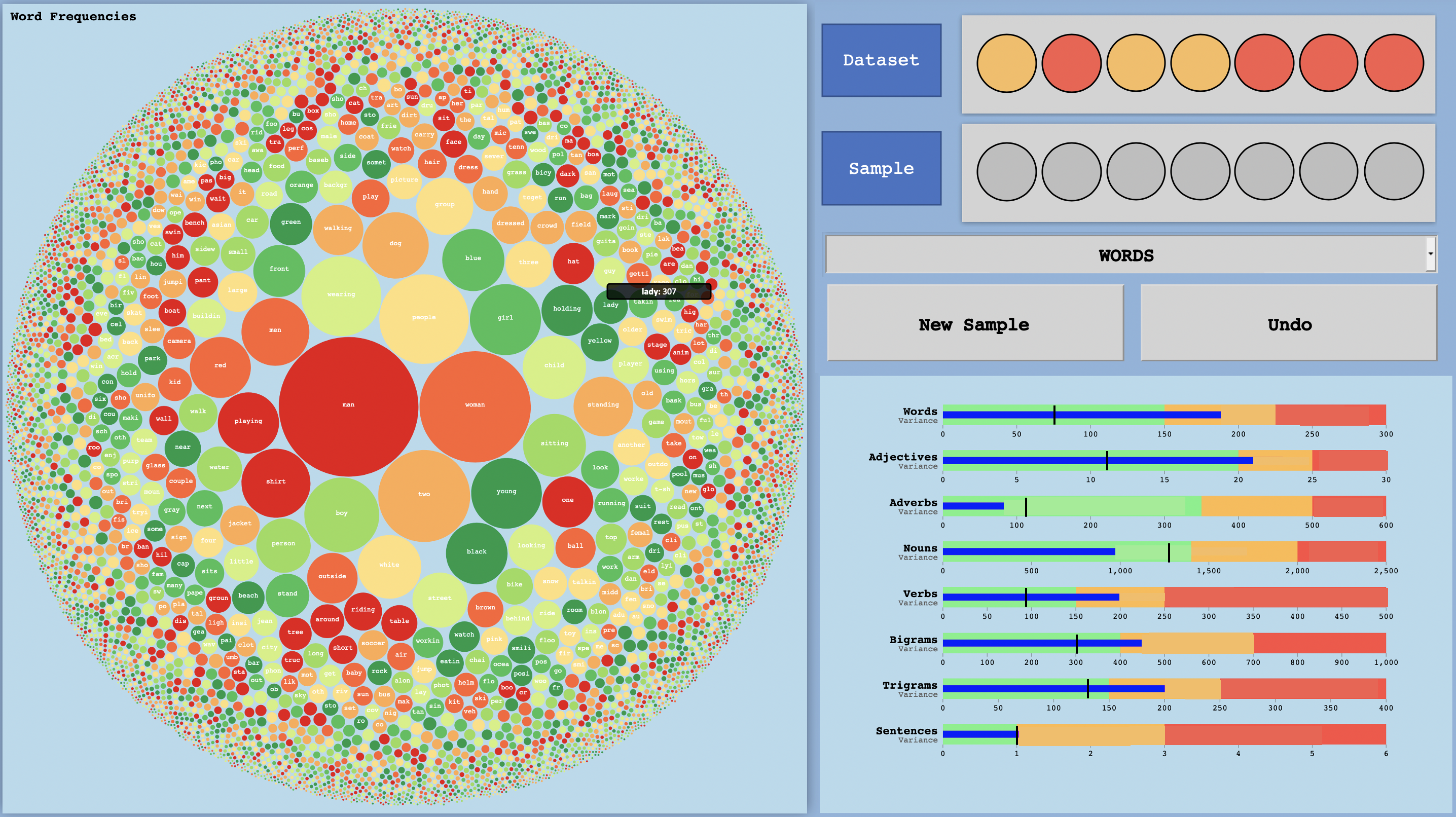}
  \caption{$DQI_{c2}$ Visualization Prior to New Sample Addition}
  \label{fig:Vis2before}
\end{figure*}
\subsection{Inter-sample N-gram Frequency and Relation}
\begin{figure*}
\includegraphics[width=\linewidth,height=8.55cm]{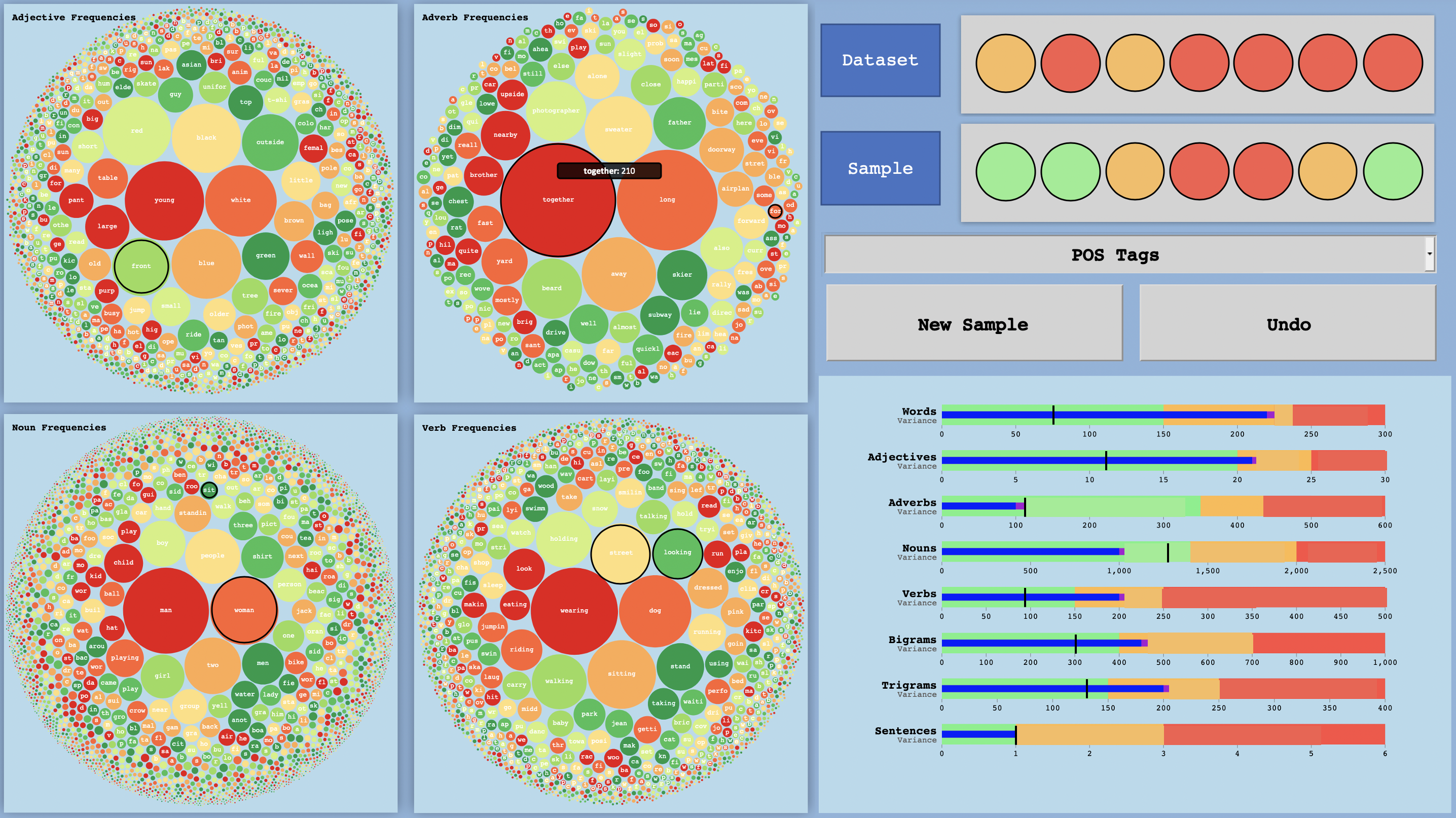}
  \caption{$DQI_{c2}$ Visualization On New Sample Addition}
  \label{fig:Vis2after}
\end{figure*}

\paragraph{Which Characteristics of Data are Visualized?}
There are different granularities of samples that are used to calculate the values of this component, namely: words, POS tags, sentences, bigrams, and trigrams. The granularities' respective frequency distributions and standard deviations are utilized for this calculation.
\paragraph{Bubble Chart for visualizing the frequency distribution:}
A bubble chart is used to visualize the frequency distribution of the respective granularity. This design choice is made in order to clearly view the contribution made by a new sample when added to the existing dataset in terms of different granularities. The bubbles are colored according to the bounds set for frequencies by the hyperparameters, and sized based on the frequency of the elements they represent. Additionally, some insight into variance can be obtained from this chart, by observing the variation in bubble size. 
\paragraph{Bullet Chart for impact of new sample:}
The impact of sample addition on standard deviation can be viewed using the bullet chart. The red-yellow-green color bands for each granularity represent the standard deviation bounds of that granularity. The vertical black line represents the ideal value of the standard deviation of that granularity. The two horizontal bars represent the value of standard deviation before and after the new sample's addition. Figure \ref{fig:Vis2before} illustrates the visualization.
\paragraph{Interactions:}
A tooltip, buttons, and a drop down are used for interactions. The tooltip displays the quantities in both charts on mouseover, and the buttons/drop down are used to update the chart. The following tasks are supported by the latter.
\begin{itemize}[noitemsep]
\item \textbf{Changing Granularity \textit{(Drop Down)}:} The drop down menu is used to select the granularity of the bubble chart displayed, as shown in Figure \ref{fig:Vis2before}.
\item \textbf{Addition of a New Sample \textit{(New Sample)}:} The new sample is added to the dataset, and an updated bubble chart of the word frequency distribution is generated. The new words that are added/ existing words that are updated are highlighted with thick black outlines in the chart. The granularity of the view can be changed using the drop down. The additions/modifications in the frequency distribution are similarly highlighted across all granularities, as illustrated in Figure \ref{fig:Vis2after}. The component value panels are updated as well. The previous state of the visualization is saved in a set of variables.
\item \textbf{Removal of a New Sample \textit{(Undo)}:} This reverses the operations of 'addition of a new sample' by using the saved state variables to restore the visualizations back to their original state.
\end{itemize}
\begin{figure*}
\includegraphics[width=\linewidth,height=8.55cm]{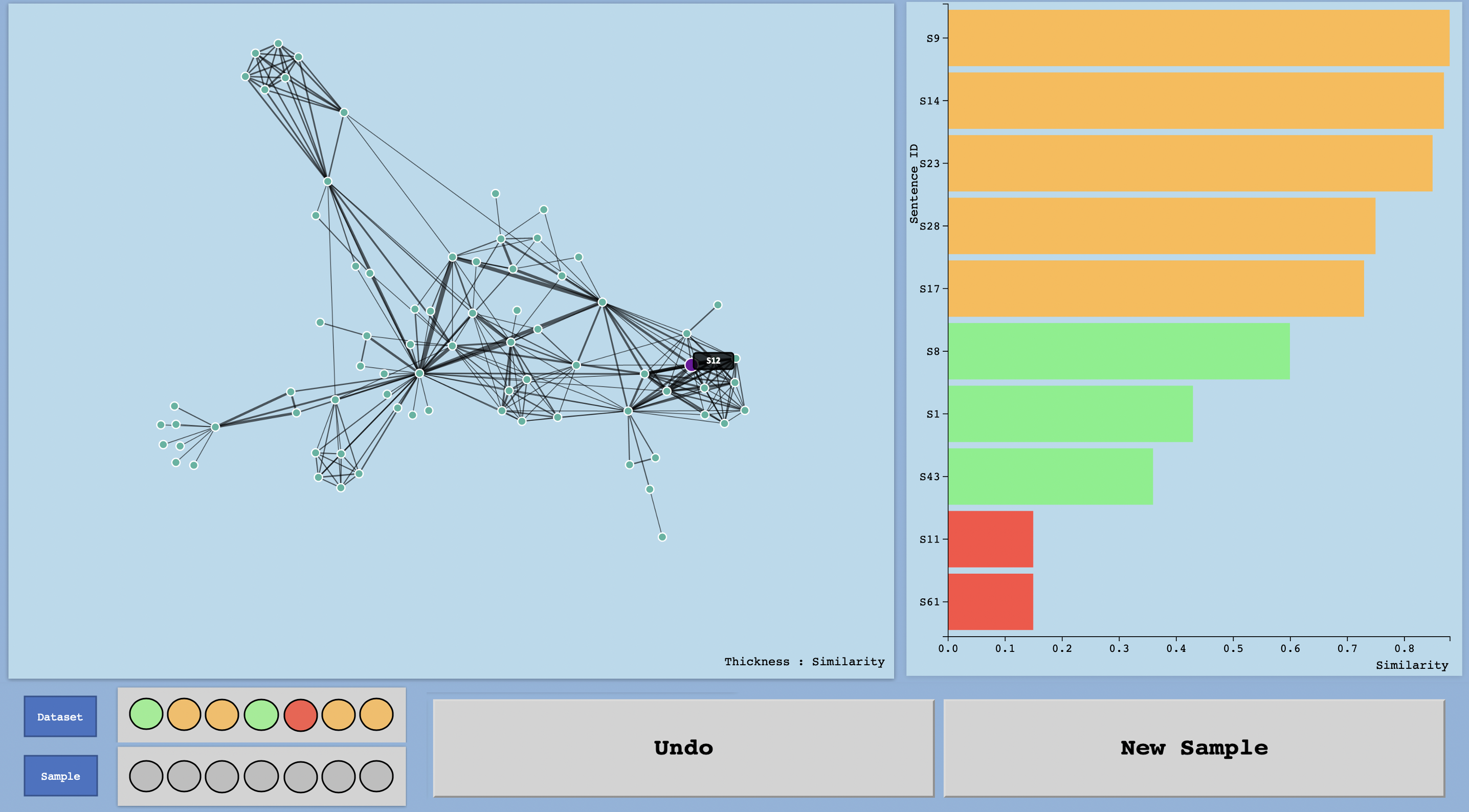}
  \caption{$DQI_{c3}$ Visualization Prior to New Sample Addition}
  \label{fig:Vis3before}
\end{figure*}
\begin{figure*}
\includegraphics[width=\linewidth,height=8.55cm]{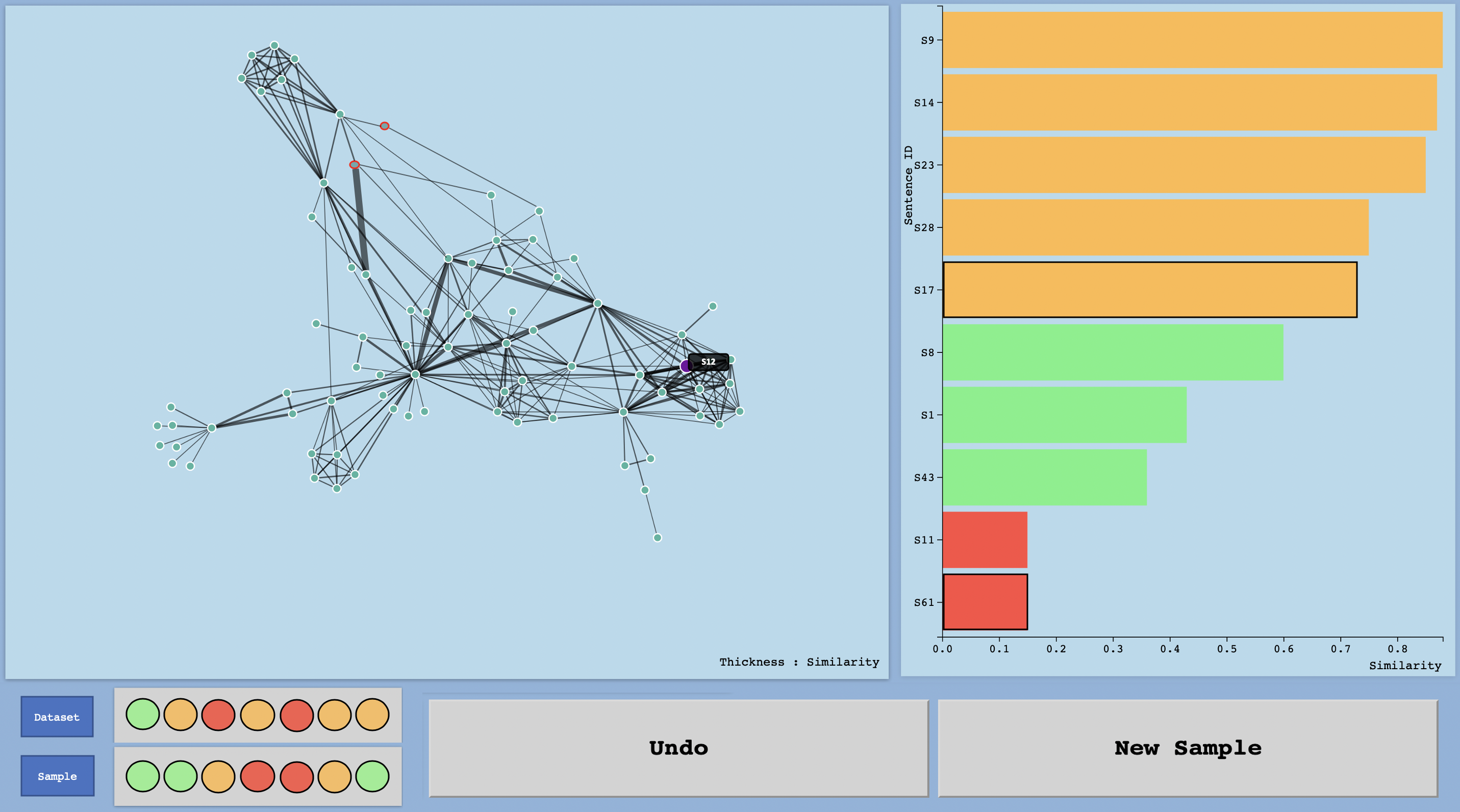}
  \caption{$DQI_{c3}$ Visualization On New Sample Addition}
  \label{fig:Vis3after}
\end{figure*}
\begin{figure*}
\includegraphics[width=\linewidth,height=8.55cm]{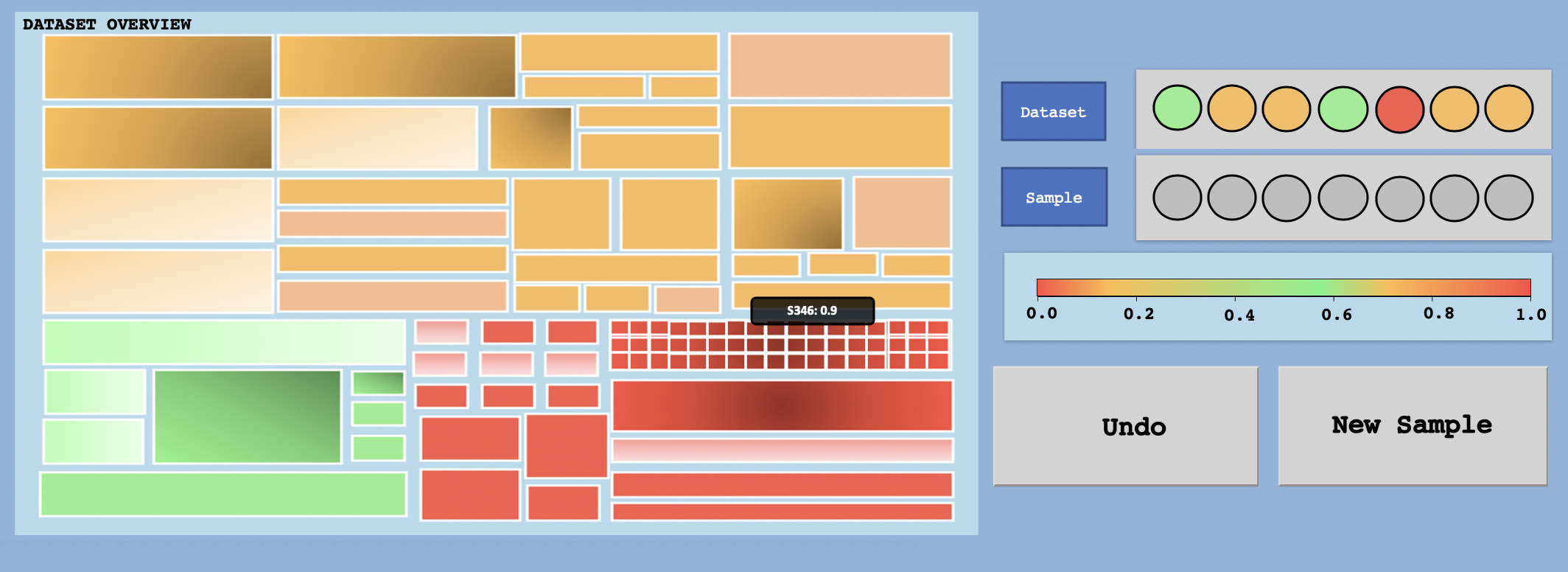}
  \caption{$DQI_{c4}$ Visualization Prior to New Sample Addition}
  \label{fig:Vis4before}
\end{figure*}
\subsection{Inter-sample STS}
\paragraph{Which Characteristics of Data are Visualized?}
The main units used in this DQI component are the similarity values between sentences across the dataset. This refers to either premise or hypothesis statements, relative to all other premise/hypothesis statements.  In order to understand the similarity relations of sentences, a force layout and horizontal bar chart are used. This is illustrated in Figure \ref{fig:Vis3before}.
%As a result, the number of sentences is twice the number of samples in the dataset.
\paragraph{Force Layout for Similar Sentence Pairs}
In the force layout, those sentence pairs with a similarity value that meets the minimum threshold are connected. Each node represents a sentence. The thickness of the connecting line depends on how close the similarity value is to the threshold. 
\paragraph{Horizontal Bar Chart for Most Similar Sentences}
In the horizontal bar chart, the sentences that are most similar to the given sentence are ordered in terms of their similarity value. The bar colors are centered around the threshold.
\paragraph{Interactions:}
Interactions via tooltip display the sentence id- i.e., the sample id, and whether the sentence is a premise/hypothesis of that sample- and similarity value in case of both the charts. The two charts are also linked on click of a node in the force layout. Other interactions are fuelled by buttons. The complete set of tasks is as follows:
\begin{itemize}[noitemsep]
    \item \textbf{Displaying Horizontal Bar Chart \textit{(on node click)}:} By selecting a node in the force layout, a horizontal bar chart is produced, that displays the ten most similar sentences to the sentence represented by the node. The benefits of the bar chart are two-fold. First, the bar chart  accounts for sentence links not present in the force layout. It displays those sentences whose similarity value is below the minimum threshold. This can help if certain sentences are isolated without links in the force layout. Second, it enhances the readability of information present in the force layout by drilling down on a subset, if the dataset size is very large.
    \item \textbf{Addition of a New Sample \textit{(New Sample)}:} The new sample is added to the dataset, and two new nodes are created in the force layout. The outline of these two nodes is in black, and by default, the premise is auto-selected to generate the bar chart. If the new sample's sentences appear in the bar chart for any other sample, then the outline of those bars is in black, as illustrated in Figure \ref{fig:Vis3after}. The component value panels are updated as well. The previous state of the visualization is saved in a set of variables.
    \item \textbf{Removal of a New Sample \textit{(Undo)}:} This reverses the operations of 'addition of a new sample' by using the saved state variables to restore the visualizations back to their original state.
\end{itemize}
\subsection{Intra-sample Word Similarity}
\begin{figure*}
\includegraphics[width=\linewidth,height=8.55cm]{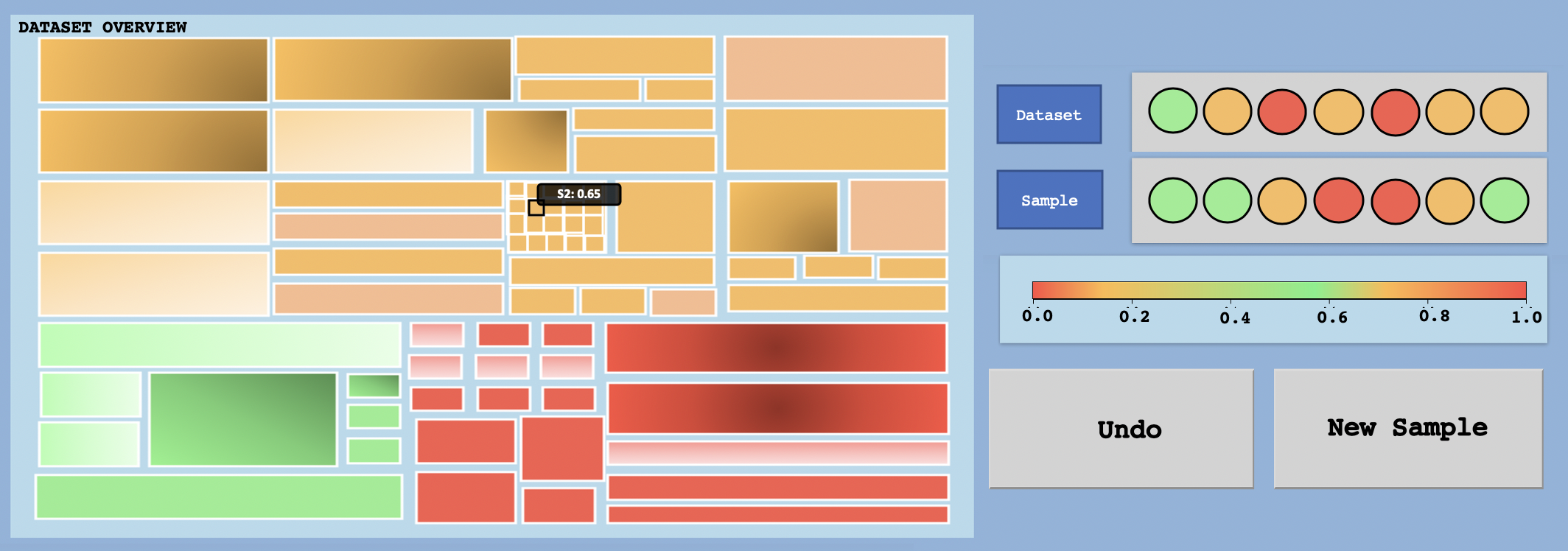}
  \caption{$DQI_{c4}$ Visualization On New Sample Addition: Dataset View}
  \label{fig:Vis4after}
\end{figure*}

\begin{figure*}
\includegraphics[width=\linewidth,height=8.55cm]{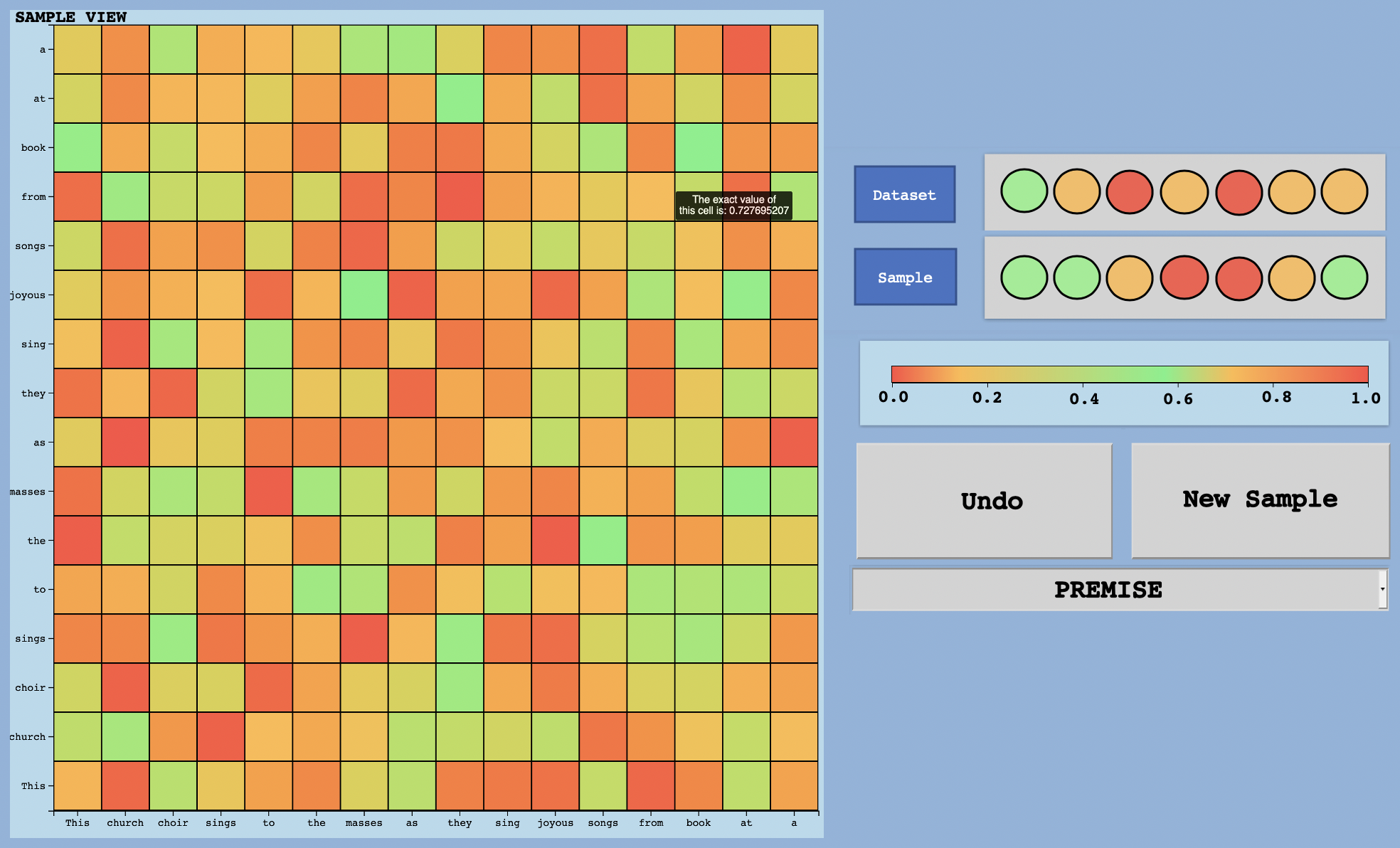}
  \caption{$DQI_{c4}$ Visualization On New Sample Addition: Sample View}
  \label{fig:Vis4component}
\end{figure*}

\paragraph{Which Characteristics of Data are Visualized?}
In this section, A sample's word similarity is viewed in terms of premise-only, hypothesis-only, and both. The relationship between non-adjacent words in the sample's sentences is analyzed specifically.
\paragraph{Overview Chart for Average Word Similarities and Heatmap for Single Sample}
The overview chart that is used is a one-level tree map, which uses the average value of all word similarities per sample- i.e., concatenated premise and hypothesis- to color and group its components. This is illustrated in Figure \ref{fig:Vis4before} The detailed view is a heat map of all the words in a single sample, ass shown in Figure \ref{fig:Vis4component}.
\paragraph{Interactions:}
Tooltips display the sample id for the tree map, and the similarity value between words for the heat map. Other interactions include a drop down used to select the sentence to be viewed in the heat map, linking the heat map to the tree map on click, and buttons to modify the visualizations. The tasks are as follows:
\begin{itemize}[noitemsep]
\item \textbf{Displaying Heat Map \textit{(on Tree Map click)}:} By clicking on a box of the tree map, the user is shown the heat map of the  clicked on sample. 
\item \textbf{Displaying the Tree Map \textit{(on Heat Map click)}:} By clicking anywhere on the heat map, the user is taken back to the tree map view.
\item \textbf{Addition of a New Sample \textit{(New Sample)}:} The new sample is added to the dataset, and a new box is added to the tree map, with a black outline to highlight it, as illustrated in Figure \ref{fig:Vis4after}. The component value panels are updated as well. The previous state of the visualization is saved in a set of variables.
\item \textbf{Removal of a New Sample \textit{(Undo)}:} This reverses the operations of 'addition of a new sample' by using the saved state variables to restore the visualizations back to their original state.
\item \textbf{Change Heat Map View \textit{(Drop Down)}:} Using the drop down, the heatmap can be changed to show word similarities for the (a) premise, (b) hypothesis, or (c) both sentences.
\end{itemize}
\begin{figure*}
\includegraphics[width=\linewidth,height=8.55cm]{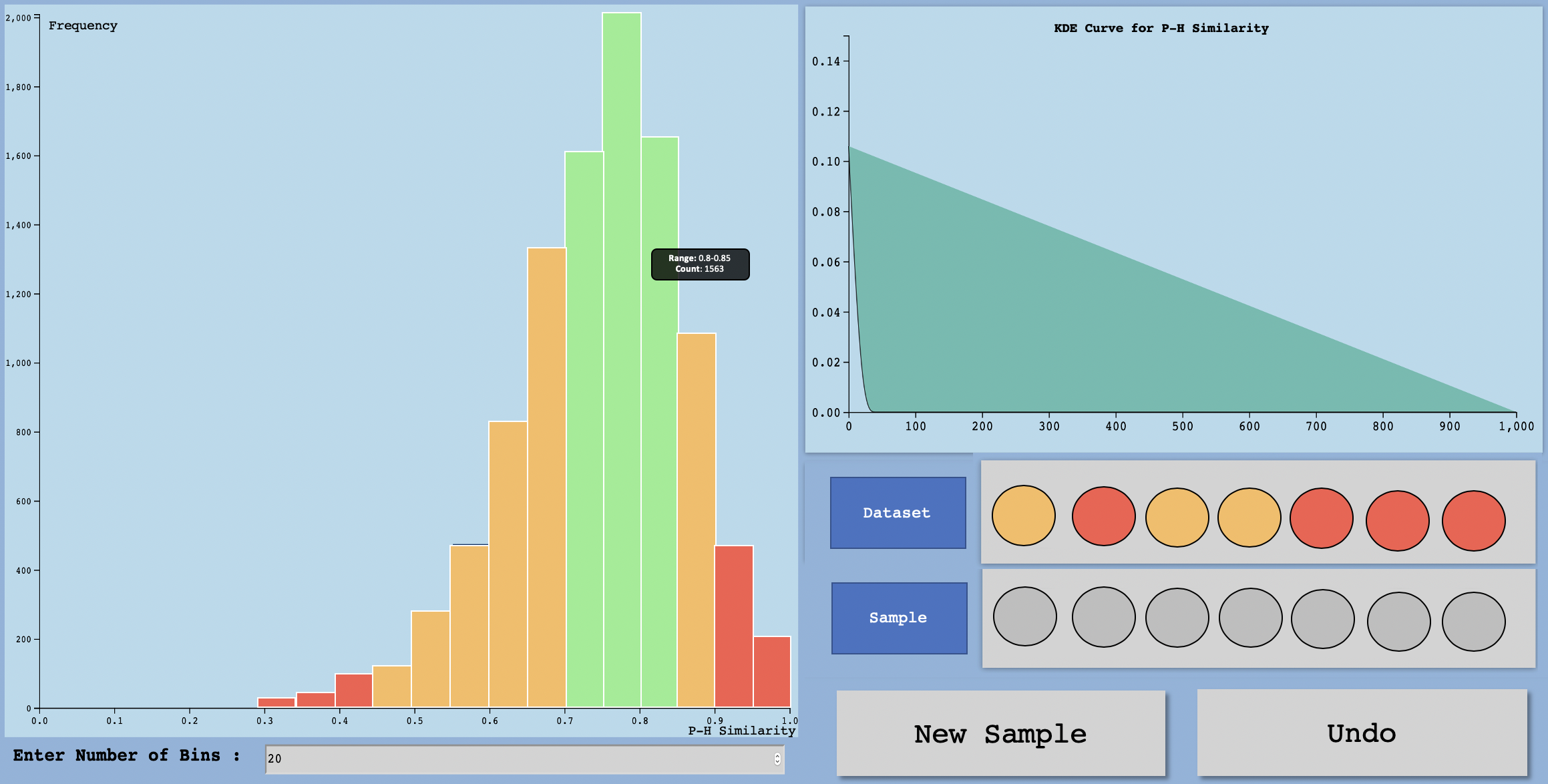}
  \caption{$DQI_{c5}$ Visualization Prior to New Sample Addition}
  \label{fig:Vis5before}
\end{figure*}
\subsection{Intra-sample STS}
\begin{figure*}
\includegraphics[width=\linewidth,height=8.55cm]{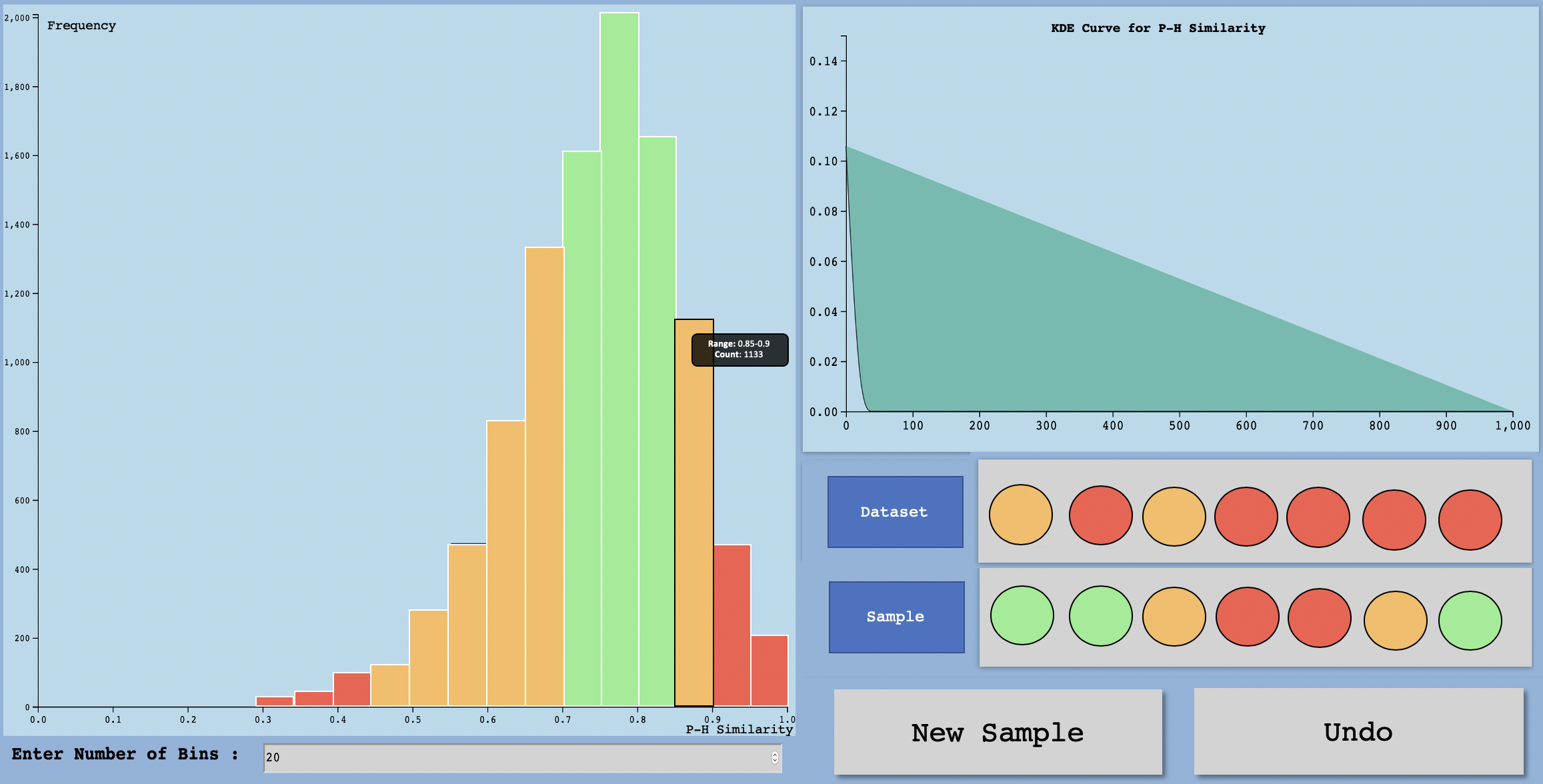}
  \caption{$DQI_{c5}$ Visualization On New Sample Addition}
  \label{fig:Vis5after}
\end{figure*}

\paragraph{Which Characteristics of Data are Visualized?}
Premise-Hypothesis similarity is analyzed on the basis of length variation, meeting a minimum threshold, and similarity distribution across the dataset. The first is addressed already in the vocabulary property by viewing the sentence length distribution. The other two are visualized using a histogram and kernel density estimation curve, as shown in Figure \ref{fig:Vis5before}.
\paragraph{Histogram and Kernel Density Curve for Sample Distribution}
The histogram represents the distribution of the samples, and is colored by centering around the threshold as the ideal value. The number of bins can be changed, and therefore multi-level analysis can be conducted. The kernel density curve is used to check for the overall skew of the distribution.
\paragraph{Interactions:}
Tooltips on the histogram display the number of samples per bin. Buttons and a text box are used for implementing other interactions:
\begin{itemize}[noitemsep]
    \item \textbf{Re-binning Histogram \textit{(textbox)}:} By filling a new value in the textbox, the number of bins in the histogram changes to that value.
    \item \textbf{Addition of a New Sample \textit{(New Sample)}:} The new sample is added to the dataset, the histogram and density plot are updated accordingly. The bar in the histogram to which the sample contributes is outlined in black across all histogram binnings, as illustrated in Figure \ref{fig:Vis5after}. The component value panels are updated as well. The previous state of the visualization is saved in a set of variables.
    \item \textbf{Removal of a New Sample \textit{(Undo)}:} This reverses the operations of 'addition of a new sample' by using the saved state variables to restore the visualizations back to their original state.
\end{itemize}
\begin{figure*}
\includegraphics[width=\linewidth,height=8.55cm]{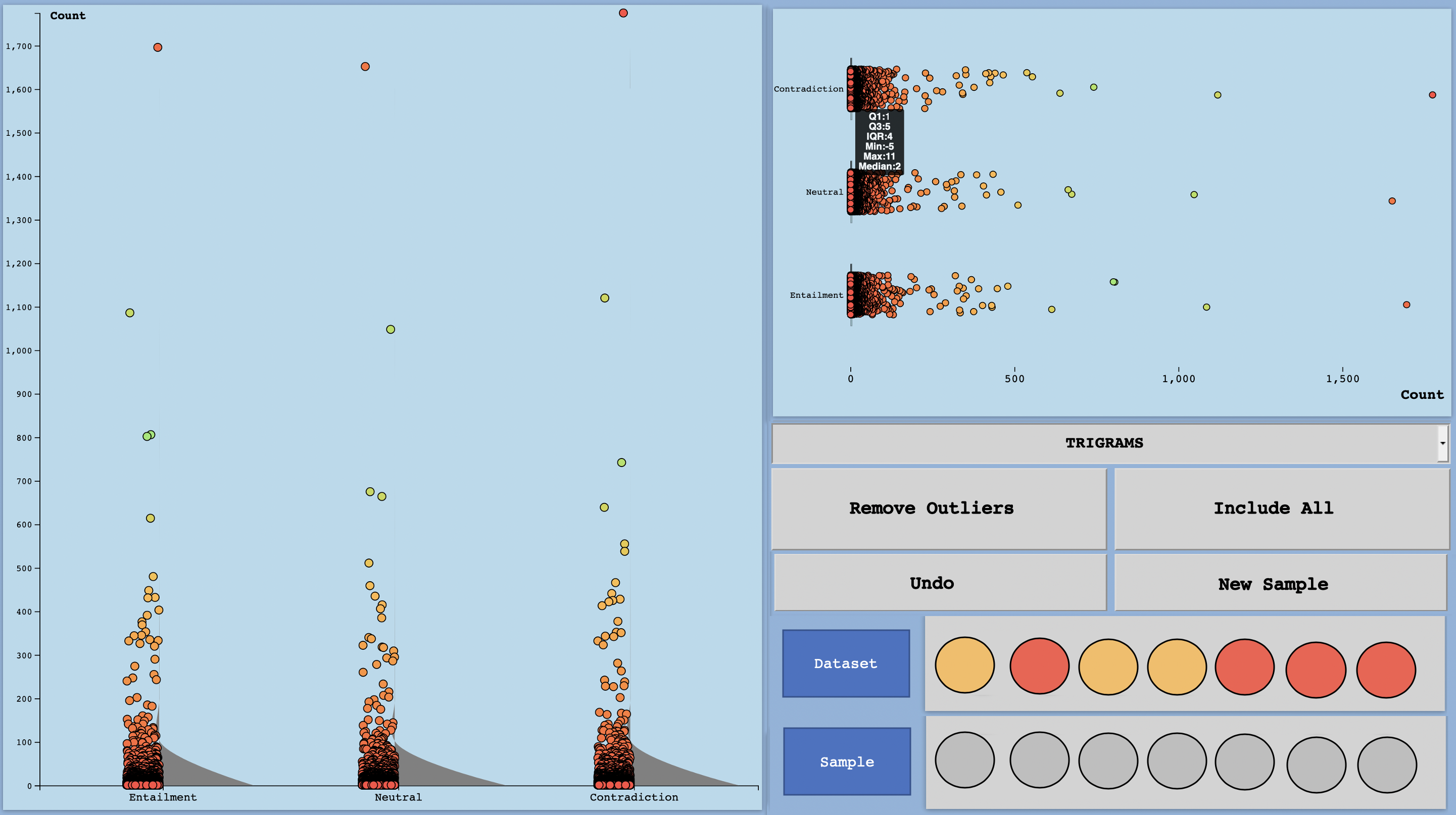}
  \caption{$DQI_{c6}$ Visualization Prior to New Sample Addition}
  \label{fig:Vis6before1}
\end{figure*}
\begin{figure*}
\includegraphics[width=\linewidth,height=8.55cm]{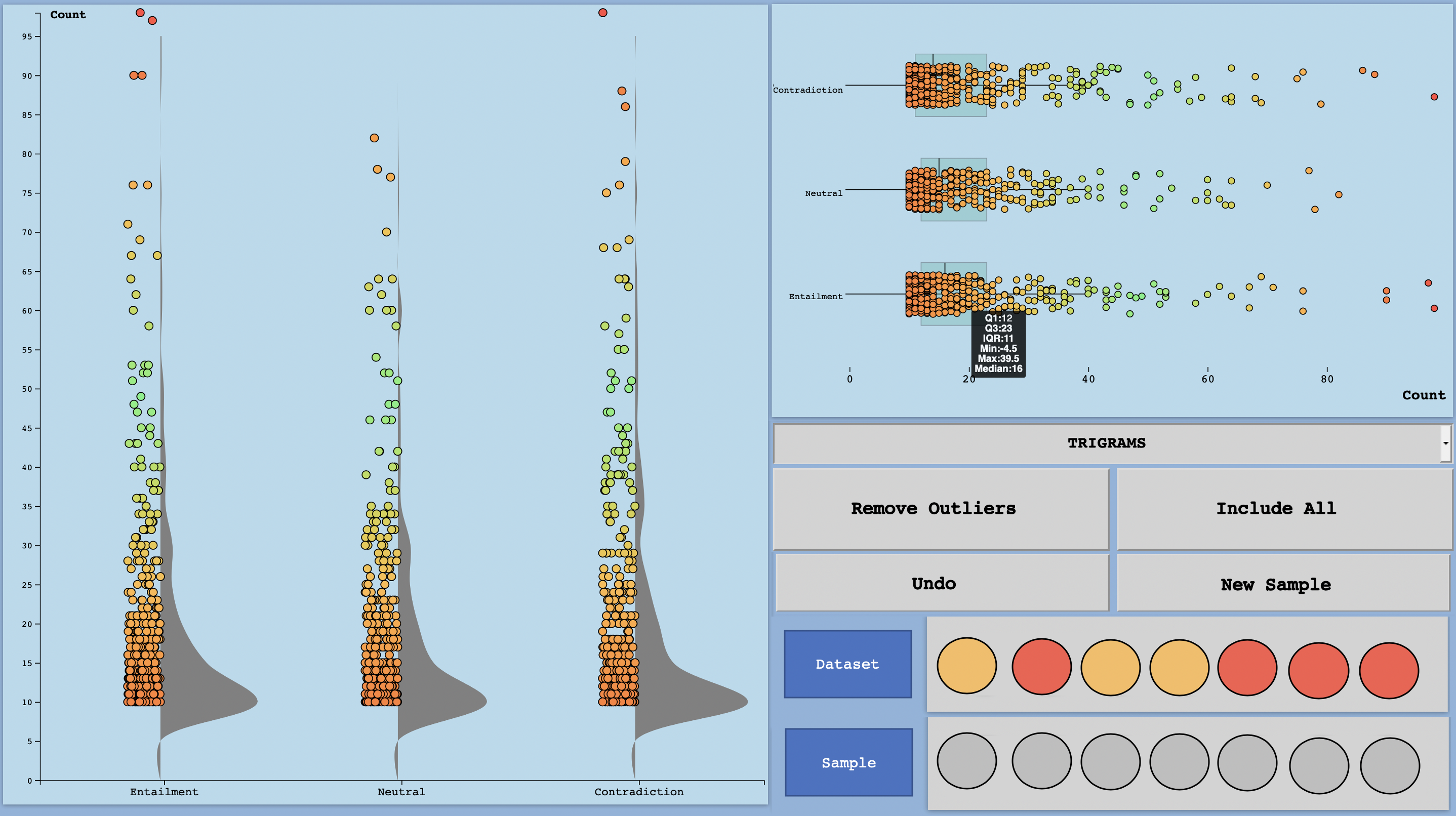}
  \caption{$DQI_{c6}$ Visualization after removing outliers Prior to New Sample Addition}
  \label{fig:Vis6before2}
\end{figure*}
\subsection{N-Gram Frequency per Label}
\begin{figure*}
\includegraphics[width=\linewidth,height=8.55cm]{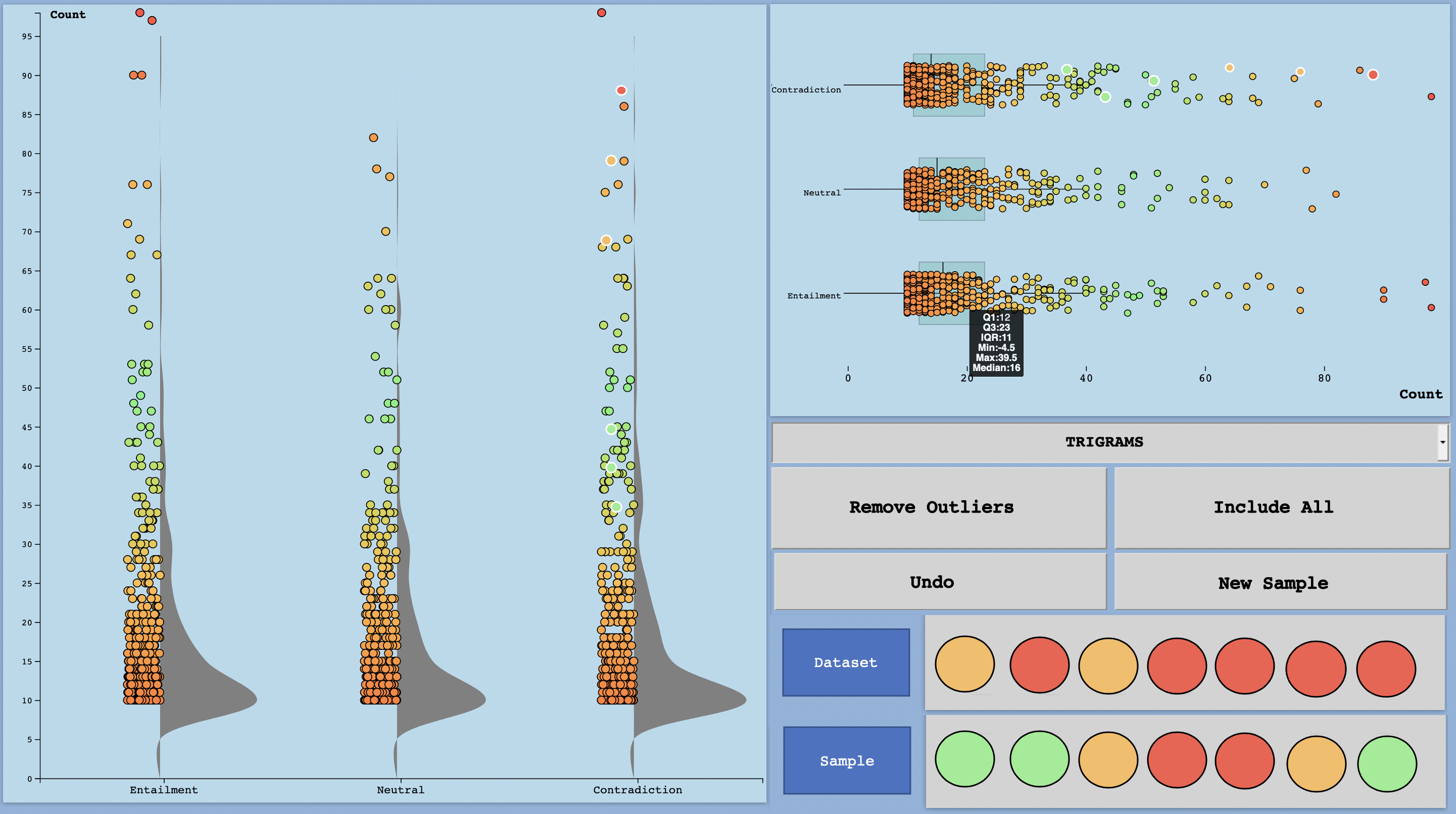}
  \caption{$DQI_{c6}$ Visualization On New Sample Addition}
  \label{fig:Vis6after1}
\end{figure*}

\begin{figure*}
\includegraphics[width=\linewidth,height=8.55cm]{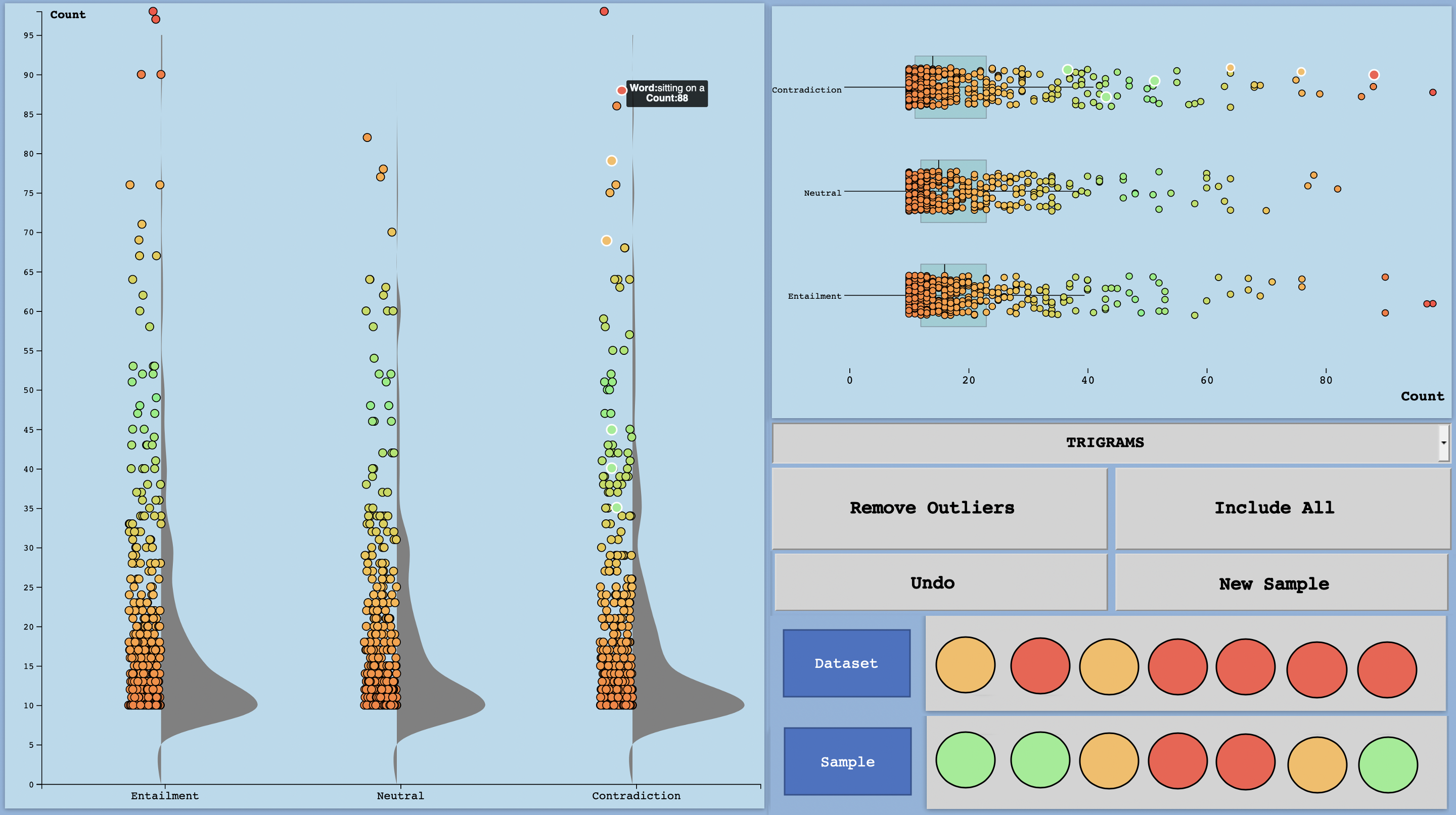}
  \caption{$DQI_{c6}$ Visualization with mouseover On New Sample Addition}
  \label{fig:Vis6after2}
\end{figure*}

\paragraph{Which Characteristics of Data are Visualized?}
This component drills down on the second component, to view the patterns seen in granularities per label. There are two small multiples charts, divided based on label, used in this view- a violin plot and a box plot.
\paragraph{Violin plot and Kernel Density Curve for Skew of Distribution:}
The violin plots are structured to display both jittered points, according to their frequency distribution, as well as a kernel density curve to judge the skew of the distribution. The points each represent an element of the granularity. 
\paragraph{Box Plots for More Information}
The box plots are used to garner more information about the distribution, in terms of its min, max, median, mean, and inter quartile range. These help further characterize the distribution, as well as provide a quantitative definition of the skew seen using density curves. Jittered points representing elements are present in this plot as well.
\paragraph{Interactions:}
On mouseover of a point in both visualizations, the element and its frequency are displayed in a tooltip. Other interactions are based on a dropdown and buttons as follows:
\begin{itemize}[noitemsep]
\item \textbf{Changing Granularity \textit{(Drop Down)}:} The drop down menu is used to select the granularity of the plots displayed, as shown in Figure \ref{fig:Vis6before1}. This granularity can be in terms of words, POS tags, bigrams, trigrams, or sentences.
\item \textbf{Addition of a New Sample \textit{(New Sample)}:} The new sample is added to the dataset, and updated plots of the word frequency distribution are generated. The new words that are added/ existing words that are updated are highlighted with thick white outlines in the chart. The granularity of the view can be changed using the drop down. The additions/modifications in the frequency distribution are similarly highlighted across all granularities. This is shown in Figure \ref{fig:Vis6after1} and \ref{fig:Vis6after2} .The component value panels are updated as well. The previous state of the visualization is saved in a set of variables.
\item \textbf{Removal of a New Sample \textit{(Undo)}:} This reverses the operations of 'addition of a new sample' by using the saved state variables to restore the visualizations back to their original state.
\item \textbf{Outlier Handling \textit{(Remove Outliers)}:} This removes elements with frequency counts less than the median to get a less skewed picture of the remainder of the distribution.  The component value panels are updated as well, as illustrated in Figure \ref{fig:Vis6before2}. The previous state of the visualization is saved in a set of variables.
\item \textbf{Full Distribution View \textit{(Include All Samples)}:} This reverses the operations of 'outlier handling' by using the saved state variables to restore the visualizations back to their original state.
\end{itemize}
\begin{figure*}
\includegraphics[width=\linewidth,height=8.55cm]{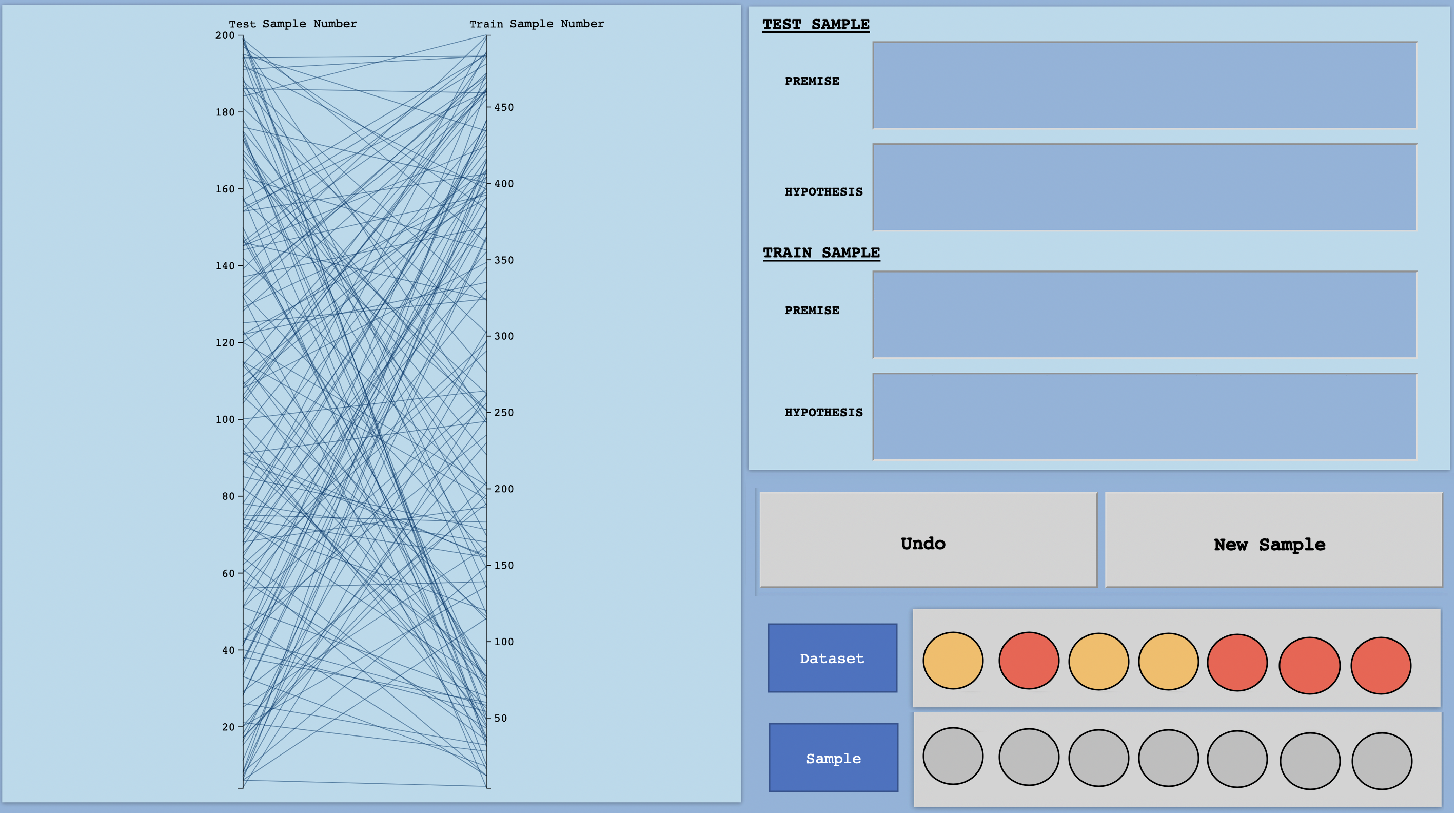}
  \caption{$DQI_{c7}$ Visualization Prior to New Sample Addition}
  \label{fig:Vis7before}
\end{figure*}
\begin{figure*}
\includegraphics[width=\linewidth,height=8.55cm]{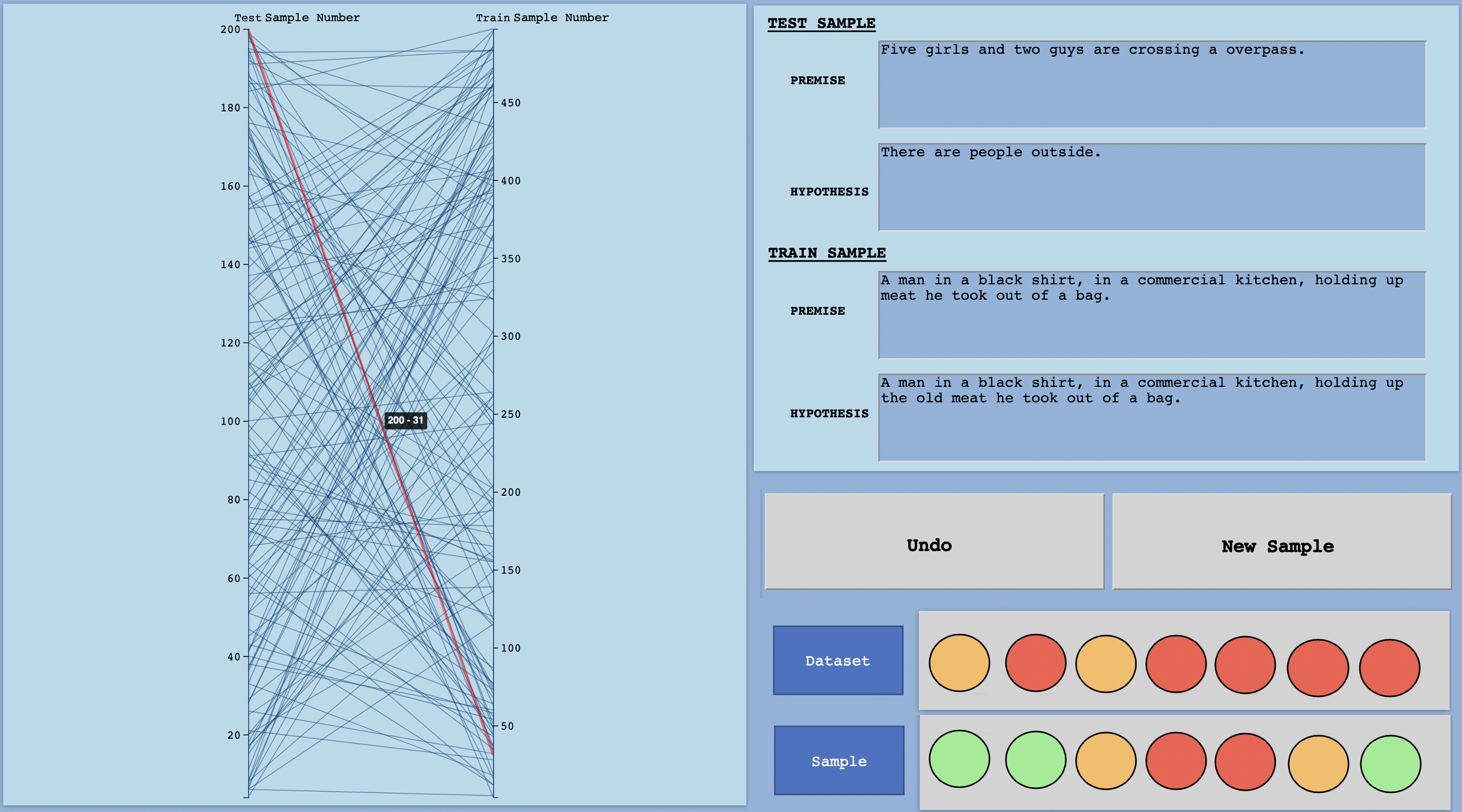}
  \caption{$DQI_{c7}$ Visualization On New Sample Addition}
  \label{fig:Vis7after}
\end{figure*}
\subsection{Inter-split STS}
\paragraph{Which Characteristics of Data are Visualized?}
Train-Test similarity must be kept minimal to prevent data leakage. This component's main feature is finding the train split sample that is most similar to a given test split sample.
\paragraph{Parallel Coordinate Graph for Train-Test Similarity:} A subset of test and train samples, all found to have close similarity within their respective splits, and significant similarity across the splits are plotted as a one step parallel coordinate graph, with test samples along one axis, and train samples along the other. This subset is seeded with those samples closest in similarity to the new sample to be introduced, based on the third component's visualization. The links connecting points on the two axes are drawn between the most similar matches across the split, as shown in Figure \ref{fig:Vis7before}. 
\paragraph{Interactions:}
Interactions include a tooltip that displays the sample ids connected on mouseover of a link, text boxes filled on click of a link, and other tasks by buttons:
\begin{itemize}[noitemsep]
\item \textbf{Details of Linked Pair \textit{(on click of link)}:} Clicking on a link causes the link to turn red, and the premises and hypotheses of the two samples are displayed in the text boxes on the screen. Clicking on another link changes the values of the textboxes, and highlights only the new link.
\item \textbf{Addition of a New Sample \textit{(New Sample)}:} The new sample is added to the dataset, and the sample is added to the axis of the parallel coordinates plot depending on the split that it belongs to, as determined by the component one visualization. The sample's link is auto-selected and the textboxes are accordingly updated. The component value panels are updated as well, as illustrated in Figure \ref{fig:Vis7after}. The previous state of the visualization is saved in a set of variables.
\item \textbf{Removal of a New Sample \textit{(Undo)}:} This reverses the operations of 'addition of a new sample' by using the saved state variables to restore the visualizations back to their original state.
\end{itemize}
\section{User Interface}
\paragraph{UI for Data Creation and Valiation:}
The UI design is two-fold. It targets two aspects of data creation- crowd source worker creation, and analyst review. The first phase uses colored flags to provide feedback to a crowd source worker about the quality of the sample they have created, so that they can fix it manually/with autofix assistance before submitting for higher return. The second phase uses the data visualizations discussed in section \ref{dqivizl} to help the analyst determine if the sample should be added, rejected, or fixed.
\subsection{Crowd-Source Worker:}
The design choices made are heavily focused on the notion of providing simple, yet critical feedback to the crowd source worker, to enhance the quality of data created by means of minimizing spurious bias. The methods and principles used in building the interface used for SNLI's \cite{bowman2015large} data collection process are the basis of our interface design. There are two types of feedback given in the UI, pre-submission and post-submission of the sample.
\paragraph{Instructions}
A sliding panel instruction tab is on the left corner of the screen. It consists of two sets of instructions. The first set goes over all general interface functionality descriptions, including post-submission user feedback. The second set specifically focuses on the pre-submission feedback loop.

\begin{figure*}
\includegraphics[width=\linewidth,height=8.55cm]{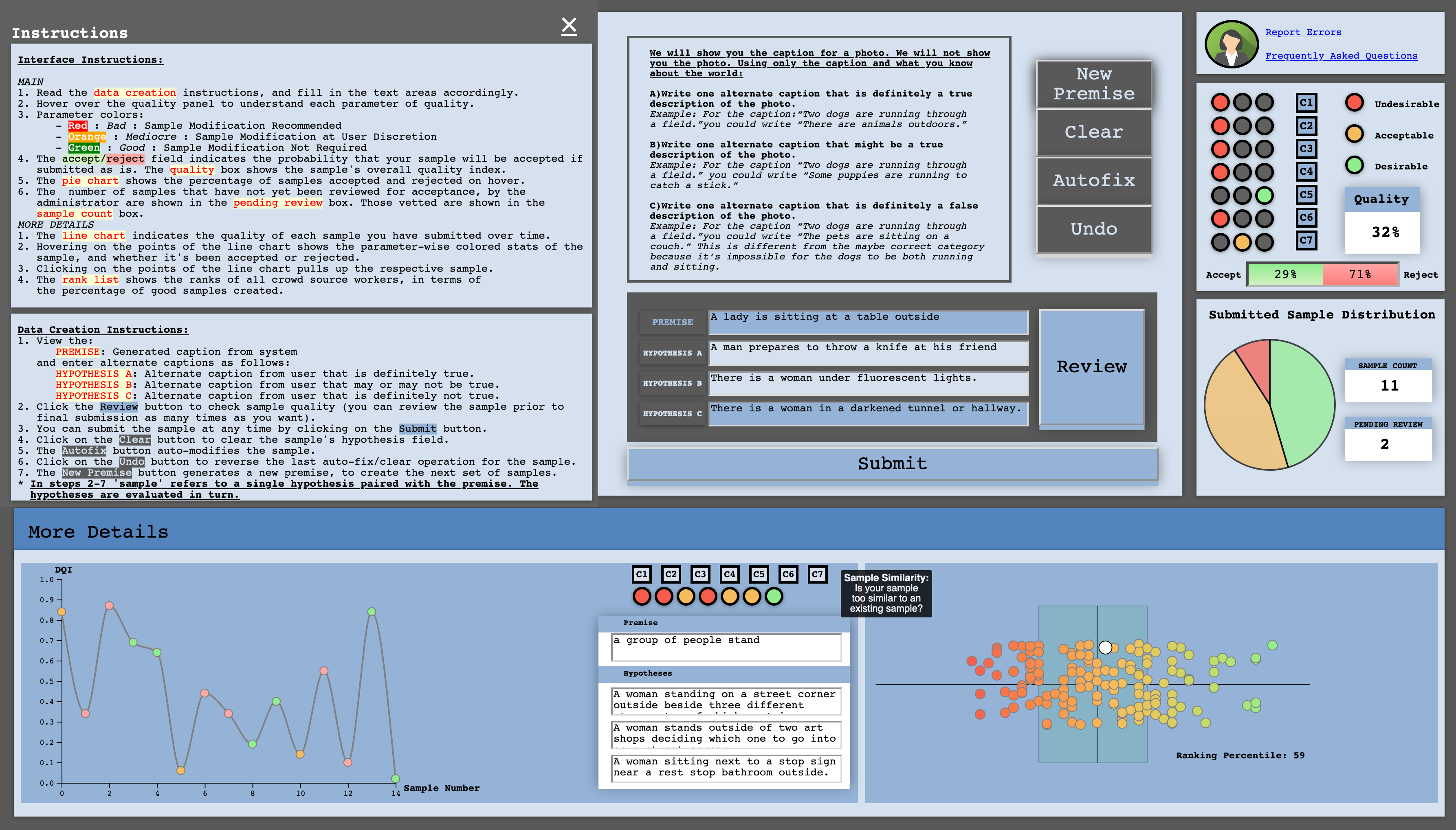}
  \caption{Crowd Source Worker View}
  \label{fig:csview}
\end{figure*}

\paragraph{Pre-Submission Feedback Loop:}
After reviewing the main instruction panel, the user can begin data creation. There is an instructions box displayed at all times on the main creation panel, which gives examples used in the original SNLI interface design, to make users understand the nature of the samples they are required to create. The premise field is auto-filled with captions from the Flickr30k corpus. This field can be changed to a fresh premise at any time by clicking on the 'new premise' button. The 3 types of hypothesis (entailment, neutral, and contradiction) must be entered in their respective fields. 
\paragraph{DQI based on past history}
Following this, each hypothesis is evaluated individually with the premise. Henceforth, the use of the term sample denotes premise and only the hypothesis under consideration. The hypothesis under consideration can be cleared at any time by clicking the 'clear' button. The user must click the 'Review' button at least once before submitting. The 'Review' button populates the DQI indication panel, which displays the values of the DQI components with respect to both the newly created sample and the existing set of accepted samples. The general aspect of data that is being analyzed by a component can be viewed on a tooltip, on mouseover of the component label. The messages displayed are as follows:

\begin{itemize}
    \item Vocabulary: Does your sample contribute new words?
    \item Combinations: Does your sample contribute new combinations of words and phrases?
    \item Sentence Similarity: How similar is your hypothesis to all other premises or hypotheses?
    \item Word Similarity: How similar are all the words within your sample?
    \item PH Score: How similar is your hypothesis to the premise?
    \item Label Giveaway: Is your hypothesis too obvious for our system?
    \item Sample Similarity: Is your sample too similar to an existing sample?
\end{itemize}
\paragraph{Feedback Flags}
The values of the DQI components are indicated using a traffic signal analogy (red, yellow, and green), thereby indicating if a particular aspect of the data created might lead to bias. The colors respectively advise the user to stop, revise, and proceed in their sample creation tactics. The probability of the newly created sample being accepted/rejected is also displayed. Based on this feedback, the user can choose to: (i) manually fix their sample and review it again, (ii) 'auto-fix' the sample by paraphrasing it using concept net, (iii) submit the sample as is. Once the user is satisfied with the sample created, they can submit the sample. Once the sample has been submitted, the 'pending review' box is accordingly updated, as is the 'count' box for total number of submitted samples.

\paragraph{Post-Submission Feedback Loop:}
 We retain the notion of a background expert reviewing samples to ensure that the sentences use appropriate ideas and language. Once the analyst reviews the sample and marks it as accepted/rejected (see section 8.2), the following updates occur on the crowdsource worker's UI \footnote{these updates are only loaded at the start of each new user login session} :
\begin{itemize}
    \item The line chart on the secondary panel indicates the quality of the user's submitted samples over time. It is color coded according to whether the sample was accepted or rejected. On hovering over any one sample, the quality level of that sample are displayed on a tooltip. On click the sample appears in a text box.
    \item The 'pending review' box count on the main panel is decremented by one.
    \item The ranks are displayed using a box plot that calibrates ranks based on the percentage of accepted samples created by each user.
    \item The pie chart on the main panel is updated according to the accept/reject percentages.
\end{itemize}
\paragraph{Additional Communication Links:}
There are additional FAQ and Reporting Problem links present in the interface. The FAQs deal with data creation guidelines, and the Reporting Problems form is intended for technical issues only. This is in accordance with similar functionalities from the original SNLI interface. Figure \ref{fig:csview} illustrates the crowdsource worker's UI.

\subsection{Analyst:}
\begin{figure*}
\includegraphics[width=\linewidth,height=8.55cm]{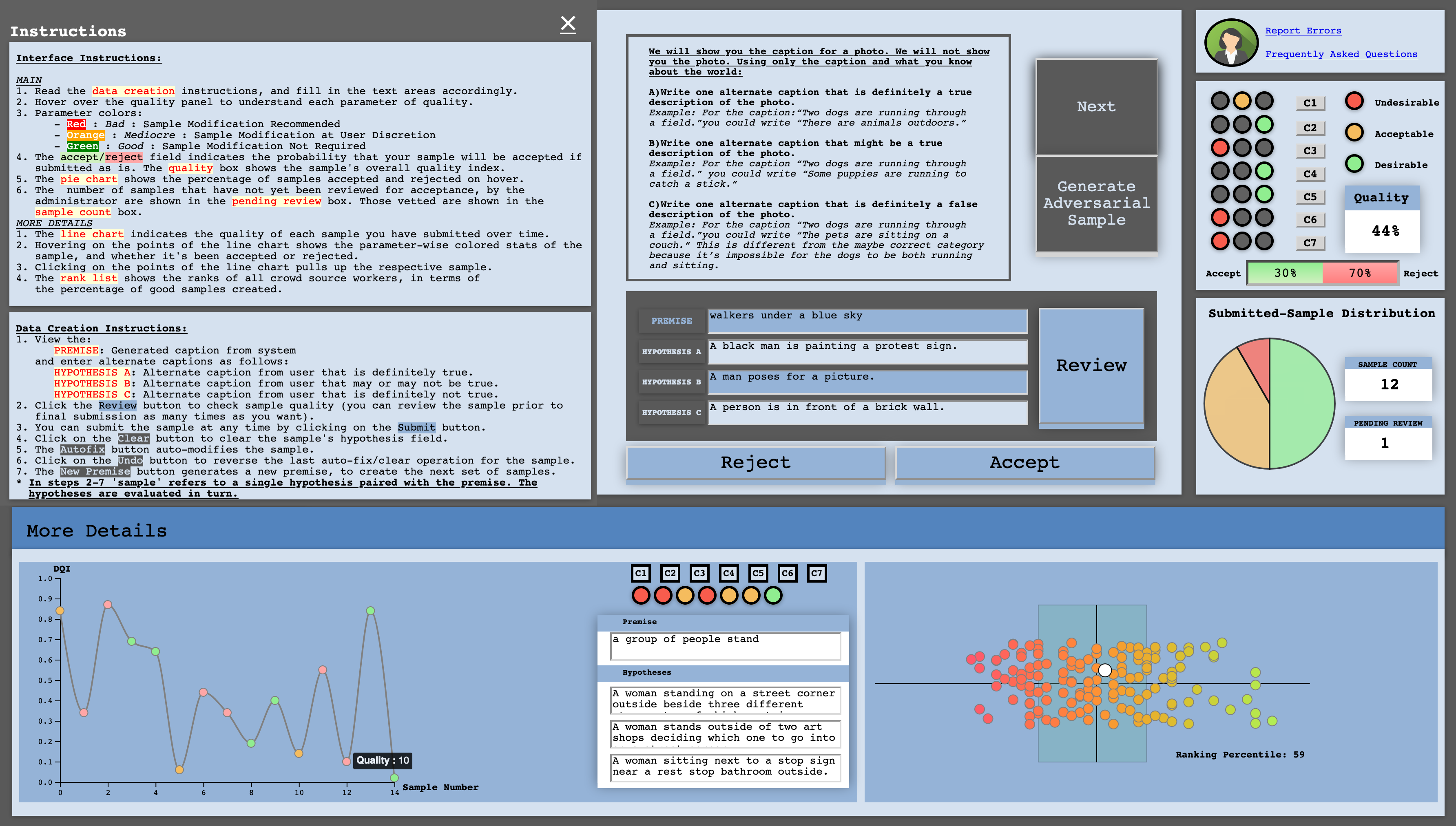}
  \caption{Analyst View}
  \label{fig:anaview}
\end{figure*}

\paragraph{Analysts' basic interface similar to crowdsource workers':}
The analyst interface is focused on the data validation process. The layout of the interface follows the same pattern as that of the crowd source workers interface. This is done so that the analyst understands the environment presented to the crowd source worker for data creation. The sliding panel for instructions, data entry boxes, DQI indication panel, and communication links are retained as is. The piechart, count box, pending review box, line chart, and rank box plot change depending on the annotator id associated with the sample being evaluated, as they represent the performance of that particular annotator.
\paragraph{Review Button}
The 'Next' buttons loads the next created sample set that must be reviewed. The text fields are filled with the premise and all hypotheses statements matching that premise. On clicking 'Review', the analyst reviews each hypothesis paired with the premise individually, as done in the crowdsource worker interface.
\paragraph{Buttons for Appropriate Visualizations:}
The DQI indication panel has buttons that link to each component's respective visualization, as outlined in section \ref{dqivizl}. There are buttons present instead of labels for each component in this panel that can be used to navigate to each visualization in turn. The sample considered in the visualizations as the 'new sample' is the sample that is under review.
\paragraph{Data Validation}
The 'Accept' button can be used to accept the sample as is, and causes the piechart, pending review box, count box, rank box plot, and line chart for the annotator of the sample to be updated. The 'Reject' button is used mainly to discard samples that contain obscenities, have incoherent/ungrammatical hypothesis statements, and have hypothesis statements of length less than three words. If the sample has low quality, but can be converted to a higher quality adversarial sample with some modification and resubmitted, the 'Generate Adversarial Sample' button sends the sample to Text-Fooler. Samples that are auto-fixed at the analyst end in this manner are displayed as the yellow slice of the pie chart. Crowdsource workers receive lesser rewards for these samples. Figure \ref{fig:anaview} illustrates this.

\section{Proving Efficacy}

Test cases have been developed to show the efficacy of DQI in our proposed data creation paradigm, with varying numbers of preexisting samples. We tune the hyperparameters proportionally, based on the dataset size. The value ranges for the DQI component colors are also set accordingly. DQI has been calculated for the following cases:

(i) No Preexisting Samples

(ii) 100 Preexisting Samples from the Good Split of the SNLI Test Set

In case (i), DQI of the new sample is calculated. In case (ii), first, DQI for the preexisting sample set is computed, as $x_{1}$. Then, the new sample is added and DQI is recalculated for the updated sample set, as $x_{2}$. The new samples, shown in Table \ref{tab:dqitc}, have been taken from a recent work on adversarial filtering, AFLite \cite{bras2020adversarial}.

Then, the difference $\Delta x = x_{1}-x_{2}  $ is calculated. On the main interface, the crowd source worker views the colors of DQI components corresponding to $\Delta x$. The analyst views $\Delta x$ as `Sample' and $x_{2}$ as `Dataset' component colors on the visualizations. 

\subsection{Case(i) - Addressing Cold Start}
Case (i) addresses the situation of cold-start for DQI. Unlike adversarial filtering algorithms, DQI can be used even with low data levels. In the situation of cold start, the component initialization is as follows:

\paragraph{Vocabulary:}
The first term is scaled appropriately as it takes the size of the dataset into account. The second term returns the standard deviation between the premise and hypothesis lengths. Since the third term defines upper and lower bounds on sentence length, it takes a value of one as long as the lengths of both the premise and hypothesis statements exceed three words, and zero if it is three words or less, as seen for sample 5 in Table \ref{tab:dqic1i}.

\paragraph{Inter-sample N-gram Frequency and Relation:}
Term 1 captures the inverse of standard deviation, and hence yields infinity in the case of POS tags, when a word with that POS tag does not occur at all, or only occurs once as standard deviation tends to zero. In some cases, the standard deviation can be zero, as seen in Table \ref{tab:dqic2i8} for trigrams, as each trigram occurs an equal number of times. High non-infinite values for term one are seen for bigrams and trigrams due to their balanced distributions in a sample, as in Table \ref{tab:dqic2i11}.

Sentences are seen to differ across samples in terms of the language used, and their length. Therefore, when setting the  upper and lower bounds of granularities for Term 2, standardizing the bounds for cold start fails in the case of POS tags, particularly adverbs, as in seen Tables  \ref{tab:dqic2i1} - \ref{tab:dqic2i12}. These bounds therefore need to be reset at cold start particular to the sample's language.

\paragraph{Inter-sample STS:}
The first term focuses on the standard deviation of similarity values that cross a threshold between all sentences. Since there is only one similarity value calculated, the value of Term 1, as in Table \ref{tab:dqic3i}, is set to that similarity value to prevent it from becoming infinity. The second term is always taken to have a value of 2, as there is no definite set threshold for taking a maximum.

\paragraph{Intra-sample Word Simlarity:}
The fourth component scales appropriately, as it takes the size of the dataset into account and can therefore be directly computed, as in Table \ref{tab:dqic3i}.

\paragraph{Intra-sample STS:}
The first term, in Table \ref{tab:dqic5i1}, deals with whether the Premise-Hypothesis similarity crosses a threshold. This scales as it takes dataset size into account, and can be calculated for different threshold values. The second and third terms, Table \ref{tab:dqic5i234}, involve the calculation of the mean and standard deviation of length difference between the premise and hypothesis. Therefore, the second term is directly computed, while the third is always zero, since only one value is present. The fourth term's value, in Table \ref{tab:dqic5i234}, also uses standard deviation and is therefore is directly taken to be the similarity between the premise and hypothesis, as only one value is calculated. The fifth and sixth terms look at word overlap and word similarity levels between the premise and hypothesis, and can be directly calculated. These are represented in Tables \ref{tab:dqic5i56} - \ref{tab:dqic5i56e}.

\paragraph{N-gram Frequency per Label:}
Since cold start only involves the text data of a single sample, the label of that sample is the only one with initialized values in $DQI_{C6}$. Table \ref{tab:dqic5i1} has Terms 1 and 2 of $DQI_{C6}$, as they are equivalent to the terms of $DQI_{C2}$ for the label of the new sample. These terms are set to zero for the other two labels. Table \ref{tab:dqic5i234} has Terms 3 and 4, which are the same as terms 2 and 3 of $DQI_{C5}$, and are only computed for the label of the new sample. Also, since the counts of all granularities are only initialized for a single label, the fifth term is set to zero for all samples.

\paragraph{Inter-split STS:}
Since $DQI_{C7}$ is calculated on the basis of the most similar training sample for every test set sample, it is not applicable to the case of cold start, as there is only one sample. Hence, its value is taken as zero.

\subsection{Case(ii)-Adding to the Test Good Split}
A 100 samples are taken at random from the good split of the SNLI Test set and $x_{1}$ is calculated. Then the new sample is added to the dataset. $x_{2}$ and $\Delta x$ are calculated. For all components, DQI values are calculated using the same hyperparameter values as those used for the full test set. The results, shown in Tables \ref{tab:44} - \ref{tab:60}, indicate the need for hyperparameter scaling. 
\paragraph{What requires Scaling?}
From tables \ref{tab:45} and \ref{tab:51-54}-\ref{tab:54}, we find that the hyperparameters used to set upper and lower bounds for POS tag frequencies across and within labels are the ones that require significant scaling. Additionally, we find that sentence, bigram, and trigram terms should be omitted when calculating the DQI until their overall frequencies and variance reach a certain threshold. This is because terms inversely proportional to the standard deviation of the distributions of those granularities are found to explode for lesser numbers of samples.
\subsection{Assigning Colors}
The new sample set has six samples removed by AFLite, that belong to the bad split of the Dev set, and six that are retained, i.e., that belong to the good split of the Dev set. In both case (i) and case (ii), we find that on adding samples to the existing dataset, there is no significant difference in the term/component values except in the cases of word overlap and word similarity, seen in T5 and T6 of $DQI_{C5}$. We observe that DQI component colors are correctly predicted 10/12 times on an average. Also, the change in $DQI_{C5}$ corresponding to word overlap and word similarity is as expected as per the findings of Bras et.al. \cite{bras2020adversarial}.

\begin{table*} 
\centering
\scriptsize
\resizebox{2.0\columnwidth}{!}{%
\begin{tabular}{lllll}
\hline
\textbf{Sample ID} & \textbf{Premise} & \textbf{Hypothesis} & \textbf{Label} & \textbf{Split} \\
\hline
\textbf{S1} &A woman, in a green shirt, &A woman is preparing to &contradiction& Dev-Bad\\&preparing to run on a treadmill.&sleep on a treadmill.&&\\
\textbf{S2} &The dog is catching a treat. &The cat is not catching a treat. &contradiction & Dev-Bad\\
\textbf{S3} &Three young men are watching &Three young men watching&neutral  & Dev-Bad\\&a tennis match on a&a tennis match on a screen&& \\&large screen outdoors.& outdoors, because their&&\\&&brother is playing. &&\\
\textbf{S4} &A girl dressed in a pink shirt,&A funny person in a shirt. &neutral & Dev-Bad\\& jeans, and flip-flops &&&\\&sitting down playing &&&\\&with a lollipop machine. &&&\\
\textbf{S5} &A man in a green apron &A man smiles. &entailment & Dev-Bad\\&smiles behind a food stand. &&&\\
\textbf{S6} &A little girl with a hat &The girl is wearing a hat. &entailment & Dev-Bad\\&sits between a woman’s feet &&&\\&in the sand in front of &&&\\&a pair of colorful tents.  &&&\\
\textbf{S7} &People are throwing tomatoes &The people are having a &entailment & Dev-Good\\&at each other.&food fight. &&\\
\textbf{S8} &A man poses for a photo in &The man is prepared&&\\&front of a Chinese building&for his photo. &entailment & Dev-Good\\ &by jumping.&&&\\
\textbf{S9} &An older gentleman &A man giving a speech. &neutral & Dev-Good\\&speaking at a podium.&&&\\
\textbf{S10} &A man poses for a photo in &The man has experience&neutral & Dev-Good\\&front of a Chinese building &in taking photos.&&\\&by jumping.&&&\\
\textbf{S11} &People are waiting in &People sit and wait for&contradiction & Dev-Good\\&line by a food vendor.&their orders at a nice&&\\&&sit down restaurant.&& \\
\textbf{S12} &Number 13 kicks a soccer &A player passing the&contradiction & Dev-Good\\&ball towards the goal during&ball in a soccer game.&& \\&children’s soccer game.&&&\\
 \hline

              &            &            &        &          
\end{tabular}%
}
\caption{Samples used for Test Cases}
\label{tab:dqitc}
\end{table*}

\begin{table}
\resizebox{1.0\columnwidth}{!}{%
  \begin{tabular}{lllll}
  \hline
     
    \multirow{2}{*}{\textbf{Sample}} &
      \multicolumn{3}{c}{\textbf{Terms}}&
      \multicolumn{1}{c}{\textbf{DQI C1}}\\
      &\textbf{T1}&\textbf{T2}&\textbf{T3}&  \\\hline
      
    \textbf{S1} & 0.0693 & 2.121 & 1.0000 & 2.1906  \\
    \textbf{S2} & 0.0396 & 0.7071 & 1.0000 & 0.7467  \\
    \textbf{S3} & 0.1089 & 2.1213 & 1.0000 & 2.2302  \\
    \textbf{S4} & 0.1188 & 7.7781 & 1.0000 & 7.8969  \\
    \textbf{S5} & 0.06930 & 5.6568 & 0.0000 & 0.0693  \\
    \textbf{S6} & 0.1188 & 11.3137 & 1.0000 & 11.4325  \\
    \textbf{S7} & 0.0594& 0.0000& 1.0000&0.0594  \\
    \textbf{S8} & 0.0792&4.9497&1.0000&5.0289  \\
    \textbf{S9} & 0.0693&1.4142&1.0000&1.4835  \\
    \textbf{S10} & 0.0891&4.9497&1.0000&5.0388  \\
    \textbf{S11} & 0.0990&2.8284&1.0000&2.9274  \\
    \textbf{S12} & 0.1089&2.8284&1.0000&2.9373  \\\hline
       
  \end{tabular}%
  }
 \caption{$DQI_{C1}$ for Case (i)}
 \label{tab:dqic1i}
\end{table}

\begin{table}
\resizebox{1.0\columnwidth}{!}{%
  \begin{tabular}{lllll}
  \hline
     
    \textbf{Granularity}&\textbf{Count}&\textbf{DQI C2,C6 - T1}&\textbf{DQI C2,C6 - T2}&\textbf{DQI C6 - T5}  \\\hline
      
    \textbf{Sentences} & 2 & 1.0000 & 1.0000&0 \\
    \textbf{Words} & 7 & 13.0958 & 1.0000&0\\
    \textbf{Adjectives} & 1 & inf & 1.0000&0\\
    \textbf{Adverbs} & 0 & inf & nan&0\\
    \textbf{Verbs} & 2 & 4.0000 & 1.0000&0\\
    \textbf{Nouns} & 4 & 8.0000 & 1.0000&0\\
    \textbf{Bigrams} & 15 & 32.7698 & 0.1578&0\\
    \textbf{Trigrams} & 16 & 64.0000 & 0.7647&0\\\hline
       
  \end{tabular}%
  }
 \caption{$DQI_{C2}$and $DQI_{C6}$ (contradiction) for S1, Case (i)}
 \label{tab:dqic2i1}
\end{table}

\begin{table}
\resizebox{1.0\columnwidth}{!}{%
  \begin{tabular}{lllll}
  \hline
     
    \textbf{Granularity}&\textbf{Count}&\textbf{DQI C2,C6 - T1}&\textbf{DQI C2,C6 - T2}&\textbf{DQI C6 - T5}  \\\hline
      
    \textbf{Sentences} & 2 & 1.0000 & 1.0000&0 \\
    \textbf{Words} & 4 & 6.9282 & 1.0000&0\\
    \textbf{Adjectives} & 0 & nan & nan&0\\
    \textbf{Adverbs} & 0 & nan & nan&0\\
    \textbf{Verbs} & 1 & inf & 1.0000&0\\
    \textbf{Nouns} & 3 & 6.3639 & 1.0000&0\\
    \textbf{Bigrams} & 9 & 20.4101 & 0.2727&0\\
    \textbf{Trigrams} & 8 & 22.6274 & 0.5555&0\\\hline
       
  \end{tabular}%
  }
 \caption{$DQI_{C2}$and $DQI_{C6}$ (contradiction) for S2, Case (i)}
 \label{tab:dqic2i2}
\end{table}

\begin{table}
\resizebox{1.0\columnwidth}{!}{%
  \begin{tabular}{lllll}
  \hline
     
    \textbf{Granularity}&\textbf{Count}&\textbf{DQI C2,C6 - T1}&\textbf{DQI C2,C6 - T2}&\textbf{DQI C6 - T5}  \\\hline
      
    \textbf{Sentences} & 2 & 1.0000 & 1.0000&0 \\
    \textbf{Words} & 11 & 23.5495 & 1.0000&0\\
    \textbf{Adjectives} & 3 & 6.3639 & 1.0000&0\\
    \textbf{Adverbs} & 0 & 6.3639 & nan&0\\
    \textbf{Verbs} & 2 & 4.0000 & 1.0000&0\\
    \textbf{Nouns} & 5 & 12.5000 & 1.0000&0\\
    \textbf{Bigrams} & 19 & 37.4563 & -0.1851&0\\
    \textbf{Trigrams} & 20 & 45.0185 & 0.2000&0\\\hline
       
  \end{tabular}%
  }
 \caption{$DQI_{C2}$and $DQI_{C6}$ (neutral) for S3, Case (i)}
 \label{tab:dqic2i3}
\end{table}

\begin{table}
\resizebox{1.0\columnwidth}{!}{%
  \begin{tabular}{lllll}
  \hline
     
    \textbf{Granularity}&\textbf{Count}&\textbf{DQI C2,C6 - T1}&\textbf{DQI C2,C6 - T2}&\textbf{DQI C6 - T5}  \\\hline
      
    \textbf{Sentences} & 2 & 1.0000 & 1.0000&0 \\
    \textbf{Words} & 12 & 41.5692 & 1.0000&0\\
    \textbf{Adjectives} & 3 & inf & 1.0000&0\\
    \textbf{Adverbs} & 0 & inf & nan&0\\
    \textbf{Verbs} & 4 & inf & 1.0000 &0\\
    \textbf{Nouns} & 5 & 12.5000 & 1.0000&0\\
    \textbf{Bigrams} & 20 & 89.4427 & 0.8095&0\\
    \textbf{Trigrams} & 19 & 4.6757e+16 & 1.0000&0\\\hline
       
  \end{tabular}%
  }
 \caption{$DQI_{C2}$and $DQI_{C6}$ (neutral) for S4, Case (i)}
 \label{tab:dqic2i4}
\end{table}

\begin{table}
\resizebox{1.0\columnwidth}{!}{%
  \begin{tabular}{lllll}
  \hline
     
    \textbf{Granularity}&\textbf{Count}&\textbf{DQI C2,C6 - T1}&\textbf{DQI C2,C6 - T2}&\textbf{DQI C6 - T5}  \\\hline
      
    \textbf{Sentences} & 2 & 1.0000 & 1.0000&0 \\
    \textbf{Words} & 7 & 14.3457 & 1.0000&0\\
    \textbf{Adjectives} & 1 & inf & 1.0000&0\\
    \textbf{Adverbs} & 0 & inf & nan&0\\
    \textbf{Verbs} & 1 & inf & 1.0000 &0\\
    \textbf{Nouns} & 4 & 8.0000 & 1.0000&0\\
    \textbf{Bigrams} & 11 & 36.4828 & 0.6667&0\\
    \textbf{Trigrams} & 10 & 6.8359e+16 & 1.0000&0\\\hline
       
  \end{tabular}%
  }
 \caption{$DQI_{C2}$and $DQI_{C6}$ (entailment) for S5, Case (i)}
 \label{tab:dqic2i5}
\end{table}

\begin{table}
\resizebox{1.0\columnwidth}{!}{%
  \begin{tabular}{lllll}
  \hline
     
    \textbf{Granularity}&\textbf{Count}&\textbf{DQI C2,C6 - T1}&\textbf{DQI C2,C6 - T2}&\textbf{DQI C6 - T5}  \\\hline
      
    \textbf{Sentences} & 2 & 1.0000 & 1.0000&0 \\
    \textbf{Words} & 12 & 30.8285 & 1.0000&0\\
    \textbf{Adjectives} & 3 & inf & 1.0000&0\\
    \textbf{Adverbs} & 0 & inf & nan&0\\
    \textbf{Verbs} & 1 & inf & 1.0000 &0\\
    \textbf{Nouns} & 7 & 20.0041 & 1.0000&0\\
    \textbf{Bigrams} & 25 & 125.0000 & 0.8461&0\\
    \textbf{Trigrams} & 24 & 7.0540e+16 & 1.0000&0\\\hline
       
  \end{tabular}%
  }
 \caption{$DQI_{C2}$and $DQI_{C6}$ (entailment) for S6, Case (i)}
 \label{tab:dqic2i6}
\end{table}

\begin{table}
\resizebox{1.0\columnwidth}{!}{%
  \begin{tabular}{lllll}
  \hline
     
    \textbf{Granularity}&\textbf{Count}&\textbf{DQI C2,C6 - T1}&\textbf{DQI C2,C6 - T2}&\textbf{DQI C6 - T5}  \\\hline
      
    \textbf{Sentences} & 2 & 1.0000 & 1.0000&0 \\
    \textbf{Words} & 6 & 14.6969 & 1.0000&0\\
    \textbf{Adjectives} & 1 & inf & 1.0000&0\\
    \textbf{Adverbs} & 0 & inf & nan&0\\
    \textbf{Verbs} & 1 & inf & 1.0000 &0\\
    \textbf{Nouns} & 4 & 9.2376 & 1.0000&0\\
    \textbf{Bigrams} & 11 & 36.4828 & 0.6667&0\\
    \textbf{Trigrams} & 10 & 6.8359e+16 & 1.0000&0\\\hline
       
  \end{tabular}%
  }
 \caption{$DQI_{C2}$and $DQI_{C6}$ (entailment) for S7, Case (i)}
 \label{tab:dqic2i7}
\end{table}

\begin{table}
\resizebox{1.0\columnwidth}{!}{%
  \begin{tabular}{lllll}
  \hline
     
    \textbf{Granularity}&\textbf{Count}&\textbf{DQI C2,C6 - T1}&\textbf{DQI C2,C6 - T2}&\textbf{DQI C6 - T5}  \\\hline
      
    \textbf{Sentences} & 2 & 1.0000 & 1.0000&0 \\
    \textbf{Words} & 8 & 17.2819 & 1.0000&0\\
    \textbf{Adjectives} & 2 & inf & 1.0000&0\\
    \textbf{Adverbs} & 0 & inf & nan&0\\
    \textbf{Verbs} & 2 & inf & 1.0000 &0\\
    \textbf{Nouns} & 4 & 8.0000 & 1.0000&0\\
    \textbf{Bigrams} & 19&4.6757e+16&1.0000&0\\
    \textbf{Trigrams} & 17 & inf & 1.0000&0\\\hline
       
  \end{tabular}%
  }
 \caption{$DQI_{C2}$and $DQI_{C6}$ (entailment) for S8, Case (i)}
 \label{tab:dqic2i8}
\end{table}

\begin{table}
\resizebox{1.0\columnwidth}{!}{%
  \begin{tabular}{lllll}
  \hline
     
    \textbf{Granularity}&\textbf{Count}&\textbf{DQI C2,C6 - T1}&\textbf{DQI C2,C6 - T2}&\textbf{DQI C6 - T5}  \\\hline
      
    \textbf{Sentences} & 2 & 1.0000 & 1.0000&0 \\
    \textbf{Words} & 7 & 3.3356e+16 & 1.0000&0\\
    \textbf{Adjectives} & 1 & inf & 1.0000&0\\
    \textbf{Adverbs} & 0 & inf & nan&0\\
    \textbf{Verbs} & 2 & inf & 1.0000&0\\
    \textbf{Nouns} & 4 & inf & 1.0000&0\\
    \textbf{Bigrams} & 10 & 6.8359e+16 & 1.0000&0\\
    \textbf{Trigrams} & 8 & inf & 1.0000&0\\\hline
       
  \end{tabular}%
  }
 \caption{$DQI_{C2}$and $DQI_{C6}$ (neutral) for S9, Case (i)}
 \label{tab:dqic2i9}
\end{table}

\begin{table}
\resizebox{1.0\columnwidth}{!}{%
  \begin{tabular}{lllll}
  \hline
     
    \textbf{Granularity}&\textbf{Count}&\textbf{DQI C2,C6 - T1}&\textbf{DQI C2,C6 - T2}&\textbf{DQI C6 - T5}  \\\hline
      
    \textbf{Sentences} & 2 & 1.0000 & 1.0000&0 \\
    \textbf{Words} & 9 & 20.4100 & 1.0000&0\\
    \textbf{Adjectives} & 3 & inf & 1.0000&0\\
    \textbf{Adverbs} & 0 & inf & nan&0\\
    \textbf{Verbs} & 2 & inf & 1.0000 &0\\
    \textbf{Nouns} & 4 & 8.0000 & 1.0000&0\\
    \textbf{Bigrams} & 19 & 4.6757e+16 & 1.0000&0\\
    \textbf{Trigrams} & 17 & 4.6757e+16 & 1.0000&0\\\hline
       
  \end{tabular}%
  }
 \caption{$DQI_{C2}$and $DQI_{C6}$ (neutral) for S10, Case (i)}
 \label{tab:dqic2i10}
\end{table}

\begin{table}
\resizebox{1.0\columnwidth}{!}{%
  \begin{tabular}{lllll}
  \hline
     
    \textbf{Granularity}&\textbf{Count}&\textbf{DQI C2,C6 - T1}&\textbf{DQI C2,C6 - T2}&\textbf{DQI C6 - T5}  \\\hline
      
    \textbf{Sentences} & 2 & 1.0000 & 1.0000&0 \\
    \textbf{Words} & 10 & 23.7170 & 1.0000&0\\
    \textbf{Adjectives} & 1 & inf & 1.0000&0\\
    \textbf{Adverbs} & 0 & inf & nan&0\\
    \textbf{Verbs} & 1 & inf & 1.0000&0\\
    \textbf{Nouns} & 8 & 18.4752 & 1.0000&0\\
    \textbf{Bigrams} & 20 & 1.4046e+17 & 1.0000&0\\
    \textbf{Trigrams} & 18 & 7.0027e+16 & 1.0000&0\\\hline
       
  \end{tabular}%
  }
 \caption{$DQI_{C2}$and $DQI_{C6}$ (contradiction) for S11, Case (i)}
 \label{tab:dqic2i11}
\end{table}

\begin{table}
\resizebox{1.0\columnwidth}{!}{%
  \begin{tabular}{lllll}
  \hline
     
    \textbf{Granularity}&\textbf{Count}&\textbf{DQI C2,C6 - T1}&\textbf{DQI C2,C6 - T2}&\textbf{DQI C6 - T5}  \\\hline
      
    \textbf{Sentences} & 2 & 1.0000 & 1.0000&0 \\
    \textbf{Words} & 11 & 16.3156 & 1.0000&0\\
    \textbf{Adjectives} & 1 & inf & 1.0000&0\\
    \textbf{Adverbs} & 0 & inf & nan&0\\
    \textbf{Verbs} & 1 & inf & 1.0000&0\\
    \textbf{Nouns} & 8 & 11.3137 & 1.0000&0\\
    \textbf{Bigrams} & 18 & 55.6619 & 0.6000&0\\
    \textbf{Trigrams} & 18 & 7.0027e+16 & 1.0000&0\\\hline
       
  \end{tabular}%
  }
 \caption{$DQI_{C2}$and $DQI_{C6}$ (contradiction) for S12, Case (i)}
 \label{tab:dqic2i12}
\end{table}

\begin{table}
\resizebox{1.0\columnwidth}{!}{%
  \begin{tabular}{llll}
  \hline
     
\textbf{Sample}&\textbf{DQI C5 -T2,C6 - T3}&\textbf{DQI C5 - T3,C6 - T4}&\textbf{DQI C5 - T4}\\\hline
      
    \textbf{S1}&0.2500 & nan&0.8938 \\
    \textbf{S2}&0.5000 & nan&0.9060 \\
    \textbf{S3}&0.2500 & nan&0.8722 \\
    \textbf{S4}&0.0830 & nan&0.6512 \\
    \textbf{S5}&0.1111 & nan&0.6982 \\
    \textbf{S6}&0.0588 & nan&0.6806 \\
    \textbf{S7}&1.0000 & nan&0.7443 \\
    \textbf{S8}&0.1250 & nan&0.7672 \\
    \textbf{S9}&0.3333 & nan&0.8219 \\
    \textbf{S10}&0.1250 & nan&0.7750 \\
    \textbf{S11}&0.2000 & nan&0.7616 \\
    \textbf{S12}&0.2000 & nan&0.8255 \\\hline
       
  \end{tabular}%
  }
 \caption{T2/3 and T3/4 for $DQI_{C5}$/$DQI_{C6}$, T4 for $DQI_{C5}$ , Case (i)}
 \label{tab:dqic5i234}
\end{table}

\begin{table}
\resizebox{1.0\columnwidth}{!}{%
  \begin{tabular}{llll}
  \hline
     
    \multirow{3}{*}{\textbf{Sample Set}} &
      \multicolumn{3}{c}{\textbf{Terms}}\\
      &&\textbf{T1}& \\
      &ISIM=0.5&ISIM=0.6&ISIM=0.7\\\hline
      
    \textbf{+S1} & 2.53901172& 3.40305015& 5.15852057\\
    \textbf{+S2} & 2.46282325& 3.26756734& 4.85347200\\
    \textbf{+S3} & 2.68605483& 3.67251159& 5.80405898\\
    \textbf{+S4} & 6.61292347& 19.5239860& 20.4998054\\
    \textbf{+S5} & 5.04523160& 10.1825780& 557.710874\\
    \textbf{+S6} & 5.53586344& 12.4007484& 51.6536766\\
    \textbf{+S7} & 4.09274400& 6.92833358& 22.5556185\\
    \textbf{+S8} & 3.74140198& 5.97801932& 14.8633715\\
    \textbf{+S9} & 3.10654715& 4.50651832& 8.20339191\\
    \textbf{+S10} & 3.6359872& 5.71335622& 13.3282739\\
    \textbf{+S11} & 3.8217013& 6.18568557& 16.2170311\\
    \textbf{+S12} & 3.0714259& 4.43298421& 7.96294530\\
    \hline
       
  \end{tabular}%
  }
 \caption{T1 for $DQI_{C5}$, Case (i)}
 \label{tab:dqic5i1}
\end{table}

\begin{table}
\resizebox{1.0\columnwidth}{!}{%
  \begin{tabular}{llll}
  \hline
     
\textbf{Sample}&\textbf{DQI C3 - T1}&\textbf{DQI C3 - T2}&\textbf{DQI C4}\\\hline
      
    \textbf{S1}&0.8938 & 2.0&0.9896 \\
    \textbf{S2}&0.9060 & 2.0&0.7779 \\
    \textbf{S3}&0.8722 & 2.0&1.3180 \\
    \textbf{S4}&0.6512 & 2.0&0.9093 \\
    \textbf{S5}&0.6982 & 2.0&0.0848 \\
    \textbf{S6}&0.6806 & 2.0&1.1088 \\
    \textbf{S7}&0.7443 & 2.0&0.6826 \\
    \textbf{S8}&0.7672 & 2.0&1.0860 \\
    \textbf{S9}&0.8219 & 2.0&0.5084 \\
    \textbf{S10}&0.7750 & 2.0&0.9601 \\
    \textbf{S11}&0.7616 & 2.0&1.1597 \\
    \textbf{S12}&0.8255 & 2.0&1.2076 \\\hline
       
  \end{tabular}%
  }
 \caption{T1 and T2 for $DQI_{C3}$, $DQI_{C4}$, Case (i)}
 \label{tab:dqic3i}
\end{table}

\begin{table}
\resizebox{1.0\columnwidth}{!}{%
  \begin{tabular}{llllllll}
  \hline
     
\textbf{Sample}&\textbf{DQI C1}&\textbf{DQI C2}&\textbf{DQI C3}&\textbf{DQI C4}&\textbf{DQI C5 (ISIM=0.5)}&\textbf{DQI C6}&\textbf{DQI C7}\\\hline
      
    \textbf{S1}&2.1906&80.2076&2.8938&0.9896&12.3961&80.4576&0\\
    \textbf{S2}&0.7467&32.4274&2.9060&0.7779&9.7696&32.9274& 0\\
    \textbf{S3}&2.2302&49.4839&2.8722&1.3180&15.0742&49.7339&0 \\
    \textbf{S4}&7.8969&4.6757E+16&2.6512&0.9093&18.2884&4.6757E+16&0 \\
    \textbf{S5}&0.0693&6.8359E+16&2.6982&0.0848&16.3837&6.8359E+16& 0\\
    \textbf{S6}&11.4325&7.0540E+16&2.6806&1.1088&23.0456&7.054E+16& 0\\
    \textbf{S7}&0.0594&6.8359E+16&2.7443&0.6826&16.4604&6.8359E+16& 0\\
    \textbf{S8}&5.0289&4.6757E+16&2.7672&1.0860&15.8438&4.6757E+16& 0\\
    \textbf{S9}&1.4835&1.0171E+17&2.8219&0.5084&77.4403&1.01715E+17& 0\\
    \textbf{S10}&5.0388&9.3514E+16&2.7750&0.9601&16.2461&9.3514E+16& 0\\
    \textbf{S11}&2.9274&2.1048E+17&2.7616&1.1597&20.1601&2.10487E+17& 0\\
    \textbf{S12}&2.9373&7.0027E+16&2.8255&1.2076&16.6541&7.0027E+16& 0\\
\hline
       
  \end{tabular}%
  }
 \caption{DQI Terms, Case (i)}
\end{table}

\begin{table}
\resizebox{1.0\columnwidth}{!}{%
  \begin{tabular}{lllll}
  \hline
     
    \multirow{2}{*}{\textbf{Sample Set}} &
      \multicolumn{3}{c}{\textbf{Terms}}&
      \multicolumn{1}{c}{\textbf{DQI C1}}\\
      &\textbf{T1}&\textbf{T2}&\textbf{T3}&  \\\hline
      
    \textbf{Original} & 5.8200 & 6.6656 & 0.9300 & 12.0190  \\
    \textbf{+S1} & 5.7921 & 6.6347 & 0.9307 & 11.9669  \\
    \textbf{+S2} & 5.7822 & 6.6507 & 0.9307 & 11.9719  \\
    \textbf{+S3} & 5.8020 & 6.6409 & 0.9307 & 11.9826  \\
    \textbf{+S4} & 5.8119 & 6.6550 & 0.9307 & 12.0056  \\
    \textbf{+S5} & 5.7723 & 6.6590 & 0.9208 & 11.9038  \\
    \textbf{+S6} & 5.7822 & 6.6849 & 0.9307 & 12.0038  \\
    \textbf{+S7} & 5.7822 & 6.6470 & 0.9307 & 11.9685  \\
    \textbf{+S8} & 5.7921 & 6.6422 & 0.9307 & 11.9739  \\
    \textbf{+S9} & 5.8020 & 6.6551 & 0.9307 & 11.9958  \\
    \textbf{+S10} & 5.7921 & 6.6422 & 0.9307 & 11.9739  \\
    \textbf{+S11} & 5.7921 & 6.6355 & 0.9307 & 11.9677  \\
    \textbf{+S12} & 5.8317 & 6.6355 & 0.930 & 12.0073  \\\hline
       
  \end{tabular}%
  }
 \caption{$DQI_{C1}$ for Case (ii)}
 \label{tab:44}
\end{table}

\begin{table*}
\resizebox{2.0\columnwidth}{!}{%
  \begin{tabular}{llllllllllllllllll}
  \hline
     
    \multirow{2}{*}{\textbf{Sample Set}} &
      \multicolumn{2}{c}{\textbf{Sentences}}&
      \multicolumn{2}{c}{\textbf{Words}}&
      \multicolumn{2}{c}{\textbf{Adjectives}}&
      \multicolumn{2}{c}{\textbf{Adverbs}}&
      \multicolumn{2}{c}{\textbf{Verbs}}&
      \multicolumn{2}{c}{\textbf{Nouns}}&
      \multicolumn{2}{c}{\textbf{Bigrams}}&
      \multicolumn{2}{c}{\textbf{Trigrams}}&
      \multicolumn{1}{c}{\textbf{DQI C2}}\\
      &\textbf{T1}&\textbf{T2}&\textbf{T1}&\textbf{T2}&\textbf{T1}&\textbf{T2}&\textbf{T1}&\textbf{T2}&\textbf{T1}&\textbf{T2}&\textbf{T1}&\textbf{T2}&\textbf{T1}&\textbf{T2}&\textbf{T1}&\textbf{T2}& \\\hline
      
    \textbf{Original} & 2807.2405 & 0.9800 & 137.2755 & 0.6371 & 52.0534 & 0.3111 & 20.0385 & -0.04 & 46.8398 & -0.025 & 54.2786 & 0.3888 & 707.8112 & 0.8852 & 2723.6406 & 0.8910 &   5927.1970\\
    \textbf{+S1} & 2849.6668 & 0.9802 & 137.0171 & 0.6368 & 55.6705 & 0.3065 & 21.7786 & -0.1111 & 50.8642 & -0.0356 & 49.5464 & 0.3452 & 697.9764 & 0.8815 & 2706.4317 & 0.8857 &  5922.7847\\
    \textbf{+S2} & 2849.6668 & 0.9802 & 137.0171 & 0.6368 & 55.6705 & 0.3065 & 21.7789 & -0.1111 & 50.8642 & -0.0356 & 49.5464 & 0.3452 & 697.9764 & 0.8815 & 2706.4317 & 0.8857 &  5922.7847\\
    \textbf{+S3} & 2849.6668 & 0.9802 & 137.9140 & 0.6393 & 52.6620 & 0.2414 & 17.4592 & 0.0833 & 43.8252 & -0.0661 & 55.2815 & 0.3505 & 712.9377 & 0.8847 & 2763.8091 & 0.8924 &  6009.2173\\
    \textbf{+S4} &2849.6668 &0.9802 & 138.3361 & 0.6392 & 54.2001 & 0.2576 & 24.9929 & 0.1250 & 48.5320 & -0.0313 & 50.1523 & 0.3498 & 706.9163 & 0.9043 & 2765.4396 & 0.8921 &  6021.0912\\
    \textbf{+S5} & 2849.6668& 0.9802& 135.4295& 0.6365& 49.2904& 0.2619& 23.3950& 0.0000& 49.0989&-0.0840& 52.0959& 0.3432& 697.8102& 0.9029& 2649.2411& 0.8895&  5892.6612\\
    \textbf{+S6} & 2849.6668& 0.9802& 137.1086& 0.6379& 53.9239& 0.3609& 20.0385& -0.0400& 48.0375&-0.0538& 52.8044& 0.3463& 711.5407& 0.9064& 2723.0651& 0.8903& 5984.3517\\
    \textbf{+S7} &2849.6668 &0.9802 &137.4205 &0.6359 &48.4367 &0.2015 &35.9211 &0.1538 &45.0502 &-0.0361 &54.6786 &0.4303 &710.2298 &0.9058 &2739.3807 &0.8916 & 6003.5736\\
    \textbf{+S8} & 2849.6668& 0.9802& 136.2514& 0.6368& 49.6075&0.2268& 57.0399& 0.3846& 49.9798&-0.0445& 52.5582& 0.3432& 705.7911&0.9052& 2693.8612& 0.8888& 5962.1966\\
    \textbf{+S9} & 2849.6668& 0.9802& 137.6593& 0.6375& 58.2917& 0.3388& 24.5189& -0.0244& 52.4063& 0.0041& 50.5623& 0.3237& 707.6845& 0.9048& 2742.9126& 0.8915& 6002.3536\\
    \textbf{+S10} & 2849.6668& 0.9802& 136.2477& 0.6371& 56.5772& 0.2511& 29.8974& -0.1034& 51.6379& -0.0206& 51.8621& 0.3484& 708.3581& 0.9052& 2718.4279& 0.8899& 5968.5017\\
    \textbf{+S11}& 2849.6668& 0.9802& 137.7623& 0.6373& 49.6725& 0.2197& 20.5196& -0.0667& 47.5031& -0.0370& 54.6531& 0.3741& 717.2547& 0.9062& 2767.0664& 0.8921& 6027.7480\\
    \textbf{+S12} & 2849.6668& 0.9802& 139.5281& 0.6413& 59.9832& 0.3101& 15.2008& -0.2727& 52.8410& 0.0723& 50.6446& 0.3174& 713.8007& 0.9052& 2763.0228& 0.8920& 6027.8220\\\hline
       
  \end{tabular}%
  }
 \caption{$DQI_{C2}$ for Case (ii)}
 \label{tab:45}
\end{table*}

\begin{table*}
\resizebox{2.0\columnwidth}{!}{%
  \begin{tabular}{llllllllll}
  \hline
     
    \multirow{3}{*}{\textbf{Sample Set}} &
      \multicolumn{6}{c}{\textbf{Terms}}&
      \multicolumn{3}{c}{\textbf{DQI C3 (e=0.5)}}\\
      &&\textbf{T1}&&&\textbf{T2 (SIM=0.5)}&&&&  \\
      &SIM=0.5&SIM=0.6&SIM=0.7&e=0.25&e=0.33&e=0.5&SIM=0.5&SIM=0.6&SIM=0.7\\\hline
      
    \textbf{Original} & 14.1194 & 4.9647 & 4.2968 & 200.0000 & 200.0000 & 198.4692 & 212.5886&	203.4339&	202.766  \\
    \textbf{+S1} & 14.0959 & 4.9880 & 4.2882 & 202.0000 & 202.0000 & 199.9066 &214.0025	&204.8946	&204.1948 \\
    \textbf{+S2} & 14.2729 & 4.8939 & 4.3000 & 202.0000 & 202.0000 & 200.9450 &215.2179	&205.8389	&205.245 \\
    \textbf{+S3} & 14.1055 & 4.9749 & 4.2710 & 202.0000 & 202.0000 & 199.9066 &214.0121	&204.8815	&204.1776 \\
    \textbf{+S4} & 14.1285 & 4.9797 & 4.3134 & 202.0000 & 202.0000 & 200.4539 &214.5824	&205.4336	&204.7673 \\
    \textbf{+S5} & 14.1522 & 4.9797 & 4.3072 & 202.0000 & 202.0000 & 200.4539 &214.6061	&205.4336	&204.7611 \\
    \textbf{+S6} & 14.1961 & 4.9827 & 4.3041 & 202.0000 & 202.0000 & 200.4539 &214.65	    &205.4366	&204.758 \\
    \textbf{+S7} & 14.1656 & 4.9842 & 4.3197 & 202.0000 & 202.0000 & 200.4539 &214.6195	&205.4381	&204.7736  \\
    \textbf{+S8} & 14.2711 & 4.9873 & 4.3015 & 202.0000 & 202.0000 & 200.9450 &215.2161	&205.9323	&205.2465 \\
    \textbf{+S9} & 14.2321& 4.9836 & 4.3214 & 202.0000 & 202.0000 & 200.9450 &215.1771	&205.9286	&205.2664 \\
    \textbf{+S10} & 14.2859 & 4.9888 & 4.2944 & 202.0000 & 202.0000 & 200.9450 &215.2309	&205.9338	&205.2394 \\
    \textbf{+S11} & 14.1403 & 4.9720 & 4.3122 & 202.0000 & 202.0000 & 200.4539 &214.5942	&205.4259	&204.7661 \\
    \textbf{+S12} & 14.1707 & 4.9874 & 4.3211 & 202.0000 & 202.0000 & 199.9066 &214.0773	&204.894	&204.2277  \\\hline
       
  \end{tabular}%
  }
 \caption{$DQI_{C3}$ for Case (ii)}
\end{table*}

\begin{table} 
\centering
\scriptsize
\resizebox{0.5\columnwidth}{!}{%
\begin{tabular}{lll}
\hline
\textbf{Sample Set} & \textbf{DQI C4}  \\
\hline
\textbf{Original} & 0.00657581 \\
\textbf{+S1} & 0.00653241 \\
\textbf{+S2} & 0.00652070 \\
\textbf{+S3} & 0.00654317 \\
\textbf{+S4} & 0.00652860 \\
\textbf{+S5} & 0.00610259 \\
\textbf{+S6} & 0.00653705 \\
\textbf{+S7} & 0.00651307 \\
\textbf{+S8} & 0.00653624 \\
\textbf{+S9} & 0.00649185 \\
\textbf{+S10} & 0.00653108 \\
\textbf{+S11} & 0.00653874 \\
\textbf{+S12} & 0.00654020 \\\hline
              &            &                 
\end{tabular}%
}
\caption{$DQI_{C4}$ for Case (ii)}
\label{tab:my-table}
\end{table}

\begin{table*}
\resizebox{2.0\columnwidth}{!}{%
  \begin{tabular}{llllllllll}
  \hline
     
    \multirow{3}{*}{\textbf{Sample Set}} &
      \multicolumn{8}{c}{\textbf{Terms}}&
      \multicolumn{1}{c}{\textbf{DQI C5 (ISIM=0.5)}}\\
      &&\textbf{T1}&&\textbf{T2}&\textbf{T3}&\textbf{T4}&\textbf{T5}&\textbf{T6}&  \\
      &ISIM=0.5&ISIM=0.6&ISIM=0.7&&&&&&\\\hline
      
    \textbf{Original} & 3.79338794& 5.79942751& 9.64213607&0.13869626&0.06846071&0.00106449&19.2658&0.08669236&4.00160940\\
    \textbf{+S1} & 3.77492292& 5.75927311& 9.55986754&0.13950276&0.06756993&0.00105670&19.1081&0.08686184&3.98305231\\
    \textbf{+S2} & 3.77320467& 5.75527455& 9.54885537&0.13988920&0.06771915&0.00105824&19.1048&0.08711365&3.98187126\\
    \textbf{+S3} & 3.77796738& 5.76636257& 9.57941700&0.13950276&0.06756993&0.00105429&19.0986&0.08666733&3.98609436\\
    \textbf{+S4} & 3.80946946& 5.84007436& 9.69296631&0.13797814&0.06754694&0.00105432&19.2038&0.08661618&4.01604886\\
    \textbf{+S5} & 3.80273001& 5.82425011& 9.73687404&0.13854595&0.06744772&0.00105055&19.1196&0.08696758&4.00977423\\
    \textbf{+S6} & 3.80524680& 5.83015604& 9.72041244&0.13704206&0.06799806&0.00105172&19.1444&0.08642433&4.01133864\\
    \textbf{+S7} & 3.79613706& 5.80879868& 9.69710399&0.14008322&0.06781511&0.00104881&19.1444&0.08708462&4.00508420\\
    \textbf{+S8} & 3.79286615& 5.80114342& 9.67578885&0.13873626&0.06744340&0.00104868&19.1246&0.08673365&4.00009449\\
    \textbf{+S9} & 3.78510214& 5.78300049& 9.62542175&0.13969571&0.06763740&0.00105033&19.7681&0.08710369&3.99348558\\
    \textbf{+S10}& 3.79176275& 5.79856261& 9.66861134&0.13873626&0.06744340&0.00104875&19.1295&0.08675259&3.99899116\\
    \textbf{+S11}& 3.79366621& 5.80301526& 9.68099727&0.13931034&0.06751676&0.00104867&19.1840&0.08695819&4.00154198\\
    \textbf{+S12}& 3.78458008& 5.78178193& 9.62204642&0.13931034&0.06751676&0.00105054&19.1213&0.08674638&3.99245772\\\hline
       
  \end{tabular}%
  }
 \caption{$DQI_{C5}$ for Case (ii)}
\end{table*}

\begin{table}
\resizebox{1.0\columnwidth}{!}{%
  \begin{tabular}{llllllll}
  \hline
     
    \multirow{3}{*}{\textbf{Sample Set}} &
      \multicolumn{7}{c}{\textbf{Terms}}\\
      &\textbf{entailment}&&\textbf{neutral}&&\textbf{contradiction}&&\\
      &\textbf{T1}&\textbf{T2}&\textbf{T1}&\textbf{T2}&\textbf{T1}&\textbf{T2}&\textbf{T5}\\\hline
      
    \textbf{Original} &7.1303e+16&1.0000&1045.3358&2.0833&7.1303e+16&1.0000&92.8203\\
    \textbf{+S1} &7.1303e+16&1.0000&1045.3358&2.0833&1.4267e+17&1.0417&93.7485\\
    \textbf{+S2} &7.1303e+16&1.0000&1045.3358&2.0833&1.4267e+17&1.0417&93.7485\\
    \textbf{+S3} &7.1303e+16&1.0000&1075.9298&2.1250&7.1303e+16&1.0000&93.7485\\
    \textbf{+S4} &7.1303e+16&1.0000&1075.9298&2.1250&7.1303e+16&1.0000&93.7485\\
    \textbf{+S5} &1.4267e+17&1.0000&1045.3358&2.0000&7.1303e+16&0.9600&93.7485\\
    \textbf{+S6} &1.4267e+17&1.0000&1045.3358&2.0000&7.1303e+16&0.9600&93.7485\\
    \textbf{+S7} &1.4267e+17&1.0000&1045.3358&2.0000&7.1303e+16&0.9600&93.7485\\
    \textbf{+S8} &1.4267e+17&1.0000&1045.3358&2.0000&7.1303e+16&0.9600&93.7485\\
    \textbf{+S9} &7.1303e+16&1.0000&1075.9298&2.1250&7.1303e+16&1.0000&93.7485\\
    \textbf{+S10}&7.1303e+16&1.0000&1075.9298&2.1250&7.1303e+16&1.0000&93.7485\\
    \textbf{+S11}&7.1303e+16&1.0000&1045.3358&2.0833&1.4267e+17&1.0417&93.7485\\
    \textbf{+S12}&7.1303e+16&1.0000&1045.3358&2.0833&1.4267e+17&1.0417&93.7485\\\hline
       
  \end{tabular}%
  }
 \caption{Case (ii), Sentence Granularity Terms in $DQI_{C6}$}
\end{table}
\begin{table}
\resizebox{1.0\columnwidth}{!}{%
  \begin{tabular}{llllllll}
  \hline
     
    \multirow{3}{*}{\textbf{Sample Set}} &
      \multicolumn{7}{c}{\textbf{Terms}}\\
      &\textbf{entailment}&&\textbf{neutral}&&\textbf{contradiction}&&\\
      &\textbf{T1}&\textbf{T2}&\textbf{T1}&\textbf{T2}&\textbf{T1}&\textbf{T2}&\textbf{T5}\\\hline
      
    \textbf{Original} &113.4748&0.5548&136.5557&0.6599&105.1059&0.5255&2.4416\\
    \textbf{+S1} &113.4748&0.5548&136.5557&0.6599&103.7067&0.5219&2.4509\\
    \textbf{+S2} &113.4748&0.5548&136.5557&0.6599&107.3208&0.5339&2.4325\\
    \textbf{+S3} &113.4748&0.5548&137.7114&0.6182&105.1059&0.5255&2.3670\\
    \textbf{+S4} &113.4748&0.5548&138.5993&0.6422&105.1059&0.5255&2.4336\\
    \textbf{+S5} &109.7512&0.5298&136.5557&0.6599&105.1059&0.5255&2.4566\\
    \textbf{+S6} &117.4812&0.5679&136.5557&0.6599&105.1059&0.5255&2.4518\\
    \textbf{+S7} &115.2611&0.5520&136.5557&0.6599&105.1059&0.5255&2.4241\\
    \textbf{+S8} &110.1518&0.5562&136.5557&0.6599&105.1059&0.5255&2.4491\\
    \textbf{+S9} &113.4748&0.5548&136.5917&0.6604&105.1059&0.5255&2.4467\\
    \textbf{+S10}&113.4748&0.5548&134.4891&0.6595&105.1059&0.5255&2.4267\\
    \textbf{+S11}&113.4748&0.5548&136.5557&0.6599&110.1129&0.5304&2.4310\\
    \textbf{+S12}&113.4748&0.5548&136.5557&0.6599&112.6038&0.5459&2.4524\\\hline
       
  \end{tabular}%
  }
 \caption{Case (ii), Word Granularity Terms in $DQI_{C6}$}
\end{table}
\begin{table}
\resizebox{1.0\columnwidth}{!}{%
  \begin{tabular}{llllllll}
  \hline
     
    \multirow{3}{*}{\textbf{Sample Set}} &
      \multicolumn{7}{c}{\textbf{Terms}}\\
      &\textbf{entailment}&&\textbf{neutral}&&\textbf{contradiction}&&\\
      &\textbf{T1}&\textbf{T2}&\textbf{T1}&\textbf{T2}&\textbf{T1}&\textbf{T2}&\textbf{T5}\\\hline
      
    \textbf{Original} &65.4824&0.1935&48.9086&0.1130&44.8057&-0.2113&2.6514\\
    \textbf{+S1} &74.6675&0.0909&50.8008&0.1500&57.0071&0.0164&2.8685\\
    \textbf{+S2} &61.3138&-0.0588&52.7111&0.0815&51.3651&-0.1351&3.1961\\
    \textbf{+S3} &76.2138&0.0588&46.8815&0.1339&60.6168&0.0476&3.0158\\
    \textbf{+S4} &62.4955&-0.0423&58.8794&0.2480&52.4764&-0.1389&3.2262\\
    \textbf{+S5} &71.8135&-0.0133&48.3257&0.1707&57.2251&0.0667&2.9149\\
    \textbf{+S6} &71.5360&0.0571&50.7164&0.1897&49.4934&0.0000&2.5007\\
    \textbf{+S7} &69.5736&0.1475&52.5575&0.0676&58.1186&0.0312&2.6028\\
    \textbf{+S8} &73.1520&0.1250&45.2213&0.1000&51.0064&0.0149&2.7511\\
    \textbf{+S9} &68.4000&0.0000&48.3109&0.0615&52.7210&0.0000&2.8224\\
    \textbf{+S10}&72.3354&0.0684&48.7879&0.1147&53.0237&0.0667&3.0774\\
    \textbf{+S11}&68.2115&-0.0410&47.9655&0.1355&50.9620&-0.0294&2.6320\\
    \textbf{+S12}&74.7011&0.0000&51.4393&0.0518&45.1122&-0.1384&2.6840\\\hline
       
  \end{tabular}%
  }
 \caption{Case (ii), Adjective Granularity Terms in $DQI_{C6}$}
\label{tab:51-54}
\end{table}
\begin{table}
\resizebox{1.0\columnwidth}{!}{%
  \begin{tabular}{llllllll}
  \hline
     
    \multirow{3}{*}{\textbf{Sample Set}} &
      \multicolumn{7}{c}{\textbf{Terms}}\\
      &\textbf{entailment}&&\textbf{neutral}&&\textbf{contradiction}&&\\
      &\textbf{T1}&\textbf{T2}&\textbf{T1}&\textbf{T2}&\textbf{T1}&\textbf{T2}&\textbf{T5}\\\hline
      
    \textbf{Original} &18.4752&0.2000&21.4630&0.1765&6.3640&0.0000&5.1159\\
    \textbf{+S1} &3.6029e+16&1.0000&16.4141&-0.0769&6.3640&0.0000&3.0036\\
    \textbf{+S2} &10.0021&0.3333&13.4297&0.2632&9.2376&0.0000&2.9621\\
    \textbf{+S3} &16.0997&0.4287&25.0000&0.3333&6.3640&0.0000&4.8231\\
    \textbf{+S4} &inf&1.0000&20.8025&0.0000&9.2376&0.2000&3.4788\\
    \textbf{+S5} &20.0042&0.5000&19.2428&0.1250&12.5&0.3333&4.2973\\
    \textbf{+S6} &inf&1.0000&21.4630&0.1765&6.3639&0.0000&2.9468\\
    \textbf{+S7} &28.6378&0.6000&19.0918&0.0000&6.3639&0.0000&3.5977\\
    \textbf{+S8} &18.4752&0.2000&27.6955&0.4444&9.2376&0.2000&3.4223\\
    \textbf{+S9} &21.6481&0.2727&28.6216&0.3000&6.3639&0.0000&5.3589\\
    \textbf{+S10}&8.0632&-0.2307&19.2428&0.1250&9.6096&0.0000&4.3729\\
    \textbf{+S11}&inf&1.0000&19.2428&0.1250&9.2376&0.2000&4.0262\\
    \textbf{+S12}&inf&1.0000&23.7684&0.2222&6.3639&0.0000&4.1769\\\hline
       
  \end{tabular}%
  }
 \caption{Case (ii), Adverb Granularity Terms in $DQI_{C6}$}
\end{table}\begin{table}
\resizebox{1.0\columnwidth}{!}{%
  \begin{tabular}{llllllll}
  \hline
     
    \multirow{3}{*}{\textbf{Sample Set}} &
      \multicolumn{7}{c}{\textbf{Terms}}\\
      &\textbf{entailment}&&\textbf{neutral}&&\textbf{contradiction}&&\\
      &\textbf{T1}&\textbf{T2}&\textbf{T1}&\textbf{T2}&\textbf{T1}&\textbf{T2}&\textbf{T5}\\\hline
      
    \textbf{Original} &65.4824&0.1935&51.9736&-0.0598&35.1110&-0.1081&2.7836\\
    \textbf{+S1} &40.3696&-0.2069&48.5430&-0.1525&29.9195&-0.2405&2.4728\\
    \textbf{+S2} &43.9037&-0.2424&53.3506&-0.0093&30.1625&-0.0909&2.6133\\
    \textbf{+S3} &37.4444&-0.3030&56.2047&-0.1057&27.3594&-0.2286&2.3308\\
    \textbf{+S4} &42.1040&-0.3333&46.2161&-0.0973&31.2449&-0.1667&2.5586\\
    \textbf{+S5} &38.3571&-0.3714&50.6384&-0.0182&24.4386&-0.2000&2.5610\\
    \textbf{+S6} &41.7648&-0.2537&48.9552&-0.0280&28.8722&-0.1642&2.7063\\
    \textbf{+S7} &46.5989&-0.2537&53.4887&-0.1260&31.1722&-0.2500&2.2977\\
    \textbf{+S8} &35.4040&-0.3548&48.3655&-0.0990&26.0207&-0.2615&2.7680\\
    \textbf{+S9} &40.6156&-0.2000&53.4014&-0.1056&32.0340&-0.2307&2.5957\\
    \textbf{+S10}&41.3657&-0.3230&53.0775&-0.0847&29.1653&-0.2876&2.2606\\
    \textbf{+S11}&42.3999&-0.2187&46.3814&-0.1452&33.3842&-0.1267&2.6794\\
    \textbf{+S12}&37.5858&-0.2258&49.7109&-0.1071&26.0396&-0.0667&2.6669\\\hline

  \end{tabular}%
  }
 \caption{Case (ii), Verb Granularity Terms in $DQI_{C6}$}
\end{table}
\begin{table}
\resizebox{1.0\columnwidth}{!}{%
  \begin{tabular}{llllllll}
  \hline
     
    \multirow{3}{*}{\textbf{Sample Set}} &
      \multicolumn{7}{c}{\textbf{Terms}}\\
      &\textbf{entailment}&&\textbf{neutral}&&\textbf{contradiction}&&\\
      &\textbf{T1}&\textbf{T2}&\textbf{T1}&\textbf{T2}&\textbf{T1}&\textbf{T2}&\textbf{T5}\\\hline
      
    \textbf{Original} &42.7808&-0.3056&53.6301&0.2841&38.7466&-0.2050&2.3372\\
    \textbf{+S1} &38.3026&-0.3659&52.7785&0.2989&39.4878&-0.2601&2.4916\\
    \textbf{+S2} &35.9868&-0.2752&51.9745&0.3097&41.0652&-0.2558&2.3264\\
    \textbf{+S3} &36.7162&-0.3247&52.4598&0.2667&41.5999&-0.2485&2.3551\\
    \textbf{+S4} &36.7565&-0.2617&53.2731&0.2570&37.4839&-0.2075&2.3918\\
    \textbf{+S5} &33.0670&-0.2752&54.0598&0.3030&44.1367&-0.2817&2.3645\\
    \textbf{+S6} &38.3611&-0.3250&54.9709&0.3040&42.2864&-0.2528&2.5035\\
    \textbf{+S7} &37.7188&-0.3414&51.8644&0.2844&37.6200&-0.2327&2.6013\\
    \textbf{+S8} &38.9773&-0.3254&55.4119&0.3028&41.6562&-0.2441&2.4018\\
    \textbf{+S9} &35.4958&-0.3200&50.3967&0.3313&39.9118&-0.2121&2.4067\\
    \textbf{+S10}&32.9868&-0.2765&52.1225&0.2954&38.6028&-0.2484&2.4450\\
    \textbf{+S11} &36.0093&-0.3333&55.2239&0.3352&42.8904&-0.2402&2.4570\\
    \textbf{+S12} &34.8526&-0.3509&50.4304&0.3113&51.0263&-0.2448&2.5026\\\hline

  \end{tabular}%
  }
 \caption{Case (ii), Noun Granularity Terms in $DQI_{C6}$}
 \label{tab:54}
\end{table}
\begin{table}
\resizebox{1.0\columnwidth}{!}{%
  \begin{tabular}{llllllll}
  \hline
     
    \multirow{3}{*}{\textbf{Sample Set}} &
      \multicolumn{7}{c}{\textbf{Terms}}\\
      &\textbf{entailment}&&\textbf{neutral}&&\textbf{contradiction}&&\\
      &\textbf{T1}&\textbf{T2}&\textbf{T1}&\textbf{T2}&\textbf{T1}&\textbf{T2}&\textbf{T5}\\\hline
      
    \textbf{Original} &497.2044&0.8411&620.1037&0.9075&415.2737&0.8610&0.7924\\
    \textbf{+S1} &497.2043&0.8411&620.1037&0.9075&403.4774&0.8206&0.7928\\
    \textbf{+S2} &497.2043&0.8411&620.1037&0.9075&427.4754&0.8636&0.7917\\
    \textbf{+S3} &497.2043&0.8411&625.7171&0.8873&415.2737&0.8610&0.7694\\
    \textbf{+S4} &497.2043&0.8411&616.7056&0.9055&415.2737&0.8610&0.7864\\
    \textbf{+S5} &473.5139&0.8528&620.1037&0.9075&415.2737&0.8610&0.8045\\
    \textbf{+S6} &518.7792&0.8684&620.1037&0.9075&415.2737&0.8610&0.8088\\
    \textbf{+S7} &503.1652&0.8648&620.1037&0.9075&415.2737&0.8610&0.7960\\
    \textbf{+S8} &491.4631&0.8588&620.1037&0.9075&415.2737&0.8610&0.8069\\
    \textbf{+S9} &497.2043&0.8411&617.3021&0.9064&415.2737&0.8610&0.7986\\
    \textbf{+S10}&497.2043&0.8411&619.8558&0.9072&415.2737&0.8610&0.7936\\
    \textbf{+S11} &497.2043&0.8411&620.1037&0.9075&437.4726&0.8657&0.8003\\
    \textbf{+S12} &497.2043&0.8411&620.1037&0.9075&427.2611&0.8623&0.7915\\\hline
       
  \end{tabular}%
  }
 \caption{Case (ii), Bigram Granularity Terms in $DQI_{C6}$}
\end{table}
\begin{table}
\resizebox{1.0\columnwidth}{!}{%
  \begin{tabular}{llllllll}
  \hline
     
    \multirow{3}{*}{\textbf{Sample Set}} &
      \multicolumn{7}{c}{\textbf{Terms}}\\
      &\textbf{entailment}&&\textbf{neutral}&&\textbf{contradiction}&&\\
      &\textbf{T1}&\textbf{T2}&\textbf{T1}&\textbf{T2}&\textbf{T1}&\textbf{T2}&\textbf{T5}\\\hline
      
    \textbf{Original} &1567.0110&0.7652&2174.6543&0.7302&1135.1086&0.7193&1.7297\\
    \textbf{+S1} &1567.0110&0.7652&2174.6543&0.7302&1154.0280&0.7094&1.7212\\
    \textbf{+S2} &1567.0110&0.7652&2174.6543&0.7302&1157.8255&0.8636&1.7298\\
    \textbf{+S3} &1567.0110&0.7652&2215.9640&0.7163&1135.1086&0.7193&1.6799\\
    \textbf{+S4} &1567.0110&0.7652&2245.9485&0.7355&1135.1086&0.7193&1.7383\\
    \textbf{+S5} &1517.6459&0.7571&2174.6543&0.7302&1135.1086&0.7193&1.7468\\
    \textbf{+S6} &1642.3849&0.7601&2174.6543&0.7302&1135.1086&0.7193&1.7383\\
    \textbf{+S7} &1593.6394&0.7615&2174.6543&0.7302&1135.1086&0.7193&1.7406\\
    \textbf{+S8} &1529.5108&0.7521&2174.6543&0.7302&1135.1086&0.7193&1.7470\\
    \textbf{+S9} &1567.0110&0.7652&2204.5792&0.7324&1135.1086&0.7193&1.7470\\
    \textbf{+S10}&1567.0110&0.7652&2190.9585&0.7245&1135.1086&0.7193&1.7235\\
    \textbf{+S11} &1567.0110&0.7652&2174.6543&0.7302&1199.7393&0.7288&1.7470\\
    \textbf{+S12} &1567.0110&0.7652&2174.6543&0.7302&1199.7393&0.7288&1.7383\\\hline
       
  \end{tabular}%
  }
 \caption{Case (ii), Trigram Granularity Terms in $DQI_{C6}$}
\end{table}

\begin{table}
\resizebox{1.0\columnwidth}{!}{%
  \begin{tabular}{lllllll}
  \hline
     
    \multirow{3}{*}{\textbf{Sample Set}} &
      \multicolumn{6}{c}{\textbf{Terms}}\\
      &\textbf{entailment}& &\textbf{neutral}& &\textbf{contradiction}& \\
      &\textbf{T3}&\textbf{T4}&\textbf{T3}&\textbf{T4}&\textbf{T3}&\textbf{T4}\\\hline
      
    \textbf{Original} &0.1846&0.2003&0.1465&0.1226&0.1008&0.3662\\
    \textbf{+S1} &0.1846&0.2003&0.1465&0.1226&0.1037&0.3485\\
    \textbf{+S2} &0.1846&0.2003&0.1465&0.1226&0.1046&0.3514\\
    \textbf{+S3} &0.1846&0.2003&0.1480&0.1195&0.1008&0.3662\\
    \textbf{+S4} &0.1846&0.2003&0.1448&0.1195&0.1008&0.3662\\
    \textbf{+S5} &0.1811&0.1894&0.1465&0.1226&0.1008&0.3662\\
    \textbf{+S6} &0.1712&0.2065&0.1465&0.1226&0.1008&0.3662\\
    \textbf{+S7} &0.1923&0.1931&0.1465&0.1226&0.1008&0.3662\\
    \textbf{+S8} &0.1824&0.1887&0.1465&0.1226&0.1008&0.3662\\
    \textbf{+S9} &0.1846&0.2003&0.1484&0.1197&0.1008&0.3662\\
    \textbf{+S10} &0.1846&0.2003&0.1464&0.1191&0.1008&0.3662\\
    \textbf{+S11} &0.1846&0.2003&0.1465&0.1226&0.1033&0.3473\\
    \textbf{+S12} &0.1846&0.2003&0.1465&0.1226&0.1033&0.3473\\\hline
       
  \end{tabular}%
  }
 \caption{Terms 3 and 4 in $DQI_{C6}$ for Case (ii)}
\end{table}

\begin{table}
\centering
\resizebox{0.5\columnwidth}{!}{%
  \begin{tabular}{lllllll}
  \hline
     
    \multirow{1}{*}{\textbf{Sample Set}} &
      \multicolumn{1}{c}{\textbf{DQI C6}}&
\\\hline      
\textbf{Original} &228.3537\\
\textbf{+S1} &202.4647\\
\textbf{+S2} &197.6054\\
\textbf{+S3} &196.3454\\
\textbf{+S4} &196.1489\\
\textbf{+S5} &200.7986\\
\textbf{+S6} &213.8920\\
\textbf{+S7} &202.4102\\
\textbf{+S8} &202.2893\\
\textbf{+S9} &198.4766\\
\textbf{+S10} &202.7345\\
\textbf{+S11} &200.9509\\
\textbf{+S12} &197.8010\\\hline
     
  \end{tabular}%
  }
 \caption{$DQI_{C6}$ for Case (ii)}
\end{table}

\begin{table}
\resizebox{1.0\columnwidth}{!}{%
  \begin{tabular}{llll}
  \hline
     
    \multirow{2}{*}{\textbf{Sample Set}} &
      \multicolumn{3}{c}{\textbf{DQI C7}}\\
      &SSIM=0.2&SSIM=0.3&SSIM=0.4  \\\hline
      
    \textbf{Original} & 0.00304989& 0.00421324& 0.00629840\\
    \textbf{+S1} & 0.00189475& 0.00229266& 0.00290212\\
    \textbf{+S2} & 0.00216703& 0.00270372& 0.00359374\\
    \textbf{+S3} & 0.00186796& 0.00225356& 0.00283975\\
    \textbf{+S4} & 0.00196072& 0.00238996& 0.00305981\\
    \textbf{+S5} & 0.00188903& 0.00228429& 0.00288872\\
    \textbf{+S6} & 0.00190351& 0.00230549& 0.00292271\\
    \textbf{+S7} & 0.00201427& 0.00247000& 0.00319224\\
    \textbf{+S8} & 0.00187124& 0.00225832& 0.00284732\\
    \textbf{+S9} & 0.00197442& 0.00241034& 0.00309330\\
    \textbf{+S10}&0.001886216& 0.00228017& 0.00288214\\
    \textbf{+S11}&0.002048964& 0.00252237& 0.00328026\\
    \textbf{+S12}&0.002076182& 0.00256374& 0.00335058\\\hline
       
  \end{tabular}%
  }
 \caption{$DQI_{C7}$ for Case (ii)}
\label{tab:60}
\end{table}

\definecolor{myRed}{rgb}{0.97,0.36,0.3}
\definecolor{myYellow}{rgb}{0.96,0.74,0.37}
\definecolor{myGreen}{rgb}{0.56,0.93,0.56}

\begin{table}
\centering
\scriptsize
\resizebox{1.0\columnwidth}{!}{%
  \begin{tabular}{llll}
  \hline
    \textbf{Sample} & \textbf{Overlap Count}& \textbf{length(hypothesis)}\\
    & & \textbf{/ Overlap Count}\\
    \hline
      
    \textbf{S1} & 3& \cellcolor{myRed}\textbf{2.0000}\\
    \textbf{S2} & 2& \cellcolor{myRed}\textbf{1.5000}\\
    \textbf{S3} & 8& \cellcolor{myRed}\textbf{1.1250}\\
    \textbf{S4} & 1& \cellcolor{myGreen}10.0000\\
    \textbf{S5} & 2& \cellcolor{myRed}\textbf{3.5000}\\
    \textbf{S6} & 2& \cellcolor{myYellow}\textbf{5.5000}\\
    \textbf{S7} & 1& \cellcolor{myYellow}\textbf{4.0000}\\
    \textbf{S8} & 2& \cellcolor{myRed}3.5000\\
    \textbf{S9} & 0& \cellcolor{myGreen}\textbf{40.0000}\\
    \textbf{S10} & 2& \cellcolor{myRed}3.5000\\
    \textbf{S11} & 1& \cellcolor{myYellow}\textbf{5.0000}\\
    \textbf{S12} & 3& \cellcolor{myRed}3.0000\\\hline
       
  \end{tabular}%
  }
 \caption{Word Overlap, Red: $<$ 3.9375, Yellow: 3.9375-9.8333 Green: $>$ 9.8333}
 \label{tab:dqic5i56}
\end{table}

\begin{table}
\centering
\scriptsize
\resizebox{1.0\columnwidth}{!}{%
  \begin{tabular}{llll}
  \hline
    \textbf{Sample} & \textbf{Overlap Count}& \textbf{length(hypothesis+premise)}\\
    & & \textbf{/ Overlap Count}\\
    \hline
      
    \textbf{S1} & 3& \cellcolor{myRed}\textbf{3.3333}\\
    \textbf{S2} & 2& \cellcolor{myRed}\textbf{3.0000}\\
    \textbf{S3} & 8& \cellcolor{myRed}\textbf{2.3750}\\
    \textbf{S4} & 1& \cellcolor{myYellow}\textbf{13.0000}\\
    \textbf{S5} & 2& \cellcolor{myRed}\textbf{4.5000}\\
    \textbf{S6} & 2& \cellcolor{myYellow}\textbf{7.0000}\\
    \textbf{S7} & 1& \cellcolor{myYellow}\textbf{7.0000}\\
    \textbf{S8} & 2& \cellcolor{myRed}5.0000\\
    \textbf{S9} & 0& \cellcolor{myGreen}\textbf{70.0000}\\
    \textbf{S10} & 2& \cellcolor{myRed}5.5000\\
    \textbf{S11} & 1& \cellcolor{myYellow}\textbf{11.0000}\\
    \textbf{S12} & 3& \cellcolor{myRed}4.6667\\\hline
       
  \end{tabular}%
  }
 \caption{Word Overlap, Red: $<$ 5.5347, Yellow: 5.5347-17.1944 Green: $>$ 17.1944}
\end{table}

\begin{table}
\centering
\scriptsize
\resizebox{1.0\columnwidth}{!}{%
  \begin{tabular}{llll}
  \hline
    \textbf{Sample} & \textbf{Premise Word Count}& \textbf{Hypothesis Word Count}&\textbf{Sum of Word}\\
    & & &\textbf{ Similarities}\\
    \hline
      
    \textbf{S1} &10&9& \cellcolor{myGreen}5.4753\\
    \textbf{S2} &6 &7& \cellcolor{myGreen}2.7865\\
    \textbf{S3} &12 &15& \cellcolor{myYellow}\textbf{8.9008}\\
    \textbf{S4} &15 &6& \cellcolor{myYellow}\textbf{9.8715}\\
    \textbf{S5} &9 &3& \cellcolor{myGreen}6.5202\\
    \textbf{S6} &17 &6& \cellcolor{myRed}\textbf{29.0358}\\
    \textbf{S7} & 7&6& \cellcolor{myGreen}\textbf{3.6143}\\
    \textbf{S8} & 12&7& \cellcolor{myGreen}\textbf{6.5335}\\
    \textbf{S9} & 7&5& \cellcolor{myGreen}\textbf{3.6679}\\
    \textbf{S10} & 127&7& \cellcolor{myGreen}\textbf{6.0583}\\
    \textbf{S11} & 9&12& \cellcolor{myGreen}\textbf{4.3558}\\
    \textbf{S12} & 12&9& \cellcolor{myRed}28.5806\\\hline
       
  \end{tabular}%
  }
 \caption{Word Similarity With Stop Words, Red: $>$ 10.4317, Yellow: 8.8017-10.4317 Green: $<$ 8.8017}
\end{table}

\begin{table}
\centering
\scriptsize
\resizebox{1.0\columnwidth}{!}{%
  \begin{tabular}{llll}
  \hline
    \textbf{Sample} & \textbf{Premise Word Count}& \textbf{Hypothesis Word Count}&\textbf{Sum of Word}\\
    & & &\textbf{ Similarities}\\
    \hline
      
    \textbf{S1} &6&4& \cellcolor{myYellow}\textbf{5.3800}\\
    \textbf{S2} &3 &3& \cellcolor{myGreen}2.9008\\
    \textbf{S3} &10 &9& \cellcolor{myRed}\textbf{8.8910}\\
    \textbf{S4} &10 &3& \cellcolor{myRed}\textbf{7.9413}\\
    \textbf{S5} &7 &2& \cellcolor{myYellow}\textbf{6.0292}\\
    \textbf{S6} &11 &3& \cellcolor{myRed}\textbf{9.7704}\\
    \textbf{S7} & 4&3& \cellcolor{myGreen}\textbf{3.6234}\\
    \textbf{S8} & 7&3& \cellcolor{myYellow}\textbf{6.2102}\\
    \textbf{S9} & 4&3& \cellcolor{myGreen}\textbf{3.1786}\\
    \textbf{S10} & 7&4& \cellcolor{myYellow}\textbf{6.2102}\\
    \textbf{S11} & 5&6& \cellcolor{myGreen}\textbf{4.3768}\\
    \textbf{S12} & 9&5& \cellcolor{myRed}7.8905\\\hline
       
  \end{tabular}%
  }
 \caption{Word Similarity Without Stop Words, Red: $>$ 6.8188, Yellow: 5.2483-6.8188 Green: $<$ 5.2483}
 \label{tab:dqic5i56e}
\end{table}

\section{AutoFix}
% AutoFix algorithm has its roots in TextFooler algorithm. 
The crowdsource workers are provided the option of seeking assistance with improving sample quality using AutoFix. The aim is to modify a crowd worker's created hypothesis, without changing the label.

A crowdsource worker can potentially ignore the AutoFix  option completely, use a mix of manual/AutoFix modifications to their sample, or repeatedly use AutoFix to generate the highest possible quality sample after the initial review of their sample. This means that the AutoFix operation requires strict control.

By incrementally changing the hypothesis, it is possible for a worker to understand how and why their hypothesis requires modification. It also makes it easier for them to see how each change possibly changes a DQI component. This allows them to create an initial better quality sample the next time around, thereby improving their sample generation rate. It also helps ensure that workers don't get frustrated at a potential inability to generate high quality samples, a case which will arise increasingly as the dataset size becomes larger, and continue with new sample creation.

\begin{figure*}
\includegraphics[width=\linewidth]{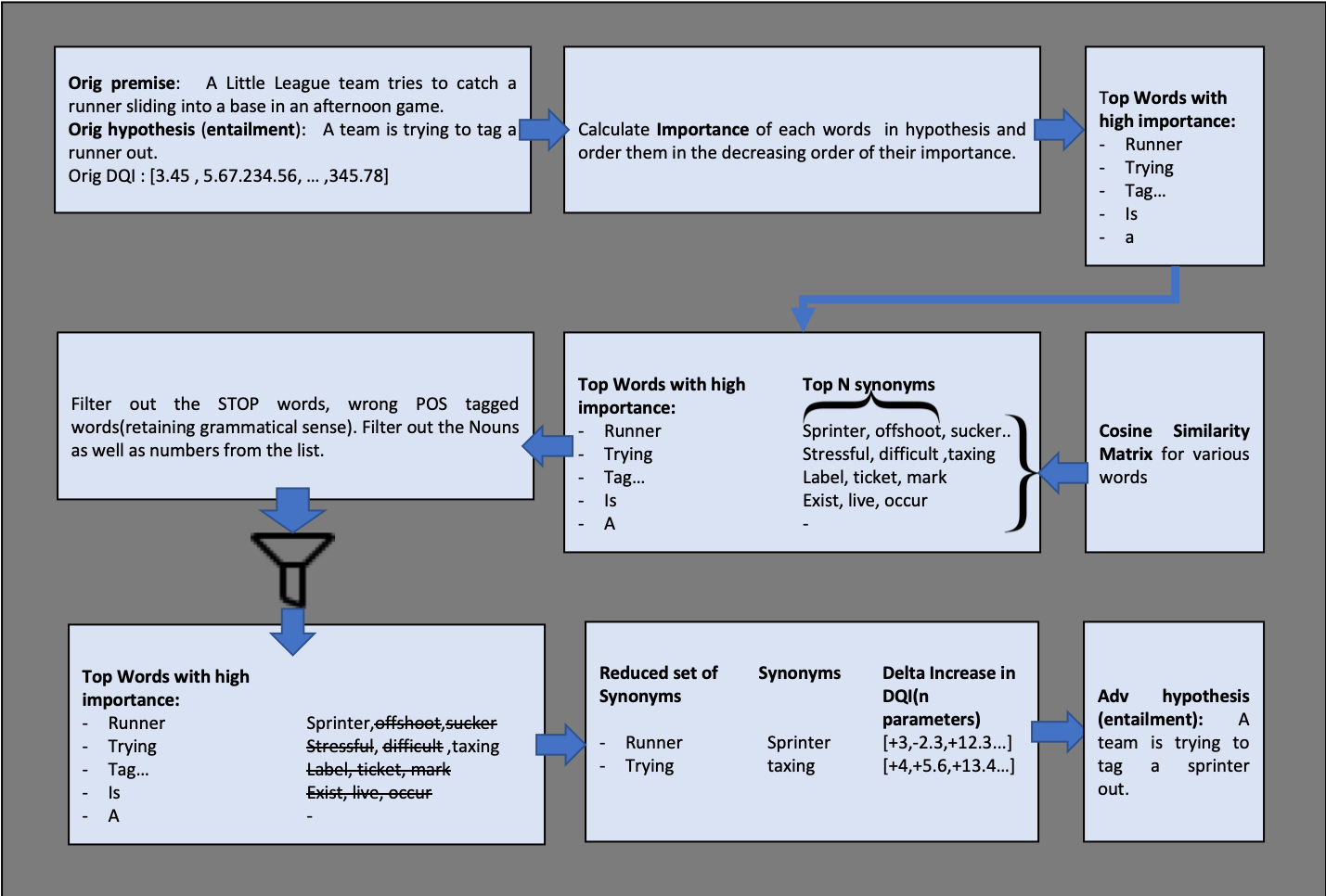}
   \caption{AutoFix Algorithm}
\label{fig:AutoFix}
\end{figure*}

Figure \ref{fig:AutoFix} explains the algorithm of AutoFix. We find important parts of the sentence and then replace those parts in order of their importance until the DQI color changes to green. The design of DQI component colors provides flexibility in Autofix, to make changes specifically according to those components that require the most fixing, i.e., those which are red. We therefore can use Autofix to create a benchmark dataset.
\section{Textfooler}\label{tfool}
Although Autofix assists a crowdsource worker in creating a better quality sample, the quality of the data sample submitted might still be too low. This might be because the crowdsource worker does not use the autofix option. It might also be due to a limited range of acceptable quality, which requires stricter analyst control, such as certain critical applications in Bio NLP. Sakaguchi et.al. \cite{sakaguchi2019winogrande} discard the bad split's data, i.e. data of lower quality, in the original setting of AFLite. In order to utilize this lower quality data we use Textfooler\cite{jin2019bert}. Textfooler's original aim is to confuse a model by strategically changing certain words of samples. The replaced words are always synonyms, and the semantics of the sentences is retained. However, the model still flips the label on the altered samples. 

\begin{figure*}
\centering
\includegraphics[width=16cm]{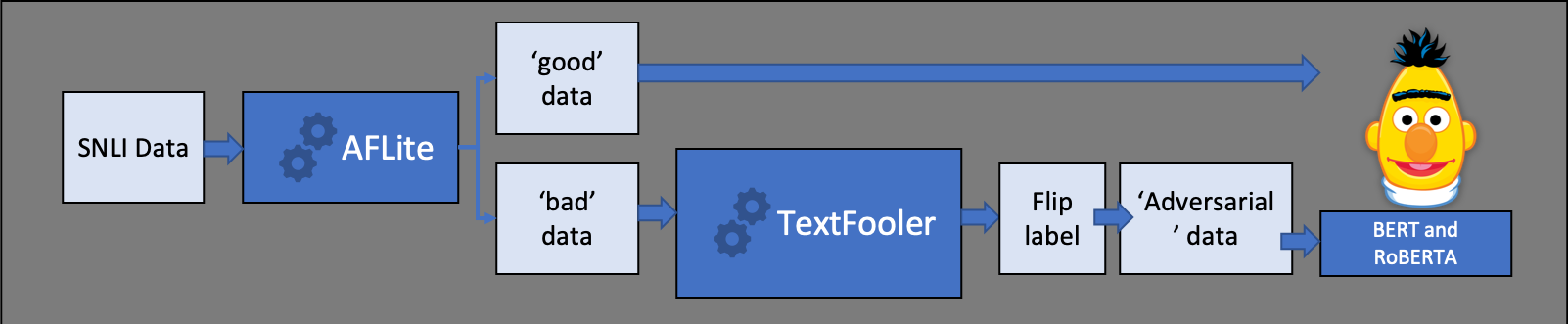}
  \caption{Role of Textfooler}
\label{fig:textfooler}
\end{figure*}

\begin{figure*}
\includegraphics[width=\linewidth,height=20cm]{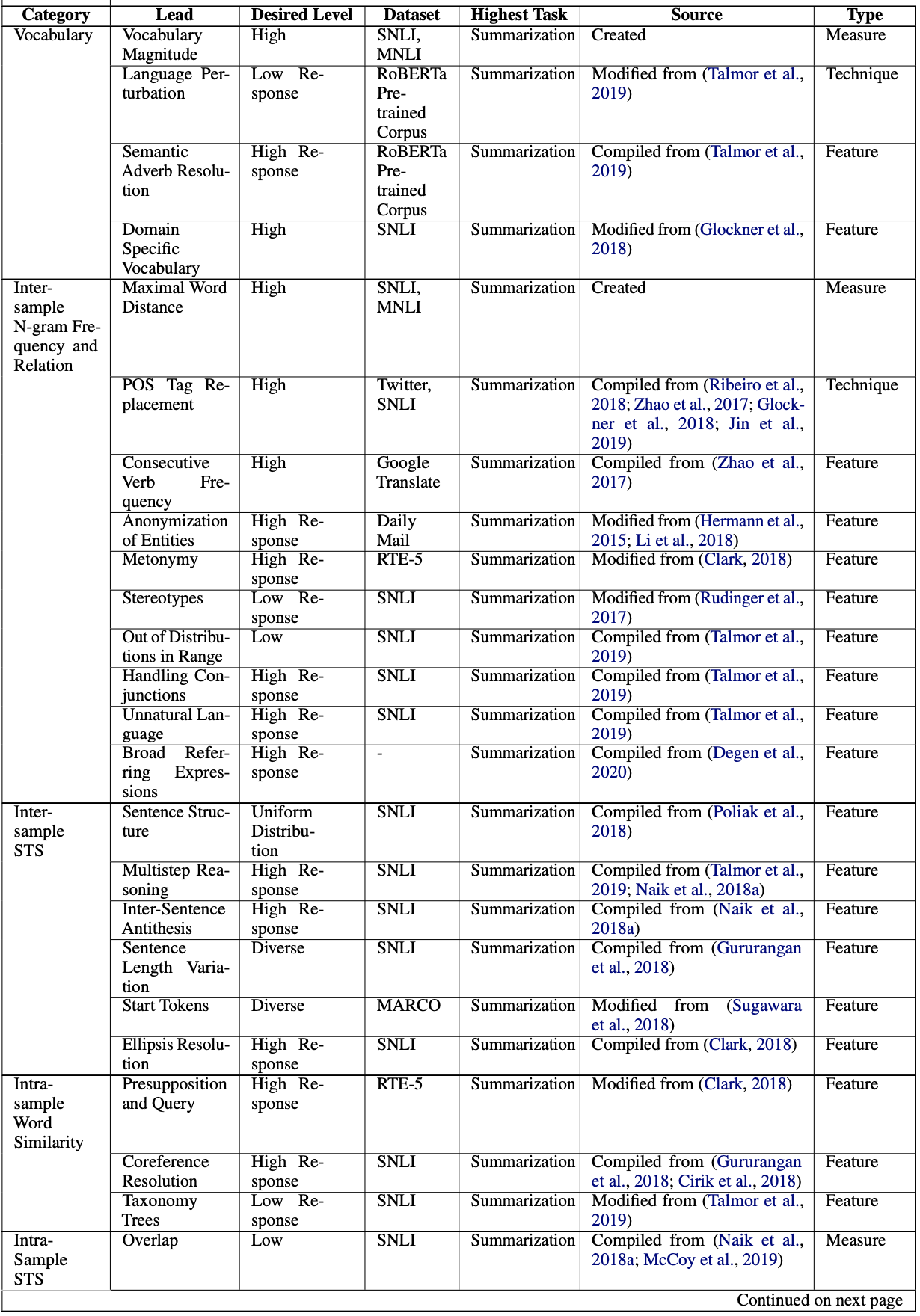}
   \caption{Detailed Information on Leads - I}
\label{fig:LeadTable1}
\end{figure*}

\begin{figure*}
\includegraphics[width=\linewidth,height=20cm]{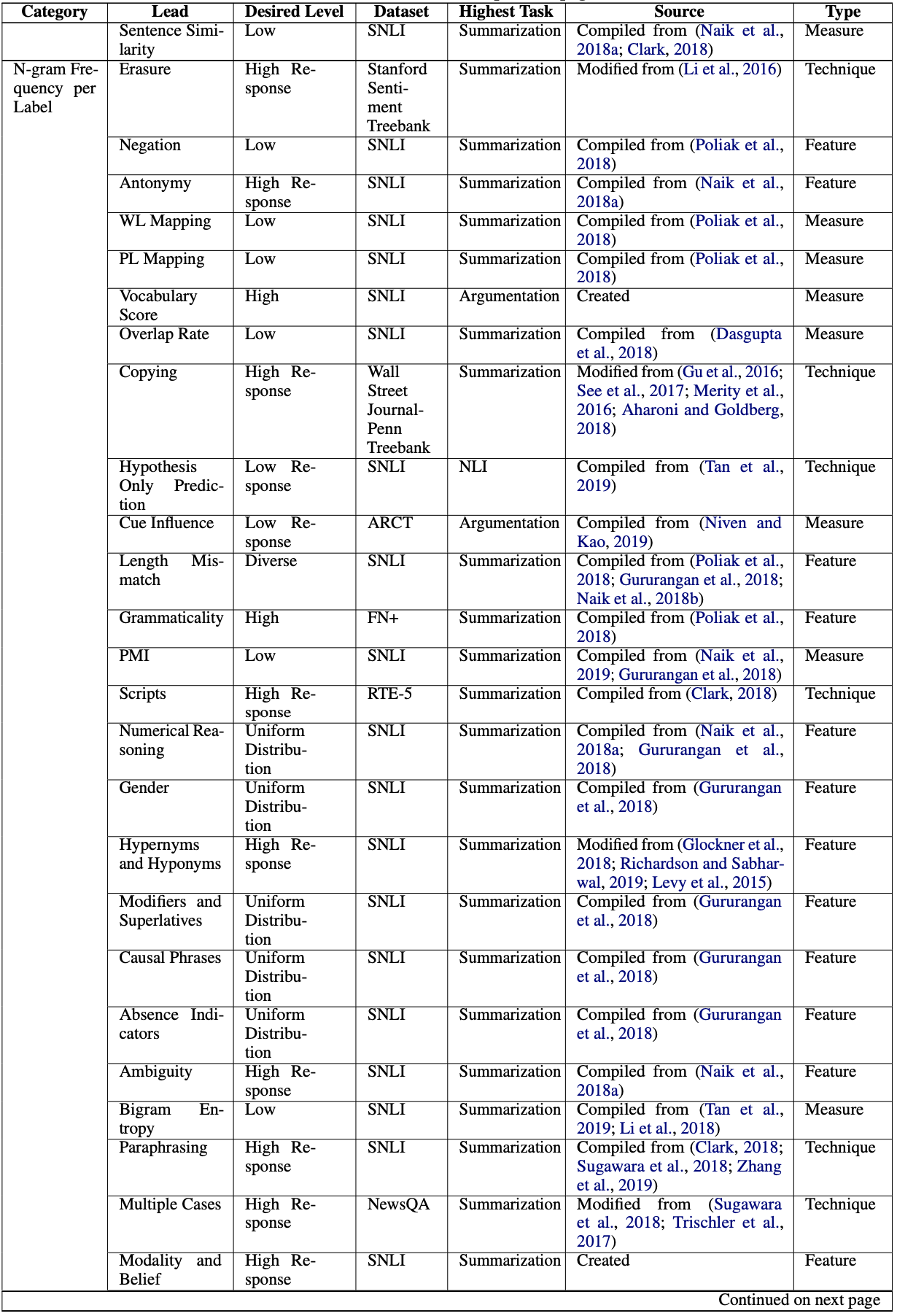}
   \caption{Detailed Information on Leads - II}
\label{fig:LeadTable2}
\end{figure*}

\begin{figure*}
\includegraphics[width=\linewidth,height=15cm]{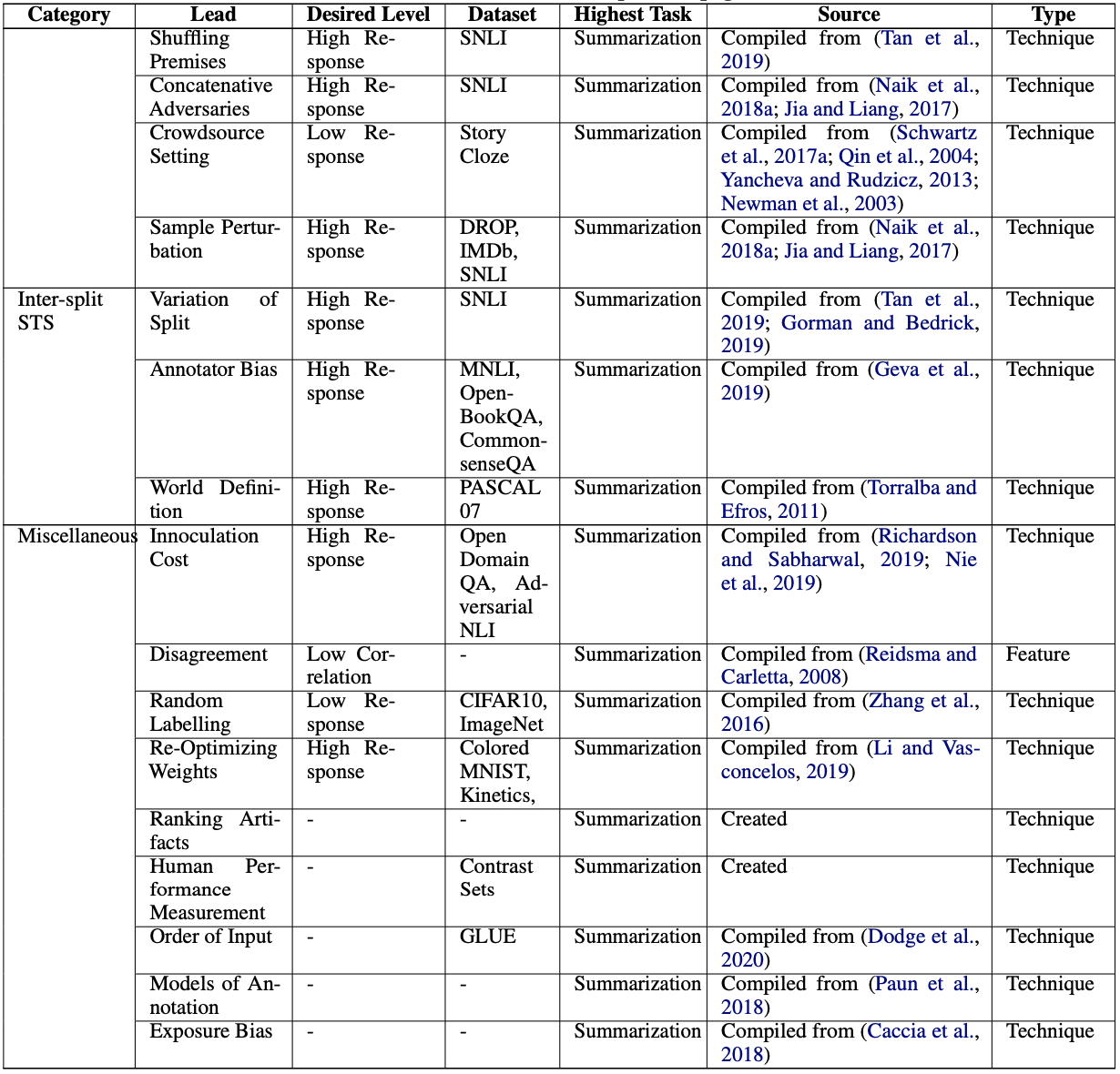}
   \caption{Detailed Information on Leads - III}
\label{fig:LeadTable3}
\end{figure*}

We use Textfooler to generate adversarial examples from low quality samples. That is, we convert low quality data to higher quality data, ensuring that the crowdsourcing effort is not wasted. This option is offered to the analyst when they are reviewing the submitted samples. This leads to much less wastage of the effort and resources involved in crowd-sourced data creation. This process is shown in Figure \ref{fig:textfooler}.

\subsection{Is Textfooler useful?}
To verify if Textfooler helped fix bad data, we perform a series of experiments, the results of which are in  Tables \ref{tab:BertTextfooler} and \ref{tab:RoBertaTextfooler}. The evaluation was also done on out of distribution datasets such as ANLI and Stress Test to evaluate the generalization ability of the models. 

\begin{table*}
\resizebox{2.0\columnwidth}{!}{%
  \begin{tabular}{lllll}
  \hline
      \textbf{No. of Adversaries}&\textbf{Test Set}&\textbf{Stress Test}&\textbf{ANLI}&\textbf{Test with Adversaries}\\\hline
 0& 0.517560074 & 0.513082991 & \textbf{0.32166302}  & 0.441785714 \\
41753     & \textbf{0.522489217} & 0.518961469 & 0.318224445 & 0.702857143 \\
162766    & 0.513247073 & 0.522762209 & 0.320100031 & 0.722738095 \\
204787    & 0.492914356 & \textbf{0.535634048} & 0.312285089 & 0.734761905 \\
289160    & 0.507701787 & 0.533201574 & 0.312910284 & \textbf{0.750714286} \\\hline
%Full data & 0.492914356 & 0.520853393 & 0.311659894 & 0.74952381  \\
\textbf{Size of the Test Set} & 1623        & 59199       & 3199        & 8400       \\\hline
  \end{tabular}%
  }
\caption{BERT Textfooler experiments}
\label{tab:BertTextfooler}
\end{table*}

\begin{table*}
\resizebox{2.0\columnwidth}{!}{%
  \begin{tabular}{lllll}
  \hline
      \textbf{Actual No. of Lines}&\textbf{Test Set}&\textbf{Stress Test}&\textbf{ANLI}&\textbf{Test with Adversaries}\\\hline
  0 & \textbf{0.590881084} & \textbf{0.74474231}  & \textbf{0.339481088} & \textbf{0.595833333} \\
  41753  & 0.585951941 & 0.737107046 & 0.336355111 & 0.523690476  \\\hline
% Full data              & 0.585951941 & 0.71658305  & 0.315411066 & 0.501428571           \\

 \textbf{Size of the Test Set}   & 1623        & 59199       & 59199       & 8400  \\\hline
  \end{tabular}%
  }
\caption{RoBERTa Textfooler experiments}
\label{tab:RoBertaTextfooler}
\end{table*}

\subsection{Results}
Using adversarial data for BERT, we see that in-sample accuracy decreases and out-of-sample accuracy increases in case of Test with Adversaries and Stress Test \cite{naik2018stress}. It decreases slightly for ANLI \cite{nie2019adversarial}. In the case of ROBERTA, we see a slight decrease in in-sample accuracy, but the out-of-sample accuracy also decreases unexpectedly. We analyze this process further to understand the issue.
% \subsection{Error analysis}
% We see a mixed trend in the results in our experiments. Looking at the BERT results we see, (i) overall slight decrease for in sample test. Even though we see slight increase in the beginning when we added the first batch of adversaries in the beginning. (ii) an increase in the accuracy for stress test set even though it is not very significant. (iii) a slight fluctuations in Adversarial SNLI test set. 
% We performed one more set of experiment for in sample accuracies. We used textfooler to generate adversaries for the 'bad' test samples as well and added them to the test set for analysis. Here we saw a major increase in the accuracies. 
% Looking at RoBERTa's as shown in Table \ref{tab:RoBertaTextfooler} we see, a slight decrease in all the cases except for the last case of test dataset with adversarial examples. 

\subsection{Error Analysis}
We find a few issues with Textfooler, which are in accordance with the authors' observation: "Our adversarial samples are susceptible to three types of
errors: word sense ambiguity, grammatical error, and task-sensitive content shift" \cite{jin2019bert}.

The task-sensitive content shift turns out to be the root cause of the issue. When adversaries are generated for a sample, we expect that the label of the sample should remain unchanged as the semantics of the sample remains unchanged. However,  we find that Textfooler is actually changing labels in many cases, such as with the replacement of numbers and certain nouns, as shown in Table \ref{tab:tferror}. Also, ROBERTA appears to be a bit harder to fool than BERT using Textfooler. 

% When Textfooler causes a label change while generating an adversarial sample, the essence of adversarial sample generation is therefore lost. For example:
% The notion of adversarial generation is to test if the model has correctly learned to solve the original sample, by checking if it fails or passes on the adversarial sample.
% For the generation of adversarial samples task sensitive content shift turned out to be a major road block. For adversaries we assume when we change some words by their synonyms the interpretation of the sentence should remain the same and thus the label should not change. We basically are blaming model to be not robust enough for these samples. Instead in many cases the model was correct and the fact we changed the label back to original did us more harm than good. A few examples to illustrate this are,

\begin{table*} 
\centering
\scriptsize
\resizebox{2.0\columnwidth}{!}{%
\begin{tabular}{lllll}
\hline
\textbf{Original Premise} & \textbf{Original Hypothesis} & \textbf{Original Label} & \textbf{Adversarial Hypothesis} & \textbf{New Label} \\
\hline
A man, woman, and child & A family of \textbf{three} is at a beach. & entailment & A family of \textbf{four} is at a beach. & contradiction\\
enjoying themselves on a beach.&&&&\\\hline
A boy is jumping on skateboard & The boy skates down the \textbf{sidewalk}. & contradiction & The boy skates down the \textbf{pavement}. & neutral\\
in the middle of a red bridge.&&&&\\\hline
              &            &            &        &          
\end{tabular}%
}
\caption{Incorrect Sample Modification in TextFooler due to Label Change}
\label{tab:tferror}
\end{table*}

% As illustrated in the above examples, the change in the Number or the Noun can be very dangerous. Such changes causes the actual meaning of the sentence to be changed completely. We address these issues by proposing a new algorithm called AutoFix explained in the next section. 
\section{Active Learning} 
DQI provides control over various properties of text. Our proposed data creation paradigm allows us to best utilize this control in the form of DQI component colors as shown in Fig \ref{fig:ourapproach}. We create our first set of benchmarks with the help of Autofix, using our default hyper parameters, as discussed in Section \ref{dqieval}. We run models and consider only those samples which are not correctly classified. We then run DQI on them to find out which DQI components are sensitive. For the sensitive DQI components, we shrink the allowed range of hyper-parameters (corresponding to the green colored DQI components) by 20 \%. We then collect our second benchmark which is harder than the first. We repeat the process once more, to generate a third benchmark, which is turns out to be the hardest of the three.

\section{Expert Review} 
The motivation behind DQI, and the method used to compute it must be thoroughly understood by a reviewer, in order to properly judge the interface design. Evaluation of the visualizations designed for the analyst interface, in particular, requires that reviewers have specialized knowledge and skills in the fields. To this end, a structured evaluation using a small set of graduate student researchers who are experts in either Data Visualization or NLP is done, i.e., the Expert Review method of evaluation\cite{elmqvist2015patterns}.

The experts are first presented  with the process flow (figure 2) and model (figure 1). Then, they are taken through the interface's functioning from both the crowd source worker and analyst perspectives(UI section figures). Finally, they review the visualizations used for each DQI components (figures from Visualization Section) and provide feedback:
\subsection{Insights}
\paragraph{1. Interface Aesthetic}
The color palettes used in the interfaces, and in particular, the use of  'traffic light' colors to indicate component-wise quality were appreciated. One concern expressed was the necessity of recoloring in the case of red-green colorblindness. This will be taken into account in future iterations. The panel placement was found satisfactory.  Also, the decision to make both the panel containing the history line chart and the rank box plot, and the instructions panel minimizable ones was commended as helping prevent the overloading of crowdsource workers. Similarly, loading the visualizations on separate tabs from the main interface was judged to prevent analyst overload, as well as account for high readability of multi-granular information irrespective of screen size. \paragraph{2. Reducing Plot Coverage}
Suggestions to limit the data represented in the  bubble plot, tree map, and kde curve plot, in a fashion similar to that of the force layout and parallel coordinates plots were made. This might improve the readability, help capture distribution skews at an earlier stage, and make the impact of sample addition more discernible in the face of a large set of preexisting samples . Future work will accordingly be directed towards analyzing different subsets of a dataset. This might also prove to be helpful in studying the impact of data ordering on bias removal.
\paragraph{3. Navigation}
The tooltip and button/dropdown controls were found to be fairly intuitive. A recommendation was made to have a minimizable/pop-up instructions panel similar to the main interface, for each visualization. This would detail the interactions present, as well as the intent behind the component's formula to direct the analysts' inference patterns. An analyst could therefore possibly learn how to effectively interpret the visualization more quickly. The actual frequency of an analyst's navigation to the visualizations was questioned. We believe that those samples that require analyst intervention via TextFooler will be the ones that require visualization navigation. As the DQI changes are affected by preexisting dataset size, the frequency of visualization usage should increase with increasing dataset size. This is because users will have the potential to submit a greater number of samples that require analyst intervention.
\paragraph{4. Annotation}
Annotation was found to be satisfactory across visualizations. Providing a minimizable panel control which contained the new sample was suggested.
\paragraph{5. Additional Features}
It was proposed that having a date element in charts to view the history of visualization transformations in accordance with quality might prove beneficial to the analyst. This could also possibly be used as a reference point for TextFooler fixed samples, in judging the amount and nature of change produced, along with the impact of this change on components and their visualizations. The crowdsource workers could also be fairly judged based on this history, if they join the data creation phase at an intermediate stage, where there already exists a significant number of data samples.
A question was raised regarding the viability of the extension of the interface to support additional natural language tasks. As mentioned in (section 2), the bias leads listed were mined using a specific task ordering based on incremental amounts of input and output data. The formulas use text properties to capture bias, which are not task specific. Therefore, we expect that the interface will only need to be changed in so far as the text fields used to display the sample data.
\paragraph{6. Learning Curve}
There was a consensus that the crowd source worker interface was straightforward for potential lay users to navigate. The analyst interface has a moderate learning curve, that is mainly attributed to the necessity of effectively interpreting visualizations used for each component. Once this is achieved, the analyst interface proves to be efficient in judging sample quality.
\section{Future Directions}
\paragraph{Weighting DQI formuale:} In this paper, we have considered DQI components individually. As we have seen in the test cases section, some components need to be scaled based on the size of dataset. It will be interesting to learn a formula consisting of all these DQI components so that appropriate weights can be given to individual components.
\paragraph{Adversarial Filtering Algorithm"} We have seen in Section \ref{dqieval} that, AFLite may be missing certain types of artifacts represented by our DQI components. Our analysis can be used to come up with a new adversarial filtering algorithm based on DQI.
\paragraph{Adversarial Technique to Fool Model:} Textfooler seems to have certain issues which are sensitive to our application, as we have seen in Section \ref{tfool}. More experiments can be performed using other types of adversarial filtering algorithms in our
data creation paradigm.
\paragraph{Create High Quality Data Using Crowdsourcing:}
The benchmarks we have created in this paper are using automation methods: TextFooler and Autofix. Both of these automation methods are secondary. The primary option, of crowdsourcing using our data creation paradigm, must be implemented to create high quality data.
\paragraph{Expansion to other tasks and domains:} Since the problem we address in this paper is a core problem in Machine Learning, our proposed generic approach for NLP needs to be expanded to other domains, especially Vision and Speech.
\section{Conclusion} 
In order to address the problem of bias in datasets, we have implemented mechanisms to stem spurious biases during the data creation process. First, we have developed a generic formula for DQI based on bias leads identified from literature. We have evaluated DQI components on the retained and removed sets produced using AFLite on the SNLI Test set. The efficacy of DQI is proved for the addition of new samples from the SNLI Dev set in the cases of (i) cold start, and (ii) a preexisting set of samples. We have proposed a data creation paradigm which is augmented by several visualizations, designed to improve the analyst's understanding of data quality and the impact of a created data instance on the overall dataset quality.  Autofix has been introduced on the crowd source worker end to assist in the creation of higher quality data. Textfooler has been used to assist analysts, in repairing any low quality data created, by adversarial sample generation. Retraining BERT and ROBERTa on the higher quality, renovated SNLI dataset has resulted in an increase in their generalization capability on out of distribution datasets. We have applied DQI in an active learning setup to renovate the SNLI dataset and produce a series of benchmarks in an increasing hierarchy of hardness. DQI takes the process of dynamic dataset creation forward, and serves as a means of benchmarking the true progress of AI.

% In order to address the problem of bias in datasets, we first identify about 50 parameters by going through relevant literature. Then, we develop basic text properties that reflect these, and accordingly build a formula for DQI. This formula is tuned through experimentation. We have also adversarially attacked AFLite, and come up with different algorithmic implementations of RAFLite, as a way to implement DQI. A user interface schematic for crowd source worker use has been developed, and will be extended to support an expert user view as well. Our next step will be to integrate these components,run experiments on the algorithm performance, build a model and feed to the UI, simulate crowd source work to create a dataset (possibly math), and run a user study. 

% \section*{Acknowledgments}

% The acknowledgments should go immediately before the references. Do not number the acknowledgments section.
% Do not include this section when submitting your paper for review.

\bibliography{anthology,emnlp2020}
\bibliographystyle{acl_natbib}

\section{Supplemental Material}
\label{sec:supplemental}
Here, we provide more details of each of the 63 parameters along with examples for better illustration.
\label{sec:eg1}\paragraph{Vocabulary Magnitude:}
A dataset of size of 100k samples and 30k unique words will have a vocabulary magnitude of 0.3.\hyperref[sec:1]{(top)}
\label{sec:eg5}\paragraph{Language Perturbation:}
The substitution of words like 'and' or 'by' with fillers such as 'blah' helps check if the original words are being used as a part of the reasoning context or not.\hyperref[sec:5]{(top)}
\label{sec:eg6}\paragraph{Semantic Adverb Resolution:}
There is a difference in the contexts created by 'always', 'sometimes', 'often', and 'never.'\hyperref[sec:6]{(top)}
\label{sec:eg7}\paragraph{Domain Specific Vocabulary:}
The names of countries such as Syria, Canada, Mexico, etc., and nationalities, such as Indian, Swiss, etc. are not recognized by language models, and performance on instances containing these words is low.\hyperref[sec:7]{(top)}
\label{sec:eg2}\paragraph{Maximal Word Distance:}
A dataset that covers the scientific domain will have words dissimilar to more commonly used language.\hyperref[sec:2]{(top)}
\label{sec:eg3}\paragraph{POS Tag Replacement:}
Consider the word 'Jordan' in vocabulary, where the context is that Jordan refers to the country. An equivalent country name (of the same POS tag) like 'Russia' can be used for replacement. Jordan could also refer to a person's name, such as 'Michael Jordan'. In this case, on replacement, 'Michael Russia' will be generated. This case does not add an example that makes sense. So such samples are discarded based on the count of the bigrams generated on replacement. In TextFooler, consider the input “The characters, cast in impossibly contrived situations, are totally estranged from reality.” The output might be: “The characters, cast in impossibly engineered circumstances, are fully estranged from reality.”\hyperref[sec:3]{(top)}
\label{sec:eg4}\paragraph{Consecutive Verb Frequency:}
It has been observed that on translation from English to German and back, sentences such as 'She was cooking dressed for a wedding' drop the second verb on retranslation, and becoming 'She was cooking for a wedding.'\hyperref[sec:4]{(top)}
\label{sec:eg8}\paragraph{Anonymization of Entities:}

Original Version: Content: 'The BBC producer allegedly struck by Jeremy Clarkson will not press charges against the “Top Gear” host.' Question: Who hosts Top Gear? Answer:Jeremy Clarkson

Anonymized Version: Content: 'The $ent1$ producer allegedly struck by $ent2$ will not press charges against the “Top Gear” host.' Question: Who hosts Top Gear? Answer: $ent2$
\hyperref[sec:8]{(top)}
\label{sec:eg12}\paragraph{Metonymy:}
'If we don't get these papers in today, the suits will be after us.' Here, suits refers to business people.\hyperref[sec:12]{(top)}
\label{sec:eg18}\paragraph{Stereotypes:}
Word associations like 'cook' or 'dolls' with 'girls', or 'temples' with 'India' are a source of bias.\hyperref[sec:18]{(top)}
\label{sec:eg21}\paragraph{Out of Distributions in Range:}
'Sheila and I' and 'Sheila or I' have different contextual meanings which can't be solved by pattern correlation. 
'Jim, John and Bob are 14, 12, and 18. Who is the second oldest?' returns the correct answer. But if their ages are '1997', '2001', and '2010', then the system returns the wrong answer.\hyperref[sec:21]{(top)}
\label{sec:eg22}\paragraph{Unnatural Language:}
The sentence: 'She was [MASK] fast, she was rapid,' has different meanings if you substitute 'not' or 'very' in it.\hyperref[sec:22]{(top)}
\label{sec:eg49}\paragraph{Broad Referring Expressions:}
Generic terms like 'this', 'the', 'that', or 'it'  can be used to refer to objects on different occasions. These must be resolved to remove ambiguity.\hyperref[sec:49]{(top)}
\label{sec:eg9}\paragraph{Sentence Structure:}
If a majority of sentence structures follow passive voice, an active voice sentence won't be easily parsed.\hyperref[sec:9]{(top)}
\label{sec:eg10}\paragraph{Multistep Reasoning:}
'When comparing a 23, a 38 and a 31 year old, the [MASK] is oldest A. second B. first C. third.'\hyperref[sec:10]{(top)}
\label{sec:eg11}\paragraph{Inter-Sentence Antithesis:}
'It was [MASK] hot, it was really cold . A. not B. really.'\hyperref[sec:11]{(top)}
\label{sec:eg14}\paragraph{Sentence Length Variation:}
Sentences with less detail are shorter, and therefore more likely to be classified as entailment.\hyperref[sec:14]{(top)}
\label{sec:eg15}\paragraph{Start Tokens:}
The candidate answer resolution is restricted by starting “wh-” and “how many" expressions.\hyperref[sec:15]{(top)}
\label{sec:eg16}\paragraph{Ellipsis Resolution:}
'I went to the mall on Monday, and she on Sunday' can be unrolled as 'I went to the mall on Monday, and she went to the mall on Sunday.'\hyperref[sec:16]{(top)}
\label{sec:eg13}\paragraph{Presupposition and Query:}
'This ban is the first ban for YouTube in China.' Here, the statement asuumes that there is a ban, and the model must reason on whether the ban was the first, not on the existence of the ban.\hyperref[sec:13]{(top)}
\label{sec:eg17}\paragraph{Coreference Resolution:}
'Tom said that he would get it done.'Here, he refers to Tom.\hyperref[sec:17]{(top)}
\label{sec:eg19}\paragraph{Taxonomy Trees:}
'Horse and crow' are grouped as animal, but 'crow and horse' are grouped as birds. This is because 'crow' is closer to 'bird' on the taxonomy tree than 'animal.'\hyperref[sec:19]{(top)}
\label{sec:eg20}\paragraph{Overlap:}
'The dog sat on the mat' and 'The dog did not sit on the chair' contain significant overlap and hence can easily be solved. In HANS, consider the premise 'The judges heard the actors resigned' and 'The judges heard the actors'. If a model relied on overlap, it would mark this sample as entailment, even though the gold label is neutral. \hyperref[sec:20]{(top)}
\label{sec:eg55}\paragraph{Erasure:}
Consider the sample 'I took my daughter and her step sister to see a show at Webster hall . It is so overpriced I’m in awe.' Using a BI-LSTM, the minimal set of words identified for 'value' is 'It is so overpriced I’m in awe.'
\hyperref[sec:55]{(top)}
\label{sec:eg24}\paragraph{Similarity:}
Similarity indicates overlapping detail. For example, 'The bird sang' and 'The robin warbled outside the window as it looked for breakfast' have less overlap due to the presence of more detail in the second sentence.\hyperref[sec:24]{(top)}
\label{sec:eg25}\paragraph{Negation:}
'She was pleased' and 'She could do nothing that did not please her' might be labeled as contradiction, due to the presence of negation terms.\hyperref[sec:25]{(top)}
\label{sec:eg26}\paragraph{Antonymy:}
Simple binary opposites are 'hot' and 'cold'. Less direct opposites are words like 'winter' and 'summer'.\hyperref[sec:26]{(top)}
\label{sec:eg27}\paragraph{WL Mapping:}
'Humans' and' instruments are found to be indicators of entailment, 'tall' and 'win' that of neutral, and 'sleep' and 'no' of contradiction.
%\begin{equation}
$P(l/w)=\frac{p(w,l)}{p(w)\cdot p(l)}$
%\end{equation}
\hyperref[sec:27]{(top)}
\label{sec:eg28}\paragraph{PL Mapping:}
For the phrase ‘x was sentient….’ ; by identifying the nature of ‘x’, a model can infer the label without looking at the rest of the sentence . Such lexical semantic exploitation indicates that context is not used in solving such samples.
%\begin{equation}
$P(l/p)=\frac{p(p,l)}{p(p)\cdot p(l)}$
%\end{equation}
\hyperref[sec:28]{(top)}
\label{sec:eg29}\paragraph{Vocabulary Score:}
Consider the word 'move' in the entailment, neutral, and contradiction classes, with counts 200, 345, and 126 respectively. Then, the score vector would be [3 200 345 126].
\hyperref[sec:29]{(top)}
\label{sec:eg30}\paragraph{Overlap Rate:}
%\begin{equation}
$Overlap Rate=\frac{number of overlap words}{number of words in sample}$
%\end{equation}
\hyperref[sec:30]{(top)}
\label{sec:eg31}\paragraph{Copying:}
Copy all possible subset of words from the premise to the hypothesis iteratively, and check when the label changes.\hyperref[sec:31]{(top)}
\label{sec:eg32}\paragraph{Hypothesis Only Prediction:}
The sample: 'People raise dogs because they are obedient' and 'People raise dogs because dogs are obedient', benefits from considering hypothesis only as there is no coreference to be resolved.\hyperref[sec:32]{(top)}
\label{sec:eg33}\paragraph{Cue Influence:}
Let $k$ be a cue, $T_{j}$ be the set of tokens in the warrant for data point $i$  with label $j$, and $n$ be the total number of data samples.

Applicability: number of data points a cue occurs with one label but not the other
%\begin{equation}
$\alpha _{k}= \sum_{i=1}^{n} [\exists j,k \in T_{j}^{(i)}  k \notin T_{\neg j}^{(i)}]$
%\end{equation}
Productivity: proportion of applicable data points for which a cue predicts the correct answer
%\begin{equation}
$\pi _{k}= \frac{\sum_{i=1}^{n} 1[\exists j,k \in T_{j}^{(i)} \wedge k \notin T_{\neg j}^{(i)} \wedge y_{i}=j]}{\alpha_{k}}$
%\end{equation}
Coverage: proportion of applicable cases of a cue over the total number of data points
%\begin{equation}
$\xi _{k}= \frac {\alpha_{k}}{n}$
%\end{equation}
\hyperref[sec:33]{(top)}
\label{sec:eg34}\paragraph{Length Mismatch:}
The sample: 'She was happy with her bonus’ and 'She decided to celebrate her raise at work by eating out,’ is more likely to be labelled as neutral.\hyperref[sec:34]{(top)}
\label{sec:eg35}\paragraph{Grammaticality:}
Consider the sample’: She has no option’ and She has no way than the others’. This is more likely to be classified as 'non-entailment.'\hyperref[sec:35]{(top)}
\label{sec:eg36}\paragraph{PMI:}
%\begin{equation}
$PMI(word,label)=\log\frac{p(word,label)}{p(word)\cdot p(label)}$
%\end{equation}
\hyperref[sec:36]{(top)}
\label{sec:eg37}\paragraph{Scripts:}
Consider the sample: 'Canada's plans to launch a satellite, but U.S. officials say the launch is a disguised long-range missile test' and 'The U.S. fears that the Canadian satellite is a ruse to hide the testing of a missile.' There is a familiar script at play here. Countries want to test  military equipment, but don't want to be seen as testing them, so may try and hide or cover up the test. Other countries are worried about this form of deceit, and may try and put political pressure on the testing country in order to prevent deceit.\hyperref[sec:37]{(top)}
\label{sec:eg38}\paragraph{Numerical Reasoning:}
'There were two major bombings in less than a week, with 10 people killed by a car bomb south of Baghdad and more than 30 dead when a suicide bomber blew himself up in the capital.' Requires a sum of 30+10 to be calculated to address the hypothesis: 'In less than a week there were 2 major bombings in Iraq, killing more than 40 people.'\hyperref[sec:38]{(top)}
\label{sec:eg39}\paragraph{Gender:}
Using terms like 'woman' and 'boy' instead of 'person' or 'child' are indicative of non-entailment.\hyperref[sec:39]{(top)}
\label{sec:eg40}\paragraph{Hypernyms and Hyponyms:}
(i) Words like 'wolf' and 'dog' are both animals, but confusion may occur during hyponym resolution as a wolf is a wild animal.
(ii) A chair might serve as a superset for its legs, which is not a true hypernym.\hyperref[sec:40]{(top)}
\label{sec:eg41}\paragraph{Modifiers and Superlatives:}
Words like 'tall' or 'popular' and 'best' or 'first' are indicative of neutral label.\hyperref[sec:41]{(top)}
\label{sec:eg42}\paragraph{Causal Phrases:}
Sentences that contain causal words like 'due to', 'because of', 'consequently', etc. are indicative of neutral label.\hyperref[sec:42]{(top)}
\label{sec:eg43}\paragraph{Absence Indicators:}
The word 'sleep' indicates the absence of activity, and hence is used as an indicator of contradiction.\hyperref[sec:43]{(top)}
\label{sec:eg44}\paragraph{Ambiguity:}
'She had a black bat' requires context and knowledge to decide if 'bat' refers to an animal, or sports equipment.\hyperref[sec:44]{(top)}
\label{sec:eg45}\paragraph{Bigram Entropy:}
%\begin{equation}
% $H(c | w) = −\sum_{c}^{}p(c | w)\log p(c | w)    $
%\end{equation}
Object bias: For example, 'playing piano' is the only class depicting pianos. This can be inferred by searching for 'piano' or 'music'. 

Scene bias: For example, 'soccer juggling' can be resolved by searching for words like 'goal', 'net', or 'ball'. 

Person bias: For example, 'military marching' can be resolved by matching to words like 'army' or 'parade'.
\hyperref[sec:45]{(top)}

\label{sec:eg46}\paragraph{Paraphrasing:}
'Same' and 'replica' are paraphrases, but 'same' and 'about same' are not.

PAWS: Word Swapping: 'Can a bad person become good? : 'Can a good person become bad?'

PAWS: Back Translation: 'The team also toured in Australia in 1953.' : 'In 1953, the team also toured in Australia.'
\hyperref[sec:46]{(top)}
\label{sec:eg47}\paragraph{Multiple Cases:}

Context: [...] This plot of land is scheduled to house the permanent United Airlines Flight 93 memorial. [...] Question: What was the name of the flight?
Answer: 93
Possible answers: United Airlines Flight 93, Flight 93
Here, multiple choices have the correct span of 93 \cite{trischler-etal-2017-newsqa}.\hyperref[sec:47]{(top)}
\label{sec:eg48}\paragraph{Modality and Belief:}

Epistemic: Agatha must be the murderer. (necessity:neutral)

Deontic: Agatha must go to jail. (obligatory:neutral)

Circumstantial: Agatha must sneeze. (possibility:entailment)

Belief for the above case is true/false in order to label them.\hyperref[sec:48]{(top)}
\label{sec:eg51}\paragraph{Shuffling Premises:}
It is a method of iteratively substituting premises to check word correlation.\hyperref[sec:51]{(top)}
\label{sec:eg52}\paragraph{Concatenative Adversaries:}
Add distractor words at the end of hypotheses such as negation, superlatives, etc. to test the model's operation over the original samples.\hyperref[sec:52]{(top)}
\label{sec:eg56}\paragraph{Crowdsource Setting:}
The length of a contradiction hypothesis is generally shorter than that of the original premise, and it uses simpler language. 
\hyperref[sec:56]{(top)}
\label{sec:eg57}\paragraph{Sample Perturbation:}

Counterfactual Sample:

P: A young dark-haired woman crouches on the banks of a river while washing dishes.

OH: A woman washes dishes in the river while camping. (Neutral)

NH: A woman washes dishes in the river. (Entailment)

Contrast Set Sample:

Original Text: Two similarly-colored and similarly-posed cats are face to face in one image.

New Text: Two differently-colored but similarly-posed chow dogs are face to face in one image.
\hyperref[sec:57]{(top)}
\label{sec:eg53}\paragraph{Variation of Split:}
Different split variations are required for proper benchmarking, to ensure a true accuracy increase.
%\begin{equation}
$\widehat{\delta }=\mathit{M(G_{test},S_{1})} - \mathit{M(G_{test},S_{2})}$
%\end{equation}

Accuracy difference: $\widehat{\delta}$

Model: $M$

Test Set: $G_{test}$ 

Systems 1 and 2: $S_{1},S_{2}$

\hyperref[sec:53]{(top)}
\label{sec:eg54}\paragraph{Innoculation Cost:}

Adversarial NLI:

Premise:A melee weapon is any weapon used in direct hand-to-hand combat; by contrast with ranged weapons which act at a distance. The term “melee” originates in the 1640s from the French word, which refers to hand-to-hand combat, a close quarters battle, a brawl, a con- fused fight, etc. Melee weapons can be broadly divided into three categories

Hypothesis: Melee weapons are good for ranged and hand-to-hand combat.
\hyperref[sec:54]{(top)}

\label{sec:eg61}\paragraph{Disagreement:}
A particular annotator overuses the label of entailment, and marks very few samples as neutral. This pattern can be used as a bias by a model.
\hyperref[sec:61]{(top)}

\pagebreak

\pagebreak

\pagebreak

\end{document}